\documentclass{article}

\PassOptionsToPackage{numbers, compress}{natbib}
\usepackage[dandb, final]{neurips_2025} 

\usepackage[utf8]{inputenc} 
\usepackage[T1]{fontenc}    
\usepackage{hyperref}       
\usepackage{url}            
\usepackage{booktabs}       
\usepackage{amsfonts}       
\usepackage{nicefrac}       
\usepackage{microtype}      

\usepackage{times}
\usepackage{epsfig}
\usepackage{graphicx}
\usepackage{amsmath}
\usepackage{amssymb}
\usepackage{adjustbox}
\usepackage{multicol}
\usepackage{multirow}
\usepackage{array}
\usepackage{tabularx}
\usepackage{booktabs}
\usepackage{bbding}
\usepackage{array}
\usepackage{makecell}
\usepackage{tabularx}
\usepackage{xspace}
\usepackage{bm}
\usepackage{graphicx}
\usepackage{booktabs}
\usepackage{amsmath}
\usepackage{color}
\usepackage{float}
\usepackage[table, dvipsnames]{xcolor}
\PassOptionsToPackage{table, usenames, dvipsnames}{xcolor}
\usepackage{colortbl}
\usepackage{amssymb}
\usepackage{pifont}
\usepackage{acronym}
\usepackage{cleveref}
\usepackage{pdflscape}
\usepackage{subcaption}
\usepackage{rotating}
\usepackage{stackengine}
\usepackage[bottom]{footmisc}

\usepackage{minitoc}

\usepackage[toc,page,header]{appendix}
\usepackage{lipsum} 

\makeatletter
\newcommand{\startappendixtoc}{%
  \newwrite\appendixtocfile
  \immediate\openout\appendixtocfile=\jobname.apptoc
  \let\oldaddcontentsline\addcontentsline
  \renewcommand{\addcontentsline}[3]{%
    \oldaddcontentsline{##1}{##2}{##3}%
    \addtocontents{apptoc}{\protect\contentsline{##2}{##3}{\thepage}}%
  }%
}

\newcommand{\printappendixtoc}{%
  \section*{Appendices}
  \begingroup
    \renewcommand{\contentsname}{Appendices}
    \@starttoc{apptoc}%
  \endgroup
}
\makeatother

\newcommand{\cmark}{\textcolor{OliveGreen}{\ding{51}}}%
\newcommand{\xmark}{\textcolor{red}{\ding{55}}}%

\definecolor{darkgreen}{rgb}{0.0, 0.5, 0.0}

\acrodef{dataset}[OrthoLoC]{Orthographic Aerial Localization and Calibration Dataset}
\acrodef{UAV}{Unmanned Aerial Vehicle}
\acrodef{DOP}{Digital Orthophoto}
\acrodef{DSM}{Digital Surface Model}
\acrodef{RTK}{Real-Time Kinematic}
\acrodef{LoD}[LoD]{Level-of-Detail}
\acrodef{SfM}[SfM]{Structure from Motion}
\acrodef{IMU}[IMU]{Inertial Measurement Unit}
\acrodef{GPS}[GPS]{Global Positioning System}
\acrodef{GCP}[GCP]{Ground Control Point}
\acrodef{UTM}[UTM]{Universal Transverse Mercator}
\acrodef{BA}[BA]{Bundle Adjustment}
\acrodef{MVS}[MVS]{Multi-View Stereo}
\acrodef{LiDAR}[LiDAR]{Light Detection and Ranging}
\acrodef{GNSS}[GNSS]{Global Navigation Satellite System}
\acrodef{AdHoP}[AdHoP]{Adaptive Homography Preconditioning}
\acrodef{ME}[ME]{Matching Error}
\acrodef{TE}[TE]{Translation Error}
\acrodef{RE}[RE]{Rotation Error}
\acrodef{RFE}[RFE]{Relative Focal Length Error}
\acrodef{RPE}[RPE]{Reprojection Error}
\acrodef{DLT}[DLT]{Direct Linear Transform}
\acrodef{RMSE}[RMSE]{Root Mean Square Error}

\usepackage{amsmath}

\DeclareMathOperator*{\argmin}{arg\,min}

\colorlet{colorFst}{Green!25}       
\colorlet{colorSnd}{SpringGreen!45} 
\colorlet{colorTrd}{Yellow!30}      
\colorlet{colorLow}{darkgray!30}    
\newcommand{\fs}{\cellcolor{colorFst}}   
\newcommand{\nd}{\cellcolor{colorSnd}}      
\newcommand{\rd}{\cellcolor{colorTrd}}      

\title{OrthoLoC: UAV 6-DoF Localization and Calibration Using Orthographic Geodata}

\newcommand{\sspace}{\hspace{10pt}}
\author{%
  Oussema Dhaouadi$^{1,2,3}$\thanks{Corresponding Author.}
  \sspace
  Riccardo Marin$^{2,3}$
  \sspace
  Johannes Meier$^{1,2,3}$ 
  \\
  \textbf{Jacques Kaiser}$^{1}$
  \sspace
  \textbf{Daniel Cremers}$^{2,3}$
  \\[.8em]
  $^{1}$DeepScenario \sspace $^{2}$TU Munich \sspace $^{3}$Munich Center of Machine Learning \\  
  \texttt{oussema.dhaouadi@tum.de}\\
    \phantom{12}
    \vspace{-0.9cm}
}
\begin{document}

\maketitle

\doparttoc 
\faketableofcontents 

\begin{abstract}
Accurate visual localization from aerial views is a fundamental problem with applications in mapping, large-area inspection, and search-and-rescue operations. In many scenarios, these systems require high-precision localization while operating with limited resources (e.g., no internet connection or GNSS/GPS support), making large image databases or heavy 3D models impractical. Surprisingly, little attention has been given to leveraging orthographic geodata as an alternative paradigm, which is lightweight and increasingly available through free releases by governmental authorities (e.g., the European Union). To fill this gap, we propose OrthoLoC, the first large-scale dataset comprising 16,425 UAV images from Germany and the United States with multiple modalities. The dataset addresses domain shifts between UAV imagery and geospatial data. Its paired structure enables fair benchmarking of existing solutions by decoupling image retrieval from feature matching, allowing isolated evaluation of localization and calibration performance. Through comprehensive evaluation, we examine the impact of domain shifts, data resolutions, and covisibility on localization accuracy. Finally, we introduce a refinement technique called AdHoP, which can be integrated with any feature matcher, improving matching by up to 95\% and reducing translation error by up to 63\%. The dataset and code are available at: \href{https://deepscenario.github.io/OrthoLoC}{https://deepscenario.github.io/OrthoLoC}.
\end{abstract}
\section{Introduction}

Visual localization for \acp{UAV} is essential for digital-twin modeling~\cite{digital_twin_1, digital_twin_2}, surveillance~\cite{wildlife_monitoring_16}, search-and-rescue~\cite{rescue_15}, and infrastructure inspection~\cite{inspection_18}, yet faces unique challenges not addressed by ground-level localization systems. While ground-level approaches~\cite{spartial_aware_19, cross_view_21, sensloc_23} benefit from similar viewpoints between images~\cite{inloc_18, superglue_20, loftr_21}, aerial applications encounter dramatic perspective differences and require scalability over large areas~\cite{render_and_compare_23, anyvisloc_25}.

Current \ac{UAV} localization algorithms rely on retrieving the closest match from a database of posed images~\cite{anyvisloc_25, university_1652_20}, which is inaccurate, or on 3D models of the scene~\cite{render_and_compare_23, uavd4l_24}, which are memory and computationally expensive. In limited resources settings, as it is often the case for connectivity-limited environments, this can result in accuracy degradation. Recent approaches like LoDLoc~\cite{lodloc_24} improve storage efficiency by using~\ac{LoD} but still assume unchanged environments, perform poorly in building-sparse areas such as highways, and its initialization depends on positioning sensors.

\begin{figure}[t]
    \centering
    \includegraphics[width=1\linewidth]{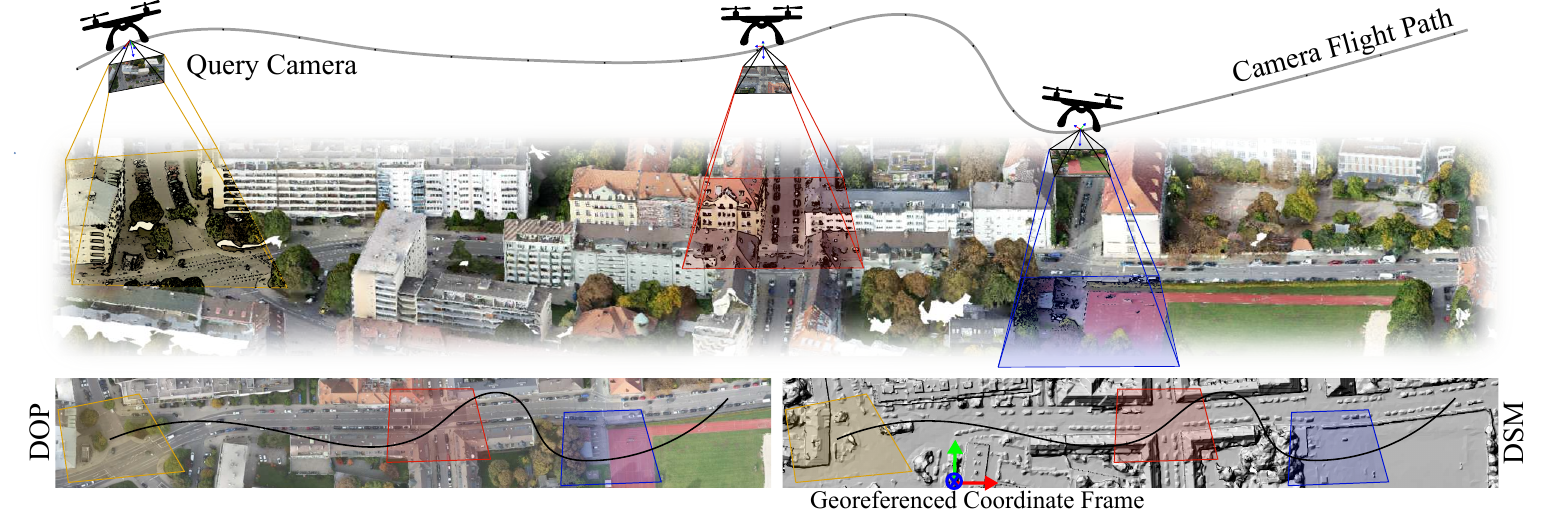}
   \caption{\textbf{Georeferenced UAV Localization / Calibration with Orthographic Geodata.} Our framework bridges the aerial-to-orthographic domain gap. It enables precise 6-DoF localization and calibration using only \acs{DOP} and \acs{DSM} geodata. This approach works even in \acs{GNSS}-denied environments without requiring expensive 3D models or image databases.}
    \label{fig:teaser}
    \vspace{-1.5\baselineskip}
\end{figure}

In contrast, a compelling solution involves geodata, such as orthographic aerial views (\acp{DOP}) and elevation maps (\acp{DSM}). These provide a reliable, lightweight source for localizing \ac{UAV} images, as shown in~\Cref{fig:teaser}. Such data is increasingly accessible through free releases from European government geoportals~\cite{eu_19, eu_23}, and where public access is limited, can be synthesized using photogrammetric tools~\cite{dji_terra}. Geodata are scalable and better suited for low-resource settings. For example, covering an area of approximately 0.265 km² would require a 3D model of around 8 GB~\cite{render_and_compare_23}, whereas geodata requires about 30 times less memory. Surprisingly, no existing \ac{UAV} localization approach seems to fully leverage these data sources. We believe this is mainly due to the absence of aligned cross-domain datasets and the lack of full-pose paired large-scale benchmarks specifically designed for localization using these types of geodata.

To fill this gap, we capture and release the \acf{dataset}. It comprises 5 main modalities such as \ac{UAV} imagery, \acp{DOP}, \acp{DSM}, 3D point maps, and 3D meshes with a total of 16.4K images captured in 47 regions in 19 cities across 2 countries. Our dataset is the first to offer three key advantages: (1) paired UAV-geodata structure that decouples pose estimation from image retrieval, eliminating confounding error sources in the evaluations; (2) precise 6-DoF poses obtained through multi-view georeferenced photogrammetric reconstruction; and (3) additional reference data sources to increase the domain gaps in the dataset.

We have evaluated state-of-the-art methods on this novel localization and calibration task in a comprehensive benchmark. Additionally, we introduce a method-agnostic refinement technique called \textit{\acf{AdHoP}} that further improves localization and calibration accuracy. The technique exploits the uniform structure of \acp{DOP} to perform homography-based warping by assuming quasi-planar surfaces common in built environments.

Our evaluation reveals several insights. First, state-of-the-art matching algorithms can generalize to aerial perspectives but struggle with the substantial domain gap between perspective \ac{UAV} imagery and orthographic reference data. Second, our \textit{AdHoP} technique significantly reduces the perspective disparity, improving all metrics across the tested methods, particularly achieving up to 95\% and 63\% enhancements in matching and translation accuracy, respectively. Third, camera calibration in aerial settings presents unique challenges due to fundamental geometric ambiguities that affect parameters estimation. Finally, reference data characteristics including domain shifts, data resolutions, and covisibility. significantly impact localization performance, with higher resolution geodata providing improvement in accuracy.

The main contributions of this paper are: (1) \ac{dataset}, the first \ac{UAV} dataset providing alignment with geodata across multiple modalities and locations; (2) a unified benchmarking framework for \ac{UAV} localization and calibration that integrates with state-of-the-art matching algorithms and includes our \ac{AdHoP} technique for addressing perspective disparities; and (3) benchmarking results for camera localization and calibration and an analysis of performance factors including cross-domain challenges, data resolution effects, and covisibility.

\section{Related Work}
\subsection{UAV Localization Datasets}

The advancement of \ac{UAV} localization research has been hampered by dataset limitations. Most existing collections fail to support comprehensive 6-DoF evaluation due to several shortcomings. Datasets such as University-1652~\cite{university_1652_20} and DenseUAV~\cite{vision_uav_self_pos_23} provide only \textit{partial pose information} (typically 2-DoF or 3-DoF), insufficient for applications requiring complete 6-DoF estimation. Collections derived from Google Earth~\cite{vision_based_abs_loc_14, clkn_17, vpair_22} predominantly feature nadir views, exhibiting \textit{limited viewpoint diversity} that fails to capture oblique perspectives common in practical \ac{UAV} operations. Several datasets incorporate \textit{synthetic data}---either entirely synthetic environments~\cite{matrix_city_23, gta_uav_25} or synthetically rendered views~\cite{crossloc_22, uavd4l_24}---introducing domain gaps that affect generalization to real-world scenarios. 

Most importantly, existing datasets lack \textit{integrated geodata resources} crucial for evaluating localization methods leveraging lightweight orthographic representations. While the concurrent AnyVisLoc~\cite{anyvisloc_25} dataset includes orthographic geodata, its primary pose evaluation focus is on 3-DoF rather than full 6-DoF. Additionally, it presents a misalignment between its low-resolution satellite imagery and aerial photogrammetry data, which compromises effective evaluation of cross-domain geodata-based localization. We illustrate this misalignment in the supplementary material.

In contrast, \ac{dataset} provides complete 6-DoF ground-truth poses with calibration information, diverse viewpoints across multiple altitudes and angles, real-world imagery from different geographic environments, carefully aligned high-resolution geodata, and paired structure that facilitates isolated evaluation of localization algorithms, independent of retrieval errors. This comprehensive design establishes a foundation for decoupled evaluation of \ac{UAV} localization and calibration methods using lightweight orthographic references, filling a critical research gap.

\subsection{Visual Localization}
\paragraph{Image retrieval-based localization.} 
Image retrieval methods~\cite{dislocation_15, netvlad_2016, large_scale_location_recog_16, place_recog_24_7_15} use global descriptors to match query images against geo-tagged databases. CNN-based approaches such as NetVLAD~\cite{netvlad_2016} and Dislocation~\cite{dislocation_15} are efficient but struggle with large viewpoint and illumination changes in \ac{UAV} imagery. Recent works~\cite{self_supervised_fine_grained_20, robust_img_retrieval_20, end2end_learning_img_retrieval_17} mitigate these issues through view synthesis and self-supervised learning, yet performance drops under extreme perspective shifts. Chen et al.~\cite{chen2025visual} introduced ComplexUAV, a high-resolution UAV dataset covering diverse terrains, along with a contrastive learning framework that improves retrieval robustness and generalization. Nonetheless, retrieval-based methods remain insufficient for accurate 6-DoF UAV localization, motivating alternatives that leverage geodata directly.

\paragraph{Matching-based localization.} Structure-based methods typically build a 3D model using \ac{SfM} techniques~\cite{colmap_16} and establish 2D–3D correspondences, either using mesh models~\cite{landscapear_20, ref_pose_gen_21, meshloc_22, render_and_compare_23} or dense depth maps~\cite{uavd4l_24}. Pose estimation is then performed via PnP algorithms~\cite{review_pnp_94, complete_sol_p3p_03, accurate_non_iter_pnp_07, novel_param_p3p_11} coupled with RANSAC optimization~\cite{lo_ransac_03, optimal_rand_ransac_08, expert_sample_consensus_19, neural_ransac_19, magsac_19}.
Recent advances in feature matching have produced three primary categories of matchers: dense matchers (e.g., DKM~\cite{dkm_23}, ROMA~\cite{roma_24}), semi-dense matchers (e.g., LoFTR~\cite{loftr_21}, eLoFTR~\cite{eloftr_24}, XoFTR~\cite{xoftr_24}), and sparse matchers (e.g., SuperGlue~\cite{superglue_20}, DeDoDe~\cite{dedode_24}, XFeat~\cite{xfeat_24}). Geometry-aware techniques such as MASt3R~\cite{mast3r_24} and DUSt3R~\cite{dust3r_24} further improve matching by integrating geometric constraints. While these state-of-the-art matchers provide robust performance across many scenarios, their effectiveness with orthographic geodata remains unexplored until now.

\paragraph{\ac{UAV}-specific localization.}
Aerial vehicle positioning systems have evolved from 2-DoF to 6-DoF approaches to address specific challenges. Early CNN-based techniques employed multiscale block attention~\cite{faster_effective_cross_view_uav_21} and capsule networks~\cite{geo_semantic_21}, while recent transformer-based frameworks integrate semantic guidance~\cite{semantic_guidance_transformer_22} and relation-aware global attention~\cite{vision_uav_self_pos_23, cross_view_geo_loc_23, geoformer_24} to address scale variations and urban uncertainties. However, most of these methods target only 2-DoF or 3-DoF localization rather than full 6-DoF pose estimation required for advanced applications.

For extended pose estimation, several approaches have emerged with increasing degrees of freedom. For 4-DoF estimation, methods align \ac{UAV} observations with rendered or autoencoded satellite imagery~\cite{vis_loc_google_earth_20, uav_loc_autoenc_21}. 5-DoF methods employ dual Siamese networks with visual odometry and Kalman filtering~\cite{uav_pose_estim_19}. Recent 6-DoF frameworks leverage curriculum learning~\cite{curriculumloc_24}, viewpoint-robust feature extraction~\cite{oblique_robust_abs_vis_loc_23}, attention-based architectures~\cite{absolute_pose_estim_24}, visibility-aware registration~\cite{real_time_geo_loc_21}, and photorealistic synthetic data~\cite{uavd4l_24, crossloc_22}.
While all these methods depend on complete 3D models that require extensive manual effort to create, our approach utilizes widely accessible geodata for 6-DoF pose estimation and camera calibration. Benchmarking results demonstrate accurate localization despite temporal gaps between geodata acquisition and \ac{UAV} flight. This simplifies deployment by utilizing standardized, government-provided resources rather than requiring custom 3D reconstruction for each operational~area.
\section{The \acs{dataset} Dataset}
\begin{figure}[t]
    \centering
    \includegraphics[width=1\linewidth]{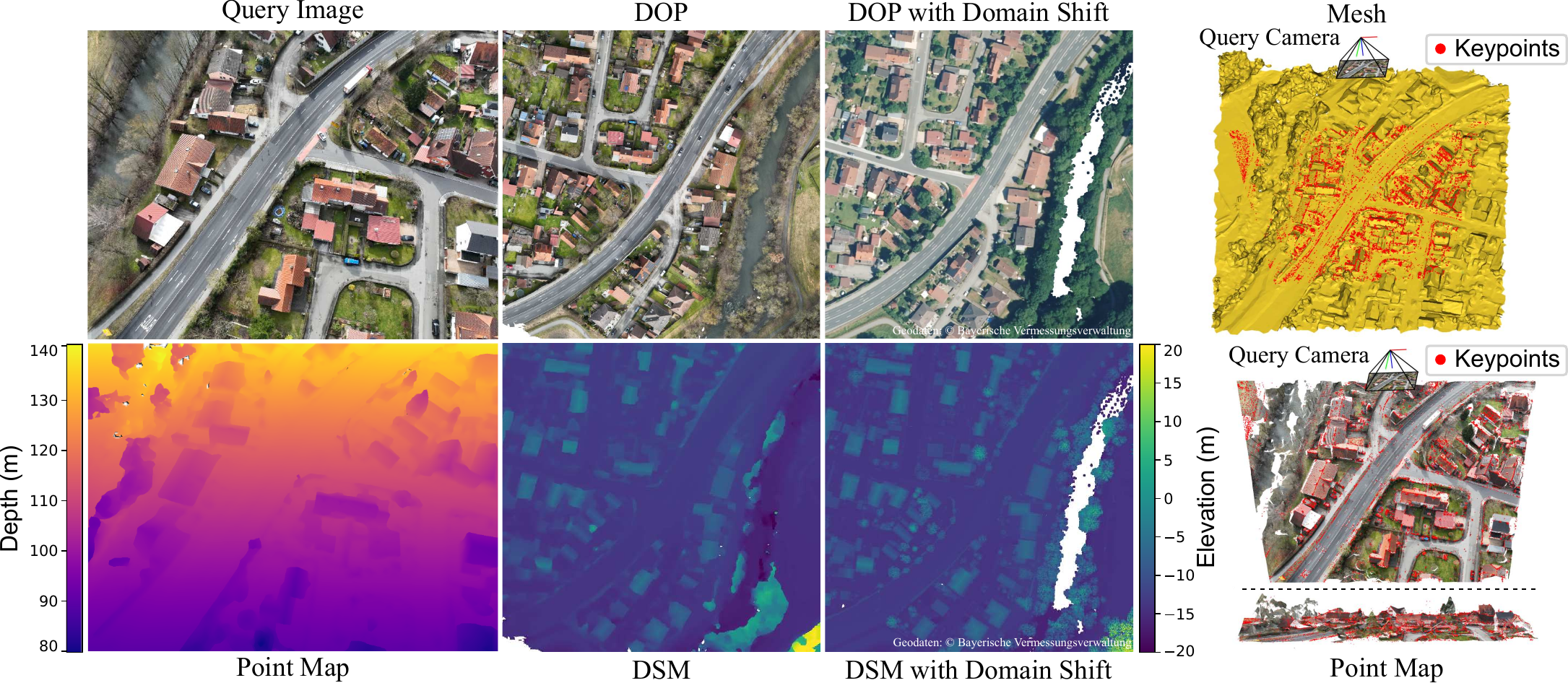}
    \caption{\textbf{Data Modalities in \ac{dataset}.} Each sample includes a query image, a point map (represented as a depth map), a local mesh, visible 3D keypoints, and photogrammetrically reconstructed \ac{DOP}/\ac{DSM}. The dataset also includes an augmented version of \ac{DOP}/\ac{DSM} derived from secondary sources, introducing domain gaps for increased variability.}
    \label{fig:dataset_sample}
    \vspace{-1.0\baselineskip}
\end{figure}

We introduce a comprehensive \ac{UAV} localization dataset that addresses key limitations in existing benchmarks. Our dataset comprises 16.4k real \ac{UAV} images spanning 47 locations across 19 cities in Germany and the United States, captured in diverse environmental contexts including urban, suburban, rural, and highway scenes. Each sample provides a query image with precise ground-truth 6-DoF pose, camera intrinsics, and rich 3D scene representations: point maps, 3D keypoints, local meshes, and aligned 2.5D geodata rasters derived from multiple sources. \Cref{fig:dataset_sample} illustrates the data modalities in our dataset. \Cref{fig:dataset_pipeline} presents the complete creation pipeline. Dataset details are provided in the supplementary material.

\subsection{Data Acquisition and Processing}

Data collection employed commercial drones equipped with \ac{GPS}. For each location, we performed 3D scene reconstruction using \ac{SfM} and \ac{MVS} techniques to generate camera poses, dense point clouds, and textured meshes. The reconstructions were georeferenced using \ac{RTK} measurements or manually annotated \acp{GCP} to ensure precise spatial alignment.

From these reconstructions, we generated orthographic \acp{DOP} via camera renderings and \acp{DSM} through rasterization at 5\,cm/pixel resolution. We complemented these with SIFT~\cite{sift_2004} keypoints extracted from the \ac{DOP} and lifted to 3D using corresponding \ac{DSM} elevations, providing reliable landmarks for pose verification.

\begin{figure}[t]
    \centering
    \includegraphics[width=\linewidth]{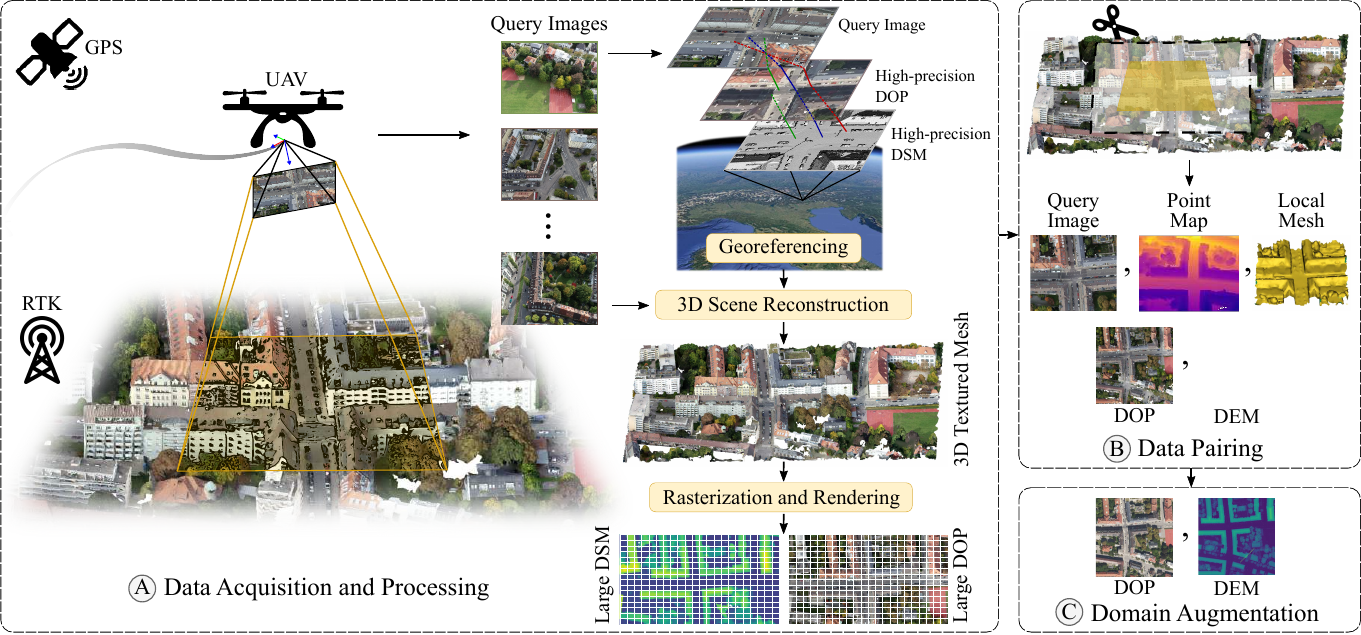}
    \caption{\textbf{Dataset Creation Pipeline.} First, (A)~data acquisition involves \ac{UAV} imagery collection. This data, combined with georeferencing techniques like \acp{GCP} and \ac{RTK}, reconstructs a georeferenced 3D textured mesh. Subsequently, geodata is derived through rasterization and orthographic rendering. Then, (B)~data pairing identifies regions of interest for each query image via raycasting. These areas undergo random expansion, followed by cropping geometric elements to form samples. Finally, (C)~the data is augmented with geodata from external sources, where spatial alignment is verified.}
    \vspace{-1\baselineskip}
    \label{fig:dataset_pipeline}
\end{figure}

\subsection{Data Pairing}
To recover the pose and intrinsic parameters, visual localization methods often require solving image retrieval before running the proper estimation algorithm. These two steps are coupled, making it difficult to disentangle the contribution of each component. Hence, we pair the query with reference data to isolate the contribution of different components and evaluate localization algorithms independent of retrieval performance.

To achieve this, we establish precise correspondences by ray-tracing from each query image with known camera parameters onto the 3D mesh model and exact cropping regions in the \ac{DOP} and \ac{DSM} that geometrically align with the query viewpoint. To quantify how positional uncertainty affects localization accuracy, we extend the reference area beyond the visible query region through spatial perturbations by applying random offsets of 0-10 meters that simulate realistic retrieval imprecision.

In summary, a dataset sample consists of the tuple $(I, \mathbf{P}, \mathbf{R}^{\text{DOP}}, \mathbf{R}^{\text{DSM}}, \mathbf{K}, \mathbf{T}, \mathcal{V}, \mathcal{F}, \mathcal{S})$, where $I \in \mathbb{R}^{H \times W \times 3}$ is the UAV image, $\mathbf{P} \in \mathbb{R}^{H \times W \times 3}$ is the point map, $\mathbf{R}^{\text{DOP}} \in \mathbb{R}^{H^{\text{DOP}} \times W^{\text{DOP}} \times 3}$ is the orthophoto raster, $\mathbf{R}^{\text{DSM}} \in \mathbb{R}^{H^{\text{DSM}} \times W^{\text{DSM}}}$ is the elevation raster, $\mathbf{K} \in \mathbb{R}^{3 \times 3}$ is the camera intrinsic matrix, $\mathbf{T} \in SE(3)$ is the camera pose, $\mathcal{V} \in \mathbb{R}^{N \times 3}$ represents the mesh vertices, $\mathcal{F} \in \mathbb{N}^{M \times 3}$ defines the mesh faces, and $\mathcal{S}$ is the set of 3D keypoints. All geometric elements are transformed into a local coordinate system to preserve privacy while maintaining precise geometric relationships.

\subsection{Domain Augmentation}
Solving \ac{UAV} localization requires robustness to natural changes in scenes due to time passing. Typically, reference geodata may have been collected months or years before a \ac{UAV} flight, creating significant domain gaps that cannot be easily addressed through simple data augmentation or domain adaptation techniques. These gaps are particularly challenging because they involve both appearance and structural changes that vary unpredictably across locations and seasons.

We can divide these challenges into two categories: (1) visual domain gaps in \acp{DOP} through appearance changes (color shifts, illumination variations, seasonal differences) while maintaining structural consistency; and (2) structural domain gaps in \acp{DSM} through geometric modifications (construction changes, vegetation growth, infrastructure evolution).

Including real-world domain gaps in our dataset is essential because synthetic alternatives cannot replicate the complex natural variations occurring over time. Our dataset provides three sample categories: minimal to no domain gap (i.e., same-domain) samples that include geodata from the 3D reconstruction, visual domain gaps only (i.e., cross-domain \ac{DOP}), and both visual and structural disparities (i.e., cross-domain \ac{DOP} and \ac{DSM}). Cross-domain samples were created by incorporating open geodata from European locations and visually verifying alignment with same-domain samples.

\begin{table}[t]
  \centering
  \setlength{\tabcolsep}{2.3pt}
  \caption{\textbf{Comparison of Existing \ac{UAV} Localization Datasets.}\newline
  \footnotesize \textbf{Legend}: 
  \textit{Country codes}: Switzerland~(CH), China~(CN), United States~(US), Germany~(DE);
  \textit{Geographic}: Urban~(U), Suburban~(SU), Rural~(R), Campus~(C), Highway~(H); 
  \textit{\ac{UAV} images}: Real~(Re), Synthetic~(Sy); 
  \textit{View}: top-down~(nadir), angled~(oblique), mixed views~(both);
  \textit{Altitude}: $\leq$150\,m~(low), $>$150\,m~(high), mixed altitudes~(both);
  \textit{3D}: Depth~(D), Point Map~(PM), Level of Detail~(LoD);
  \textit{Task}: Image Retrieval~(IR);
  \textit{Platform}: + indicates georeferencing techniques (\ac{RTK}, \ac{GCP});
  \textit{XD}: cross-domain (reference data are from external sources).
  }
  \resizebox{\linewidth}{!}{
  \renewcommand{\arraystretch}{1.1}
  \begin{tabular}{l|ccc|ccccc|cccc|c}
  \Xhline{1pt}
  \rowcolor{gray!15}
   & \multicolumn{3}{c|}{\textbf{Geographic Coverage}} & \multicolumn{5}{c|}{\textbf{UAV Data}} & \multicolumn{4}{c|}{\textbf{Reference Data}} & \\
  \rowcolor{gray!15}
  \textbf{Dataset} & \textbf{Country} & \textbf{Scene} & \textbf{\#Loc} & \textbf{Imgs (Re+Sy)} & \textbf{View} & \textbf{Alt} & \textbf{3D} & \textbf{Platform} & \textbf{Amount} & \textbf{Type} & \textbf{3D} & \textbf{XD} & \textbf{Task} \\
  \Xhline{1pt}
  \multicolumn{14}{c}{\cellcolor{gray!10}\textbf{Unpaired}} \\
  \hline
  MatrixCity~${\cite{matrix_city_23}}_{2023}$ & - & U & 1 & 0+519k & oblique & low & D & virtual & \xmark & \xmark & \xmark & \xmark & 6-DoF\\
  \hline
  CrossLoc~${\cite{crossloc_22}}_{2022}$ & CH & U & 2 & 4.5k+19.5k & both & low & D/PM & drone+ & 1 & \xmark & \xmark & \xmark & 6-DoF \\
  \hline
  AirLoc~${\cite{render_and_compare_23}}_{2023}$ & CN & U & 1 & 2.7k+0 & both & low & \xmark & drone+ & \xmark & \xmark & Mesh & \xmark & 6-DoF \\
  \hline
  UAVD4L~${\cite{uavd4l_24}}_{2024}$ & CN & U & 2 & 0.9k+18k & both & low & D & drone+ & 1 & \xmark & Mesh/DSM & \xmark & 6-DoF \\
  \hline
  Swiss-EPFL~${\cite{lodloc_24}}_{2024}$ & CH & U & 2 & 2.2k+14.7k & both & low & \xmark & drone+ & 2 & \xmark & LoD & \xmark & 6-DoF \\
  \hline
  UAVD4L-LoD~${\cite{lodloc_24}}_{2024}$ & CN & U & 2 & 3.7k+18k & both & low & \xmark & drone+ & 1 & \xmark & LoD & \xmark & 6-DoF \\
  \hline
  UAV-VisLoc~${\cite{uav_visloc_24}}_{2024}$ & CN & U & 11 & 6.7k+0 & nadir & high & \xmark & drone+ & 11 & DOP & \xmark & \xmark & IR \\
  \hline
  GTA-UAV~${\cite{gta_uav_25}}_{2025}$ & - & U & 1 & 0+33k & nadir & both & \xmark & virtual & \xmark & \xmark & \xmark & \xmark & 6-DoF \\
  \hline
  AnyVisLoc~${\cite{anyvisloc_25}}_{2025}$ & CN & U,R,SU & 25 & 18k+0 & both & both & \xmark & drone+ & 25 & DOP & DSM & \cmark & 3-DoF \\
  \hline
  \multicolumn{14}{c}{\cellcolor{gray!10}\textbf{Paired}} \\
  \hline
  University-1652~${\cite{university_1652_20}}_{2020}$ & US & C & 39 & 701+50.2k & oblique & both & \xmark & web & 951 & Images & \xmark & \cmark & IR \\
  \hline
  DenseUAV~${\cite{vision_uav_self_pos_23}}_{2023}$ & CN & C & 14 & 9k+0 & nadir & low & \xmark & drone & 18k & Images & \xmark & \cmark & 3-DoF \\
  \hline
  SUES-200~${\cite{sues_200_23}}_{2023}$ & CN & U & 200 & 40k+0 & both & high & \xmark & drone & 200 & DOP & \xmark & \cmark & IR \\
  \hline
  ALTO~${\cite{alto_22}}_{2022}$ & US & U,R,SU & 1 & 15.4k+0 & nadir & high & \xmark & aircraft+ & 16.5k & DOP & LiDAR & \xmark & 6-DoF \\
  \hline
  VPAIR~${\cite{vpair_22}}_{2022}$ & DE & U,R,SU & 1 & 2.7k+0 & nadir & high & \xmark & aircraft+ & 2.7k & Images & Depth & \cmark & 6-DoF \\
  \hline
  \textbf{\ac{dataset} (Ours)} & US,DE & U,R,SU,H & 47 & 16.4k+0 & both & both & PM & drone+ & 16.4k & DOP & DSM & \cmark & 6-DoF \\
  \Xhline{1pt}
  \end{tabular}
  }
  \label{tab:datasets_overview}
  \vspace{-1\baselineskip}
\end{table}

\subsection{Comparison with Existing Datasets}
\ac{dataset} presents the first \ac{UAV} localization dataset for 6-DoF pose estimation using governmental geodata (\acp{DOP} and \acp{DSM}) as the only reference. This eliminates costly posed image databases, meshes, or point clouds, enabling real-time localization without preprocessing.

Our dataset spans 47 locations across 2 countries with 16.4\,k real \ac{UAV} images, paired multi-modal data (\acp{DOP}, \acp{DSM}, and 3D reconstructions), diverse viewpoints from nadir to oblique perspectives, and high-precision ground-truth achieving approximately 5\,cm median error via \acp{GCP} evaluation. Over 4\,k governmental orthoimages and surface models enable robust domain adaptation assessment. \ac{dataset} uniquely provides aligned \ac{DOP}+\ac{DSM} pairs with accurate 6-DoF poses across multiple altitudes and privacy-preserving georeferencing decoupling.

As shown in~\Cref{tab:datasets_overview}, existing datasets suffer from (1) restricted geographic coverage~\cite{crossloc_22, vpair_22, alto_22, matrix_city_23, render_and_compare_23, uavd4l_24, lodloc_24}, (2) synthetic data dependency~\cite{university_1652_20, matrix_city_23, gta_uav_25, crossloc_22, lodloc_24}, or (3) incomplete pose information~\cite{university_1652_20, sues_200_23, vision_uav_self_pos_23, uav_visloc_24}. Our geometric consistency analysis reveals significant projection errors in CrossLoc~\cite{crossloc_22}, UAVD4L~\cite{uavd4l_24}, and AnyVisLoc~\cite{anyvisloc_25}, which provides only 3-DoF poses with misaligned reference data. Assessment details are in the supplementary material.
\section{Localization with Orthographic Geodata}

Unlike traditional approaches that rely on image retrieval or 3D models, we explore the novel paradigm of \ac{UAV} localization using 2.5D orthographic geodata. No existing methods are directly applicable to this scenario, as previous work has not leveraged the combination of \ac{DOP}s and \ac{DSM}s for \ac{UAV} pose estimation. This section presents the problem formulation, our benchmarking framework, and our refinement technique.

\subsection{Problem Formulation}

\paragraph{Goal.} Given an orthophoto raster $\mathbf{R}^{\text{DOP}}$, an elevation raster $\mathbf{R}^{\text{DSM}}$, and a query \ac{UAV} image $I$ taken from an arbitrary viewpoint, we aim to \textit{determine the georeferenced 6-DoF pose $\mathbf{T}$ of the camera (localization) and, optionally, its intrinsic parameters $\mathbf{K}$ (calibration)}.

\paragraph{Challenges.} The key challenge is bridging two fundamentally different projection models: perspective projection for \ac{UAV} imagery and orthographic projection for geodata. This difference creates a domain gap that is particularly pronounced in oblique views where perspective distortion is significant. Additionally, another domain gap arises from the visual and structural discrepancies between the query and reference data caused by differences in acquisition time. We provide the mathematical principles for both projection types, with particular focus on deriving a formulation for nadir orthographic projection in the supplementary material.

\begin{figure}[t]
    \centering
    \includegraphics[width=1.0\linewidth]{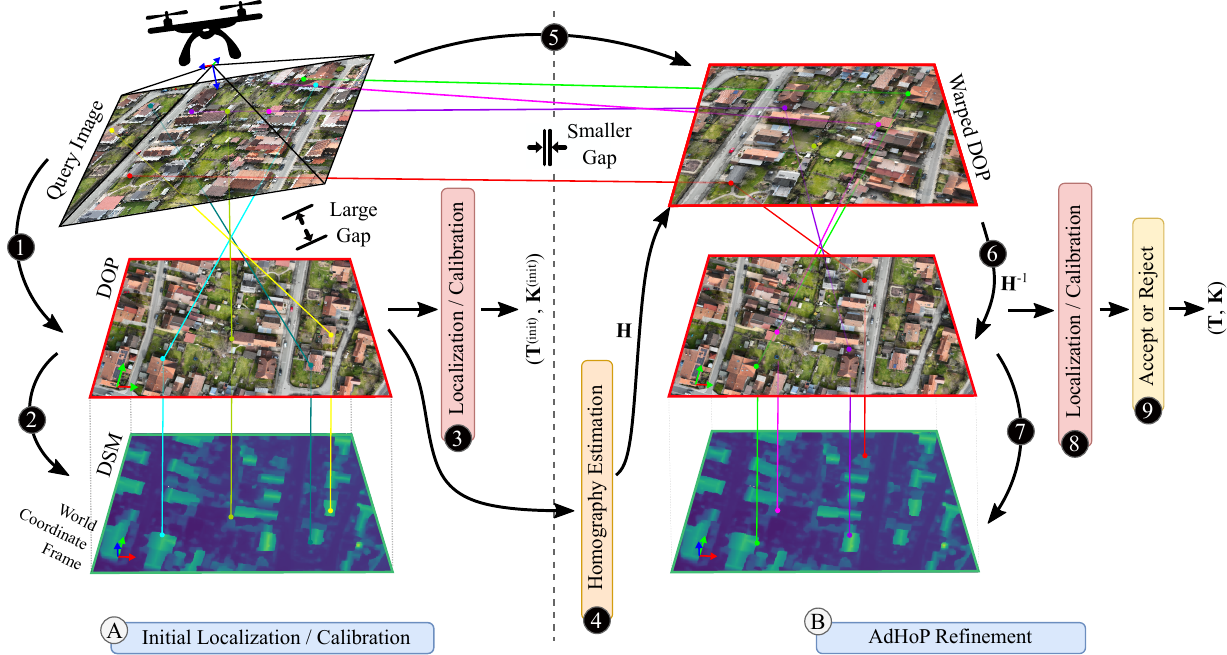}
  \caption{\textbf{UAV 6-DoF Localization and Calibration with \ac{AdHoP}:} 
(A) \textit{Initial Localization / Calibration:} We match features between the query image and \ac{DOP} (1), lift the correspondences to 3D using the \ac{DSM} (2), and compute an initial pose and optional intrinsics (3). 
(B) \textit{\ac{AdHoP} Refinement}: Using the initial 2D-2D correspondences, we estimate a homography to warp the \ac{DOP} (4), thereby reducing perspective differences. This enables enhanced feature matching on the warped orthophoto~(5). The new correspondences are then mapped back to the original unwarped coordinate space (6), lifted to 3D using the \ac{DSM} (7), and used to compute refined camera parameters (8). The refinement is accepted only when it reduces the reprojection error (9).}
    \label{fig:method_pipeline}
    \vspace{-1\baselineskip}
\end{figure}

\paragraph{Benchmarking framework.}
Given the absence of existing methods that directly tackle \ac{UAV} localization using orthographic geodata, we propose a comprehensive benchmarking framework to evaluate various combinations of matching algorithms as backbones. Our framework is entirely backbone-agnostic, enabling integration with any feature matching method, as illustrated in~\Cref{fig:method_pipeline} and detailed in the following subsections.

\subsection{Initial Camera Calibration / Localization}
\label{sec:initial_cam_cal_loc}
We establish 2D-2D correspondences between the query image $I$ and the orthophoto $\mathbf{R}^{\text{DOP}}$ using state-of-the-art matching methods such as GIM+DKM~\cite{gim_24}, RoMA~\cite{roma_24}, SuperGlue~\cite{superglue_20}, and LoFTR~\cite{loftr_21}. Each 2D point matched in $\mathbf{R}^{\text{DOP}}$ is lifted to a 3D point using the corresponding elevation value from $\mathbf{R}^{\text{DSM}}$, providing the necessary 3D-2D correspondences for pose estimation (details in the supplementary material).
Next, we filter correspondences by excluding matches with low confidence scores (below 0.5), invalid 3D points (missing data values), and points outside the field of view. Our calibration approach employs a two-stage optimization strategy. In the first stage, we use an initial guess of the focal length to estimate the camera pose by optimizing reprojection errors using RANSAC-EPnP~\cite{epnp} and a 5-pixel inlier threshold. In the second stage, we use this pose to initialize a Levenberg-Marquardt optimization that jointly refines camera intrinsics and extrinsics. For pure localization tasks, we only perform the first stage, as intrinsics are assumed to be known.

\subsection{\acs{AdHoP} Refinement}
\label{sec:adhop}
Perspective differences between query and reference images are a major challenge in \ac{UAV} localization, especially for oblique viewpoints. Our geodata-based approach addresses this with \acf{AdHoP}, a method-agnostic refinement technique that exploits the approximate planarity of many aerial elements (roads, building roofs, fields). Formally, \ac{AdHoP} estimates a homography matrix $\mathbf{H} \in \mathbb{R}^{3 \times 3}$ from initial 2D--2D correspondences using normalized \ac{DLT} with RANSAC. We adopt this straightforward formulation to avoid the complexity and biases of learning-based methods, requiring no training, dataset dependencies, or ad-hoc domain assumptions while providing a transparent and general baseline. The homography warps the orthophoto to better match the query perspective, enabling a second round of feature matching with improved similarity. The new matches are mapped back using $\mathbf{H}^{-1}$, lifted to 3D via the \ac{DSM}, and used to refine pose estimation. The refinement is accepted only if it reduces mean reprojection error.

In our experiments, we demonstrate that combining different matching algorithms with \ac{AdHoP} significantly improves localization and calibration accuracy, with GIM+DKM+AdHoP emerging as the most effective combination across diverse scenarios. This approach highlights the practical advantages of integrating geodata into localization pipelines.
\section{Experimental Results}

In this section, we introduce the evaluation metrics (\Cref{sec:evaluation_metrics}) and present our benchmarking results. We evaluate both localization (\Cref{sec:localization_results}) and calibration performance (\Cref{sec:calibration_results}) using state-of-the-art feature matchers as backbones. We then analyze some factors affecting performance across different scenarios (\Cref{sec:data_characteristics}). For complete experimental results and additional analyses, please refer to the supplementary material.

\subsection{Evaluation Metrics}
\label{sec:evaluation_metrics}

We report several metrics: \acf{ME} in pixels as the median distance between ground-truth and estimated matching coordinates; \acf{TE} in meters and \acf{RE} in degrees for pose accuracy; \acf{RPE} in pixels for keypoint reprojection errors; recall percentages at thresholds 1m-1°, 3m-3°, and 5m-5°; and \acf{RFE} in percent for calibration accuracy. We also report computation time in seconds.

\subsection{Camera Localization}
\label{sec:localization_results}

\begin{table}[t]
  \centering
   \caption{\textbf{Quantitative Localization Results on \ac{dataset} Test Sets.} Rankings between matchers are highlighted as \colorbox{colorFst}{first}, \colorbox{colorSnd}{second}, and \colorbox{colorTrd}{third}. \textbf{Bold} values indicate the best performance comparing without/with \ac{AdHoP}. RI indicates a rotation-invariant matcher (matching performed with 4 rotated versions, selecting the one with most correspondences). Abbreviations: SuperPoint~(SP), SuperGlue~(SG), LightGlue~(LG), Minima~(MM).}
  \label{tab:loc_w_wo_adhop_results}
  \scriptsize
  \setlength{\tabcolsep}{1pt}
  
\resizebox{\linewidth}{!}{\
\renewcommand{\arraystretch}{1.1}
  \begin{tabular}{c!{\vrule width 1pt}c
                  !{\vrule width 1pt}c|ccc|ccc|c
                  }
    \Xhline{1pt} \rowcolor{gray!15}
      \multirow{1}{*}{\textbf{Matcher}}
      & \multirow{1}{*}{\textbf{RI}} 
      & \textbf{ME} [px]$\downarrow$ & \textbf{TE} [m]$\downarrow$ & \textbf{RE} [°]$\downarrow$ & \textbf{RPE} [px]$\downarrow$ & \textbf{1m-1°} [\%]$\uparrow$ & \textbf{3m-3°} [\%]$\uparrow$ & \textbf{5m-5°} [\%]$\uparrow$ & \textbf{Speed} [s]$\downarrow$ \\
    \Xhline{1pt}

    SP+SG~\cite{superpoint_18,superglue_20} & \xmark & {\bf 2.2} / {\bf 2.2} & {\fs 0.36}  / {\bf {\nd 0.35} } & {\bf {\nd 0.15} } / {\bf {\nd 0.15} } & {\bf {\fs 2.8} } / {\bf {\nd 2.8} } & {\rd 63.9}  / {\bf {\nd 64.4} } & {\rd 77.4}  / {\bf {\rd 77.6} } & 78.7 / {\bf 78.9} & {\bf {\nd 0.2} } / {\nd 0.3}  \\ 
SP+LG~\cite{superpoint_18,lightglue_23} & \xmark & {\bf {\rd 2.0} } / {\bf {\rd 2.0} } & {\bf {\nd 0.37} } / {\bf {\rd 0.37} } & {\rd 0.16}  / {\bf {\nd 0.15} } & {\bf {\nd 2.9} } / {\bf {\rd 2.9} } & {\nd 64.0}  / {\bf {\rd 64.2} } & 77.0 / {\bf 77.4} & {\rd 78.8}  / {\bf {\rd 79.0} } & {\bf {\fs 0.1} } / {\fs 0.2}  \\ 
DeDoDe~\cite{dedode_24} & \xmark & {\bf {\fs 1.2} } / {\bf {\fs 1.2} } & 0.42 / {\bf 0.39} & 0.18 / {\bf {\rd 0.16} } & 3.6 / {\bf 3.2} & 27.5 / {\bf 28.2} & 33.3 / {\bf 33.6} & 35.6 / {\bf 35.7} & {\bf {\rd 0.3} } / {\bf {\nd 0.3} } \\ 
XFeat~\cite{xfeat_24} & \xmark & 257.0 / {\bf 38.1} & 1.58 / {\bf 0.96} & 0.74 / {\bf 0.45} & 13.0 / {\bf 7.8} & 42.7 / {\bf 50.8} & 57.4 / {\bf 63.0} & 61.2 / {\bf 65.1} & {\bf {\fs 0.1} } / {\fs 0.2}  \\ 
XFeat+LG~\cite{xfeat_24,lightglue_23} & \xmark & 4.3 / {\bf 3.2} & 0.57 / {\bf 0.48} & 0.25 / {\bf 0.20} & 4.7 / {\bf 3.8} & 42.5 / {\bf 45.7} & 54.4 / {\bf 56.3} & 56.3 / {\bf 57.3} & {\bf {\fs 0.1} } / {\nd 0.3}  \\ 
\Xhline{0.1pt}

LoFTR~\cite{loftr_21} & \xmark & 317.2 / {\bf 312.9} & 121.56 / {\bf 118.77} & 109.49 / {\bf 107.22} & 1451.9 / {\bf 1384.7} & 18.0 / {\bf 21.0} & 23.3 / {\bf 25.6} & 23.9 / {\bf 26.3} & {\bf {\fs 0.1} } / {\fs 0.2}  \\ 
MM+LoFTR~\cite{minima_24,lodloc_24} & \xmark & {\bf 266.9} / 269.1 & 87.17 / {\bf 84.69} & 98.89 / {\bf 97.81} & 902.4 / {\bf 841.4} & 14.5 / {\bf 18.2} & 21.5 / {\bf 23.1} & 22.5 / {\bf 23.7} & {\bf {\rd 0.3} } / {\rd 0.6}  \\ 
eLoFTR~\cite{eloftr_24} & \xmark & 329.5 / {\bf 311.9} & 124.29 / {\bf 117.53} & 109.25 / {\bf 102.50} & 1552.2 / {\bf 1471.1} & 19.0 / {\bf 22.9} & 24.0 / {\bf 27.6} & 24.8 / {\bf 28.4} & {\bf {\fs 0.1} } / {\fs 0.2}  \\ 
XoFTR~\cite{xoftr_24} & \xmark & 291.7 / {\bf 285.9} & 113.65 / {\bf 113.15} & {\bf 107.24} / 107.65 & 1322.4 / {\bf 1275.2} & 19.7 / {\bf 21.5} & 23.8 / {\bf 24.8} & 24.1 / {\bf 25.4} & {\bf {\fs 0.1} } / {\fs 0.2}  \\ 
\Xhline{0.1pt}

DKM~\cite{dkm_23} & \cmark & 8.8 / {\bf 2.7} & 3.83 / {\bf 1.40} & 1.93 / {\bf 0.63} & 31.9 / {\bf 11.9} & 33.6 / {\bf 42.2} & 44.2 / {\bf 49.8} & 45.6 / {\bf 50.4} & {\bf 0.8} / 1.7 \\ 
XFeat*~\cite{xfeat_24} & \xmark & 222.2 / {\bf 9.2} & 1.07 / {\bf 0.66} & 0.48 / {\bf 0.30} & 8.8 / {\bf 5.4} & 48.8 / {\bf 59.8} & 67.3 / {\bf 72.3} & 70.4 / {\bf 73.7} & {\bf {\fs 0.1} } / {\fs 0.2}  \\ 
GIM+DKM~\cite{gim_24,dkm_23} & \cmark & {\nd 1.5}  / {\bf {\nd 1.3} } & {\rd 0.40}  / {\bf {\fs 0.32} } & {\bf {\fs 0.12} } / {\bf {\fs 0.12} } & {\rd 3.1}  / {\bf {\fs 2.6} } & {\fs 74.1}  / {\bf {\fs 75.4} } & {\fs 86.6}  / {\bf {\fs 87.9} } & {\fs 87.4}  / {\bf {\fs 88.4} } & {\bf 1.3} / 2.6 \\ 
DUSt3R~\cite{dust3r_24} & \cmark & 5.0 / {\bf 4.9} & {\bf 3.45} / 3.68 & {\bf 1.47} / 1.53 & {\bf 25.8} / 27.3 & 3.6 / {\bf 6.4} & 33.6 / {\bf 33.8} & {\bf 51.7} / 49.8 & {\bf 1.5} / 2.1 \\ 
MASt3R~\cite{mast3r_24} & \cmark & 2.4 / {\bf 2.3} & 0.61 / {\bf 0.60} & 0.28 / {\bf 0.26} & 5.0 / {\bf 4.8} & 62.4 / {\bf 63.5} & {\nd 81.4}  / {\bf {\nd 82.0} } & {\nd 84.2}  / {\bf {\nd 84.5} } & {\bf 2.2} / 3.4 \\ 
RoMa~\cite{roma_24} & \cmark & 21.6 / {\bf 2.4} & 1.47 / {\bf 0.75} & 0.67 / {\bf 0.32} & 12.5 / {\bf 6.2} & 44.4 / {\bf 54.6} & 56.1 / {\bf 65.1} & 59.2 / {\bf 66.8} & {\bf 1.1} / 2.1 \\ 
MM+RoMa~\cite{minima_24,roma_24} & \cmark & 70.8 / {\bf 4.6} & 3.63 / {\bf 1.21} & 1.92 / {\bf 0.55} & 34.2 / {\bf 9.9} & 38.6 / {\bf 47.9} & 48.9 / {\bf 58.0} & 51.6 / {\bf 59.5} & {\bf 1.1} / 2.1 \\ 
\Xhline{1pt}

  \end{tabular}
  }
  \vspace{-1\baselineskip}
\end{table}

We report localization performance when using state-of-the-art feature matching approaches with and without our \ac{AdHoP} strategy in~\Cref{tab:loc_w_wo_adhop_results}. GIM+DKM~\cite{gim_24, dkm_23} achieves the highest performance across the majority of metrics. SP+SG~\cite{superpoint_18, superglue_20} and SP+LG~\cite{superpoint_18, lightglue_23}, along with GIM+DKM, all achieve precise localization below 40\,cm and 0.16 degrees. However, the sparse matchers (SP-based) have notably lower recall compared to GIM+DKM, indicating they successfully localize fewer images. Semi-dense approaches like LoFTR~\cite{loftr_21} and XoFTR~\cite{xoftr_24} perform poorly, with recall below~19.7\%. Our intuition is that these approaches suffer from limited training datasets or architectural constraints that prevent handling large domain shifts.

\begin{figure}[htpb]
\centering
\includegraphics[width=1.0\textwidth]{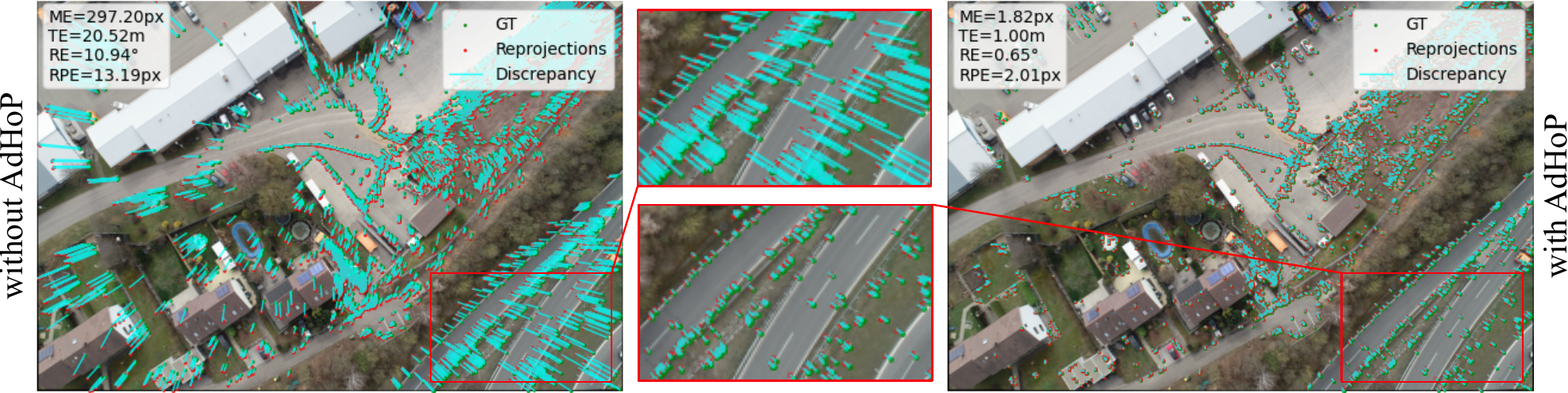}
\caption{\textbf{Localization Without and With \ac{AdHoP}.} xFeat*~\cite{xfeat_24} matching results showing 3D keypoint projections in \textcolor{darkgreen}{green} (using the ground-truth pose) and \textcolor{red}{red} (using the estimated pose). \textcolor{cyan}{Blue} lines indicate projection discrepancies between estimated and ground-truth positions.}
\vspace{-1\baselineskip}
\label{fig:localization_w_wo_adhop}
\end{figure}

Integrating \ac{AdHoP} substantially improves performance across all matchers. We observe an average matching improvement of about 30\%, yielding translational and rotational error reductions of 20\% each. The best-performing GIM+DKM~\cite{gim_24,dkm_23} with \ac{AdHoP} reduces translation error by 20\%, from 0.40\,m to 0.32\,m. Previously underperforming methods show even more dramatic improvements: XFeat*~\cite{xfeat_24} matching error decreases by 95.86\%, DKM~\cite{dkm_23} reduces translation error by 63\%, and RoMa~\cite{roma_24} increases 1m-1° recall by 23\% while reducing translation error by half. We illustrate in~\Cref{fig:localization_w_wo_adhop} the impact of \ac{AdHoP} in reducing errors.

\subsection{Camera Calibration}
\label{sec:calibration_results}

Our calibration experiments reveal a fundamental challenge in estimating camera intrinsics from \ac{UAV} imagery due to geometric ambiguity between focal length and translation estimation. We provide mathematical proof of this ambiguity and detailed calibration benchmarking results in the supplementary material. Despite this challenge, the combination of GIM+DKM~\cite{gim_24,dkm_23} with our \ac{AdHoP} technique achieves the best focal length estimation with just 1.6\% relative error with a translation error of 2.09\,m. However, the recall remains relatively low at 21.8\%, highlighting the inherent difficulty of the calibration task.

\subsection{Performance Factors}
\label{sec:data_characteristics}

\paragraph{Domain shift.}
Using cross-domain \acp{DOP} affects algorithms differently, with varying robustness to appearance changes. Even the best-performing method, GIM+DKM~\cite{gim_24,dkm_23} with \ac{AdHoP}, shows a threefold increase in translational error under these conditions. Further domain shift, combining both cross-domain \acp{DOP} and \acp{DSM}, causes significant degradation across all methods. For instance, GIM+DKM~\cite{gim_24,dkm_23} with \ac{AdHoP} experiences a sevenfold increase in translation error, rising from 0.16\,m to 1.12\,m. These findings highlight the need for localization algorithms robust against both visual and geometric domain shifts. Complete results are provided in the supplementary material.

\paragraph{Resolution and covisibility impact.}
Performance remains robust when scaling images to 30\% of original size ($\sim$~300 pixels, with highest geodata resolution at 13\,m/pixel), with degradation occurring only at lower resolutions. Additionally, localization accuracy depends heavily on the covisibility ratio between query and reference images, dropping significantly when less than 20\% of the elements seen in the query image are visible in the reference data. These findings have important implications for real-world \ac{UAV} deployment, where error-prone upstream tasks like image retrieval may result in extraction of incomplete reference data. A detailed analysis of these factors is in the supplementary.
\section{Conclusion}

We presented a novel paradigm for \ac{UAV} visual localization using widely available geodata. To support this approach, we introduced \ac{dataset}, a diverse large-scale \ac{UAV} localization and calibration dataset spanning multiple environments and regions. Our benchmarking framework matches \ac{UAV} imagery with orthophotos, which are lifted to 3D using 2.5D elevation models to solve pose estimation via PnP. Our proposed \ac{AdHoP} technique consistently enhances various matching algorithms, yielding significant improvements in both pose estimation and camera calibration performance.

Our benchmarking demonstrated that standard 2.5D geodata proved sufficient for accurate 6-DoF pose estimation in outdoor \ac{UAV} localization. Our evaluations revealed that dense matchers, specifically GIM+DKM~\cite{gim_24,dkm_23} with \ac{AdHoP}, achieved 75.4\% recall at 1m-1° threshold, though with limited robustness to domain shifts. Camera calibration performance remained challenging due to inherent geometric ambiguities between focal length and translation estimation. We also demonstrated that higher resolution data significantly improved localization accuracy, confirming that low-resolution reference data (such as satellite imagery) limited performance. Moreover, we found that the area coverage of geodata---typically determined by upstream tasks like region of interest detection through image retrieval---critically affected correspondence distribution and reliable pose estimation.

\textbf{Limitations and future work.} Calibration remains affected by translation and focal length ambiguities, which could be addressed by training end-to-end networks for improved localization. While our framework shows strong performance, it requires geodata with at least 20\% covisibility and does not yet scale to large rasters without region detection, a key factor for real-world deployment. Moreover, \ac{AdHoP} improves partially incorrect correspondences but fails when matches are completely corrupted. This highlights the potential of more advanced homography estimators, particularly learning-based approaches, to reduce reliance on initial matching and increase robustness.

\textbf{Acknowledgement.}
This work is a result of the joint research project STADT:up. The project is supported by the German Federal Ministry for Economic Affairs and Climate Action (BMWK), based on a decision of the German Bundestag. The author is solely responsible for the content of this publication.

{\small
\bibliographystyle{ieee}
\bibliography{bibliography}

\begin{thebibliography}{10}\itemsep=-1pt

\bibitem{netvlad_2016}
R.~Arandjelovic, P.~Gronat, A.~Torii, T.~Pajdla, and J.~Sivic.
\newblock Netvlad: Cnn architecture for weakly supervised place recognition.
\newblock In {\em Proceedings of the IEEE conference on computer vision and
  pattern recognition}, pages 5297--5307, 2016.

\bibitem{dislocation_15}
R.~Arandjelovi{\'c} and A.~Zisserman.
\newblock Dislocation: Scalable descriptor distinctiveness for location
  recognition.
\newblock In {\em Computer Vision--ACCV 2014: 12th Asian Conference on Computer
  Vision, Singapore, Singapore, November 1-5, 2014, Revised Selected Papers,
  Part IV 12}, pages 188--204. Springer, 2015.

\bibitem{magsac_19}
D.~Barath, J.~Matas, and J.~Noskova.
\newblock Magsac: marginalizing sample consensus.
\newblock In {\em Proceedings of the IEEE/CVF conference on computer vision and
  pattern recognition}, pages 10197--10205, 2019.

\bibitem{uav_loc_autoenc_21}
M.~Bianchi and T.~D. Barfoot.
\newblock {UAV} localization using autoencoded satellite images.
\newblock {\em CoRR}, abs/2102.05692, 2021.

\bibitem{expert_sample_consensus_19}
E.~Brachmann and C.~Rother.
\newblock Expert sample consensus applied to camera re-localization.
\newblock In {\em Proceedings of the IEEE/CVF International Conference on
  Computer Vision}, pages 7525--7534, 2019.

\bibitem{neural_ransac_19}
E.~Brachmann and C.~Rother.
\newblock Neural-guided ransac: Learning where to sample model hypotheses.
\newblock In {\em Proceedings of the IEEE/CVF International Conference on
  Computer Vision}, pages 4322--4331, 2019.

\bibitem{landscapear_20}
J.~Brejcha, M.~Luk{\'a}{\v{c}}, Y.~Hold-Geoffroy, O.~Wang, and
  M.~{\v{C}}ad{\'\i}k.
\newblock Landscapear: Large scale outdoor augmented reality by matching
  photographs with terrain models using learned descriptors.
\newblock In {\em European Conference on Computer Vision}, pages 295--312.
  Springer, 2020.

\bibitem{openmvs_20}
D.~Cernea.
\newblock {OpenMVS}: Multi-view stereo reconstruction library.
\newblock \url{https://github.com/cdcseacave/openMVS}, 2020.

\bibitem{clkn_17}
C.-H. Chang, C.-N. Chou, and E.~Y. Chang.
\newblock Clkn: Cascaded lucas-kanade networks for image alignment.
\newblock In {\em Proceedings of the IEEE conference on computer vision and
  pattern recognition}, pages 2213--2221, 2017.

\bibitem{chen2025visual}
J.~Chen, G.~Wen, H.~Jian, and X.~Fan.
\newblock A visual localization benchmark for uavs in complex multi-terrain
  environments.
\newblock {\em IEEE Journal of Selected Topics in Applied Earth Observations
  and Remote Sensing}, 2025.

\bibitem{real_time_geo_loc_21}
S.~Chen, X.~Wu, M.~W. Mueller, and K.~Sreenath.
\newblock Real-time geo-localization using satellite imagery and topography for
  unmanned aerial vehicles.
\newblock In {\em 2021 IEEE/RSJ international conference on intelligent robots
  and systems (IROS)}, pages 2275--2281. IEEE, 2021.

\bibitem{oblique_robust_abs_vis_loc_23}
Y.~Chen and J.~Jiang.
\newblock An oblique-robust absolute visual localization method for gps-denied
  uav with satellite imagery.
\newblock {\em IEEE Transactions on Geoscience and Remote Sensing}, 62:1--13,
  2023.

\bibitem{optimal_rand_ransac_08}
O.~Chum and J.~Matas.
\newblock Optimal randomized ransac.
\newblock {\em IEEE Transactions on Pattern Analysis and Machine Intelligence},
  30(8):1472--1482, 2008.

\bibitem{lo_ransac_03}
O.~Chum, J.~Matas, and J.~Kittler.
\newblock Locally optimized ransac.
\newblock In {\em Pattern Recognition: 25th DAGM Symposium, Magdeburg, Germany,
  September 10-12, 2003. Proceedings 25}, pages 236--243. Springer, 2003.

\bibitem{alto_22}
I.~Cisneros, P.~Yin, J.~Zhang, H.~Choset, and S.~Scherer.
\newblock Alto: A large-scale dataset for uav visual place recognition and
  localization.
\newblock {\em arXiv preprint arXiv:2207.12317}, 2022.

\bibitem{cledat2020mapping}
E.~Cledat, L.~V. Jospin, D.~A. Cucci, and J.~Skaloud.
\newblock Mapping quality prediction for rtk/ppk-equipped micro-drones
  operating in complex natural environment.
\newblock {\em ISPRS Journal of Photogrammetry and Remote Sensing}, 167:24--38,
  2020.

\bibitem{eu_23}
E.~Commission.
\newblock Commission implementing regulation (eu) 2023/138 laying down a list
  of specific high-value datasets and the arrangements for their publication
  and re-use, 2023.
\newblock Accessed: 2025-02-20.

\bibitem{vision_uav_self_pos_23}
M.~Dai, E.~Zheng, Z.~Feng, L.~Qi, J.~Zhuang, and W.~Yang.
\newblock Vision-based uav self-positioning in low-altitude urban environments.
\newblock {\em IEEE Transactions on Image Processing}, 33:493--508, 2023.

\bibitem{superpoint_18}
D.~DeTone, T.~Malisiewicz, and A.~Rabinovich.
\newblock Superpoint: Self-supervised interest point detection and description.
\newblock In {\em Proceedings of the IEEE conference on computer vision and
  pattern recognition workshops}, pages 224--236, 2018.

\bibitem{dji_terra}
DJI.
\newblock Dji terra.
\newblock \url{https://enterprise.dji.com/dji-terra}.

\bibitem{dkm_23}
J.~Edstedt, I.~Athanasiadis, M.~Wadenb{\"a}ck, and M.~Felsberg.
\newblock Dkm: Dense kernelized feature matching for geometry estimation.
\newblock In {\em Proceedings of the IEEE/CVF Conference on Computer Vision and
  Pattern Recognition}, pages 17765--17775, 2023.

\bibitem{dedode_24}
J.~Edstedt, G.~B{\"o}kman, M.~Wadenb{\"a}ck, and M.~Felsberg.
\newblock Dedode: Detect, don’t describe—describe, don’t detect for local
  feature matching.
\newblock In {\em 2024 International Conference on 3D Vision (3DV)}, pages
  148--157. IEEE, 2024.

\bibitem{roma_24}
J.~Edstedt, Q.~Sun, G.~B{\"o}kman, M.~Wadenb{\"a}ck, and M.~Felsberg.
\newblock Roma: Robust dense feature matching.
\newblock In {\em Proceedings of the IEEE/CVF Conference on Computer Vision and
  Pattern Recognition}, pages 19790--19800, 2024.

\bibitem{complete_sol_p3p_03}
X.-S. Gao, X.-R. Hou, J.~Tang, and H.-F. Cheng.
\newblock Complete solution classification for the perspective-three-point
  problem.
\newblock {\em IEEE transactions on pattern analysis and machine intelligence},
  25(8):930--943, 2003.

\bibitem{self_supervised_fine_grained_20}
Y.~Ge, H.~Wang, F.~Zhu, R.~Zhao, and H.~Li.
\newblock Self-supervising fine-grained region similarities for large-scale
  image localization.
\newblock In {\em Computer Vision--ECCV 2020: 16th European Conference,
  Glasgow, UK, August 23--28, 2020, Proceedings, Part IV 16}, pages 369--386.
  Springer, 2020.

\bibitem{end2end_learning_img_retrieval_17}
A.~Gordo, J.~Almazan, J.~Revaud, and D.~Larlus.
\newblock End-to-end learning of deep visual representations for image
  retrieval.
\newblock {\em International Journal of Computer Vision}, 124(2):237--254,
  2017.

\bibitem{absolute_pose_estim_24}
W.~Hanyu, S.~Qiang, D.~Zilong, C.~Xinyi, and X.~Wang.
\newblock Absolute pose estimation of uav based on large-scale satellite image.
\newblock {\em Chinese Journal of Aeronautics}, 37(6):219--231, 2024.

\bibitem{review_pnp_94}
B.~M. Haralick, C.-N. Lee, K.~Ottenberg, and M.~N{\"o}lle.
\newblock Review and analysis of solutions of the three point perspective pose
  estimation problem.
\newblock {\em International journal of computer vision}, 13:331--356, 1994.

\bibitem{wildlife_monitoring_16}
J.~C. Hodgson, S.~M. Baylis, R.~Mott, A.~Herrod, and R.~H. Clarke.
\newblock Precision wildlife monitoring using unmanned aerial vehicles.
\newblock {\em Scientific reports}, 6(1):22574, 2016.

\bibitem{curriculumloc_24}
B.~Hu, L.~Chen, R.~Chen, S.~Bu, P.~Han, and H.~Li.
\newblock Curriculumloc: Enhancing cross-domain geolocalization through
  multi-stage refinement.
\newblock {\em IEEE Transactions on Geoscience and Remote Sensing}, 2024.

\bibitem{robust_img_retrieval_20}
M.~Humenberger, Y.~Cabon, N.~Guerin, J.~Morat, V.~Leroy, J.~Revaud, P.~Rerole,
  N.~Pion, C.~de~Souza, and G.~Csurka.
\newblock Robust image retrieval-based visual localization using kapture.
\newblock {\em arXiv preprint arXiv:2007.13867}, 2020.

\bibitem{gta_uav_25}
Y.~Ji, B.~He, Z.~Tan, and L.~Wu.
\newblock Game4loc: A uav geo-localization benchmark from game data.
\newblock In {\em Proceedings of the AAAI Conference on Artificial
  Intelligence}, volume~39, pages 3913--3921, 2025.

\bibitem{minima_24}
X.~Jiang, J.~Ren, Z.~Li, X.~Zhou, D.~Liang, and X.~Bai.
\newblock Minima: Modality invariant image matching.
\newblock {\em arXiv preprint arXiv:2412.19412}, 2024.

\bibitem{inspection_18}
S.~Jordan, J.~Moore, S.~Hovet, J.~Box, J.~Perry, K.~Kirsche, D.~Lewis, and
  Z.~T.~H. Tse.
\newblock State-of-the-art technologies for uav inspections.
\newblock {\em IET Radar, Sonar \& Navigation}, 12(2):151--164, 2018.

\bibitem{poisson_2013}
M.~Kazhdan and H.~Hoppe.
\newblock Screened poisson surface reconstruction.
\newblock {\em ACM Transactions on Graphics (ToG)}, 32(3):1--13, 2013.

\bibitem{novel_param_p3p_11}
L.~Kneip, D.~Scaramuzza, and R.~Siegwart.
\newblock A novel parametrization of the perspective-three-point problem for a
  direct computation of absolute camera position and orientation.
\newblock In {\em CVPR 2011}, pages 2969--2976. IEEE, 2011.

\bibitem{epnp}
V.~Lepetit, F.~Moreno-Noguer, and P.~Fua.
\newblock Epnp: An accurate o(n) solution to the pnp problem.
\newblock {\em International journal of computer vision}, 81:155--166, 2009.

\bibitem{mast3r_24}
V.~Leroy, Y.~Cabon, and J.~Revaud.
\newblock Grounding image matching in 3d with mast3r.
\newblock In {\em European Conference on Computer Vision}, pages 71--91.
  Springer, 2024.

\bibitem{geoformer_24}
Q.~Li, X.~Yang, J.~Fan, R.~Lu, B.~Tang, S.~Wang, and S.~Su.
\newblock Geoformer: An effective transformer-based siamese network for uav
  geo-localization.
\newblock {\em IEEE Journal of Selected Topics in Applied Earth Observations
  and Remote Sensing}, 2024.

\bibitem{matrix_city_23}
Y.~Li, L.~Jiang, L.~Xu, Y.~Xiangli, Z.~Wang, D.~Lin, and B.~Dai.
\newblock Matrixcity: A large-scale city dataset for city-scale neural
  rendering and beyond.
\newblock In {\em Proceedings of the IEEE/CVF International Conference on
  Computer Vision}, pages 3205--3215, 2023.

\bibitem{lightglue_23}
P.~Lindenberger, P.-E. Sarlin, and M.~Pollefeys.
\newblock Lightglue: Local feature matching at light speed.
\newblock In {\em Proceedings of the IEEE/CVF International Conference on
  Computer Vision}, pages 17627--17638, 2023.

\bibitem{sift_2004}
D.~G. Lowe.
\newblock Distinctive image features from scale-invariant keypoints.
\newblock {\em International journal of computer vision}, 60:91--110, 2004.

\bibitem{marquardt1963algorithm}
D.~W. Marquardt.
\newblock An algorithm for least-squares estimation of nonlinear parameters.
\newblock {\em Journal of the society for Industrial and Applied Mathematics},
  11(2):431--441, 1963.

\bibitem{accurate_non_iter_pnp_07}
F.~Moreno-Noguer, V.~Lepetit, and P.~Fua.
\newblock Accurate non-iterative o(n) solution to the pnp problem.
\newblock In {\em 2007 IEEE 11th International Conference on Computer Vision},
  pages 1--8. Ieee, 2007.

\bibitem{meshloc_22}
V.~Panek, Z.~Kukelova, and T.~Sattler.
\newblock Meshloc: Mesh-based visual localization.
\newblock In {\em European Conference on Computer Vision}, pages 589--609.
  Springer, 2022.

\bibitem{eu_19}
E.~Parliament and C.~of~the European~Union.
\newblock Directive (eu) 2019/1024 on open data and the re-use of public sector
  information, 2019.
\newblock Accessed: 2025-02-20.

\bibitem{vis_loc_google_earth_20}
B.~Patel, T.~D. Barfoot, and A.~P. Schoellig.
\newblock Visual localization with google earth images for robust global pose
  estimation of uavs.
\newblock In {\em 2020 IEEE international conference on robotics and automation
  (ICRA)}, pages 6491--6497. IEEE, 2020.

\bibitem{xfeat_24}
G.~Potje, F.~Cadar, A.~Araujo, R.~Martins, and E.~R. Nascimento.
\newblock Xfeat: Accelerated features for lightweight image matching.
\newblock In {\em Proceedings of the IEEE/CVF Conference on Computer Vision and
  Pattern Recognition}, pages 2682--2691, 2024.

\bibitem{superglue_20}
P.-E. Sarlin, D.~DeTone, T.~Malisiewicz, and A.~Rabinovich.
\newblock Superglue: Learning feature matching with graph neural networks.
\newblock In {\em Proceedings of the IEEE/CVF conference on computer vision and
  pattern recognition}, pages 4938--4947, 2020.

\bibitem{large_scale_location_recog_16}
T.~Sattler, M.~Havlena, K.~Schindler, and M.~Pollefeys.
\newblock Large-scale location recognition and the geometric burstiness
  problem.
\newblock In {\em Proceedings of the IEEE conference on computer vision and
  pattern recognition}, pages 1582--1590, 2016.

\bibitem{rescue_15}
J.~Scherer, S.~Yahyanejad, S.~Hayat, E.~Yanmaz, T.~Andre, A.~Khan,
  V.~Vukadinovic, C.~Bettstetter, H.~Hellwagner, and B.~Rinner.
\newblock An autonomous multi-uav system for search and rescue.
\newblock In {\em Proceedings of the first workshop on micro aerial vehicle
  networks, systems, and applications for civilian use}, pages 33--38, 2015.

\bibitem{vpair_22}
M.~Schleiss, F.~Rouatbi, and D.~Cremers.
\newblock Vpair--aerial visual place recognition and localization in
  large-scale outdoor environments.
\newblock {\em arXiv preprint arXiv:2205.11567}, 2022.

\bibitem{colmap_16}
J.~L. Schonberger and J.-M. Frahm.
\newblock Structure-from-motion revisited.
\newblock In {\em Proceedings of the IEEE conference on computer vision and
  pattern recognition}, pages 4104--4113, 2016.

\bibitem{gim_24}
X.~Shen, Z.~Cai, W.~Yin, M.~M{\"u}ller, Z.~Li, K.~Wang, X.~Chen, and C.~Wang.
\newblock Gim: Learning generalizable image matcher from internet videos.
\newblock {\em arXiv preprint arXiv:2402.11095}, 2024.

\bibitem{uav_pose_estim_19}
A.~Shetty and G.~X. Gao.
\newblock Uav pose estimation using cross-view geolocalization with satellite
  imagery.
\newblock In {\em 2019 International Conference on Robotics and Automation
  (ICRA)}, pages 1827--1833. IEEE, 2019.

\bibitem{spartial_aware_19}
Y.~Shi, L.~Liu, X.~Yu, and H.~Li.
\newblock Spatial-aware feature aggregation for image based cross-view
  geo-localization.
\newblock {\em Advances in Neural Information Processing Systems}, 32, 2019.

\bibitem{loftr_21}
J.~Sun, Z.~Shen, Y.~Wang, H.~Bao, and X.~Zhou.
\newblock Loftr: Detector-free local feature matching with transformers.
\newblock In {\em Proceedings of the IEEE/CVF conference on computer vision and
  pattern recognition}, pages 8922--8931, 2021.

\bibitem{cross_view_geo_loc_23}
J.~Sun, R.~Yan, B.~Zhang, B.~Zhu, and F.~Sun.
\newblock A cross-view geo-localization method guided by relation-aware global
  attention.
\newblock {\em Multimedia Systems}, 29(4):2205--2216, 2023.

\bibitem{inloc_18}
H.~Taira, M.~Okutomi, T.~Sattler, M.~Cimpoi, M.~Pollefeys, J.~Sivic, T.~Pajdla,
  and A.~Torii.
\newblock Inloc: Indoor visual localization with dense matching and view
  synthesis.
\newblock In {\em Proceedings of the IEEE Conference on Computer Vision and
  Pattern Recognition}, pages 7199--7209, 2018.

\bibitem{digital_twin_1}
A.~To, M.~Liu, M.~Hazeeq Bin Muhammad~Hairul, J.~G. Davis, J.~S. Lee, H.~Hesse,
  and H.~D. Nguyen.
\newblock Drone-based ai and 3d reconstruction for digital twin augmentation.
\newblock In {\em International conference on human-computer interaction},
  pages 511--529. Springer, 2021.

\bibitem{place_recog_24_7_15}
A.~Torii, R.~Arandjelovic, J.~Sivic, M.~Okutomi, and T.~Pajdla.
\newblock 24/7 place recognition by view synthesis.
\newblock In {\em Proceedings of the IEEE conference on computer vision and
  pattern recognition}, pages 1808--1817, 2015.

\bibitem{xoftr_24}
{\"O}.~Tuzcuo{\u{g}}lu, A.~K{\"o}ksal, B.~Sofu, S.~Kalkan, and A.~A. Alatan.
\newblock Xoftr: Cross-modal feature matching transformer.
\newblock In {\em Proceedings of the IEEE/CVF Conference on Computer Vision and
  Pattern Recognition}, pages 4275--4286, 2024.

\bibitem{pixpro}
P.~UAB.
\newblock Pixpro photogrammetry software.
\newblock \url{https://www.pix-pro.com/}.

\bibitem{dust3r_24}
S.~Wang, V.~Leroy, Y.~Cabon, B.~Chidlovskii, and J.~Revaud.
\newblock Dust3r: Geometric 3d vision made easy.
\newblock In {\em Proceedings of the IEEE/CVF Conference on Computer Vision and
  Pattern Recognition}, pages 20697--20709, 2024.

\bibitem{eloftr_24}
Y.~Wang, X.~He, S.~Peng, D.~Tan, and X.~Zhou.
\newblock Efficient loftr: Semi-dense local feature matching with sparse-like
  speed.
\newblock In {\em Proceedings of the IEEE/CVF Conference on Computer Vision and
  Pattern Recognition}, pages 21666--21675, 2024.

\bibitem{uavd4l_24}
R.~Wu, X.~Cheng, J.~Zhu, Y.~Liu, M.~Zhang, and S.~Yan.
\newblock Uavd4l: A large-scale dataset for uav 6-dof localization.
\newblock In {\em 2024 International Conference on 3D Vision (3DV)}, pages
  1574--1583. IEEE, 2024.

\bibitem{uav_visloc_24}
W.~Xu, Y.~Yao, J.~Cao, Z.~Wei, C.~Liu, J.~Wang, and M.~Peng.
\newblock Uav-visloc: A large-scale dataset for uav visual localization.
\newblock {\em arXiv preprint arXiv:2405.11936}, 2024.

\bibitem{crossloc_22}
Q.~Yan, J.~Zheng, S.~Reding, S.~Li, and I.~Doytchinov.
\newblock Crossloc: Scalable aerial localization assisted by multimodal
  synthetic data.
\newblock In {\em Proceedings of the IEEE/CVF Conference on Computer Vision and
  Pattern Recognition}, pages 17358--17368, 2022.

\bibitem{render_and_compare_23}
S.~Yan, X.~Cheng, Y.~Liu, J.~Zhu, R.~Wu, Y.~Liu, and M.~Zhang.
\newblock Render-and-compare: Cross-view 6-dof localization from noisy prior.
\newblock In {\em 2023 IEEE International Conference on Multimedia and Expo
  (ICME)}, pages 2171--2176. IEEE, 2023.

\bibitem{sensloc_23}
S.~Yan, Y.~Liu, L.~Wang, Z.~Shen, Z.~Peng, H.~Liu, M.~Zhang, G.~Zhang, and
  X.~Zhou.
\newblock Long-term visual localization with mobile sensors.
\newblock In {\em Proceedings of the IEEE/CVF Conference on Computer Vision and
  Pattern Recognition}, pages 17245--17255, 2023.

\bibitem{cross_view_21}
H.~Yang, X.~Lu, and Y.~Zhu.
\newblock Cross-view geo-localization with layer-to-layer transformer.
\newblock {\em Advances in Neural Information Processing Systems},
  34:29009--29020, 2021.

\bibitem{anyvisloc_25}
Y.~Ye, X.~Teng, S.~Chen, Z.~Li, L.~Liu, Q.~Yu, and T.~Tan.
\newblock Exploring the best way for uav visual localization under low-altitude
  multi-view observation condition: a benchmark.
\newblock {\em arXiv preprint arXiv:2503.10692}, 2025.

\bibitem{vision_based_abs_loc_14}
A.~Yol, B.~Delabarre, A.~Dame, J.-E. Dartois, and E.~Marchand.
\newblock Vision-based absolute localization for unmanned aerial vehicles.
\newblock In {\em 2014 IEEE/RSJ International Conference on Intelligent Robots
  and Systems}, pages 3429--3434. IEEE, 2014.

\bibitem{digital_twin_2}
J.~Yoon, Y.~Kim, S.~Lee, and M.~Shin.
\newblock Uav-based automated 3d modeling framework using deep learning for
  building energy modeling.
\newblock {\em Sustainable Cities and Society}, 101:105169, 2024.

\bibitem{zhang2002flexible}
Z.~Zhang.
\newblock A flexible new technique for camera calibration.
\newblock {\em IEEE Transactions on pattern analysis and machine intelligence},
  22(11):1330--1334, 2002.

\bibitem{ref_pose_gen_21}
Z.~Zhang, T.~Sattler, and D.~Scaramuzza.
\newblock Reference pose generation for long-term visual localization via
  learned features and view synthesis.
\newblock {\em International Journal of Computer Vision}, 129(4):821--844,
  2021.

\bibitem{university_1652_20}
Z.~Zheng, Y.~Wei, and Y.~Yang.
\newblock University-1652: A multi-view multi-source benchmark for drone-based
  geo-localization.
\newblock In {\em Proceedings of the 28th ACM international conference on
  Multimedia}, pages 1395--1403, 2020.

\bibitem{lodloc_24}
J.~Zhu, S.~Yan, L.~Wang, z.~shengYue, Y.~Liu, and M.~Zhang.
\newblock Lod-loc: Aerial visual localization using lod 3d map with neural
  wireframe alignment.
\newblock In A.~Globerson, L.~Mackey, D.~Belgrave, A.~Fan, U.~Paquet,
  J.~Tomczak, and C.~Zhang, editors, {\em Advances in Neural Information
  Processing Systems}, volume~37, pages 119063--119098. Curran Associates,
  Inc., 2024.

\bibitem{sues_200_23}
R.~Zhu, L.~Yin, M.~Yang, F.~Wu, Y.~Yang, and W.~Hu.
\newblock Sues-200: A multi-height multi-scene cross-view image benchmark
  across drone and satellite.
\newblock {\em IEEE Transactions on Circuits and Systems for Video Technology},
  33(9):4825--4839, 2023.

\bibitem{geo_semantic_21}
Y.~Zhu, B.~Sun, X.~Lu, and S.~Jia.
\newblock Geographic semantic network for cross-view image geo-localization.
\newblock {\em IEEE Transactions on Geoscience and Remote Sensing}, 60:1--15,
  2021.

\bibitem{semantic_guidance_transformer_22}
J.~Zhuang, X.~Chen, M.~Dai, W.~Lan, Y.~Cai, and E.~Zheng.
\newblock A semantic guidance and transformer-based matching method for uavs
  and satellite images for uav geo-localization.
\newblock {\em Ieee Access}, 10:34277--34287, 2022.

\bibitem{faster_effective_cross_view_uav_21}
J.~Zhuang, M.~Dai, X.~Chen, and E.~Zheng.
\newblock A faster and more effective cross-view matching method of uav and
  satellite images for uav geolocalization.
\newblock {\em Remote Sensing}, 13(19):3979, 2021.

\end{thebibliography}
}


\clearpage
\appendix
\addcontentsline{toc}{section}{Appendix} 
\part{Appendix}
\parttoc 
\clearpage
\section{Mathematical Formulation}
\label{appendix:mathematical_formulation}

This section establishes the mathematical foundation for our approach to localization and calibration using orthographic geodata. We present the camera projection models and the 3D lifting operation from a \ac{DOP} raster to 3D coordinates using a \ac{DSM} raster.

\subsection{General Camera Projection Model}
\label{appendix:general_projection}
The general form of projecting a 3D point onto a camera image plane is given by:

\begin{equation}
    \begin{aligned}
        \pi :& \, \mathbf{P} \mapsto \mathbf{p} \, ,\\
        \lambda \tilde{\mathbf{p}} =& \, 
        \mathbf{K} \mathbf{\Pi} \mathbf{T} \tilde{\mathbf{P}} \, ,
    \end{aligned}
    \label{eq:general_projection}
\end{equation}

where $\tilde{\mathbf{P}} \in \mathbb{R}^4$ is a 3D point in homogeneous coordinates, $\mathbf{K} = \begin{bmatrix} f_x & 0 & c_x\\  0 & f_y & c_y\\ 0 & 0 & 1 \end{bmatrix} \in \mathbb{R}^{3 \times 3}$ is the camera intrinsic matrix with focal lengths $f_x$, $f_y$ and principal points $c_x$, $c_y$, $\mathbf{T} = \begin{bmatrix} \mathbf{R} & \mathbf{t} \\ \mathbf{0}^\top & 1 \end{bmatrix} \in \mathbb{R}^{4 \times 4}$ is the extrinsic matrix representing the world-to-camera pose with $\mathbf{R} \in SO(3)$ and $\mathbf{t} \in \mathbb{R}^3$, $\mathbf{\Pi} \in \mathbb{R}^{3 \times 4}$ is a projection matrix, $\tilde{\mathbf{p}} \in \mathbb{R}^3$ is the projected point in homogeneous image coordinates, and $\lambda$ is a scale factor. Note that for simplicity, we ignore distortion effects in the following.

\begin{figure}[htbp]
    \centering
    \includegraphics[width=0.5\linewidth]{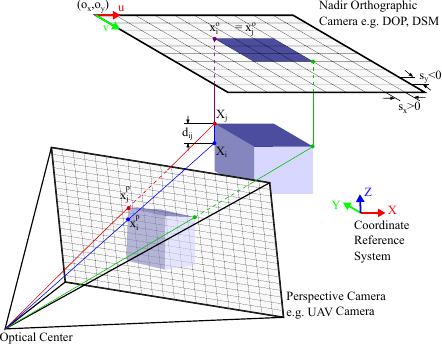}
    \caption{\textbf{Comparison of Projection Models Used in This Work.} The perspective projection, commonly employed in \ac{UAV} cameras, features rays converging at a single camera center, resulting in perspective effects where parallel lines in the real world appear to converge in the image. In contrast, the orthographic nadir projection, often applied to raster geodata, uses parallel vertical rays that preserve scale relationships and spatial accuracy in the world.}
    \label{fig:stereo_projection}
    \vspace{-1\baselineskip}
\end{figure}

Our work leverages two distinct projection models: perspective projection from \ac{UAV} cameras and orthographic projection (specifically nadir view) used in raster geodata. We illustrate both projections in~\Cref{fig:stereo_projection}.

\subsection{Perspective Projection}
\label{appendix:perspective_projection}
A characteristic of perspective projection $\pi_p$ is the intersection of all rays at a single point (camera center), accounting for perspective effects that make parallel lines in the world intersect in the image plane. In this case, $f_x > 0$, $f_y > 0$, $\lambda$ corresponds to the depth of the 3D point with respect to the camera, and $\mathbf{\Pi} = \begin{bmatrix} \mathbf{I}_3 & \mathbf{0} \end{bmatrix}$ with $\mathbf{I}_3$ being the $3 \times 3$ identity matrix and $\mathbf{0}$ a column vector of zeros. We use perspective projection for modeling \ac{UAV} imagery.

\subsection{Nadir Orthographic Projection}
\label{appendix:orthographic_projection}
In orthographic nadir projections, all rays passing through the camera are parallel and strictly vertical, yielding a camera center at infinity. Projection lines are perpendicular to the XY plane, resulting in parallel rays without perspective effects. This characteristic makes orthographic projection the standard representation for geodata such as satellite imagery and digital elevation models.

A nadir orthographic camera is defined by an origin $(o_x, o_y)$ (the top left raster position in common reference frames) with scales $s_x$ and $s_y$ defining the metric grid cell size. Unlike perspective cameras, orthographic cameras maintain the same distance relationships in pixel coordinates as in world coordinates, scaled by $s_x$ and $s_y$.

For the nadir orthographic projection function $\pi_o$, we can derive the closed-form formulation yielding $\lambda=1$, $\mathbf{R} = \mathbf{I}_3$, $\mathbf{t} = \begin{bmatrix}-o_x & -o_y & 0\end{bmatrix}^\top$, $f_x = 1/s_x$, $f_y = 1/s_y$, and $c_x = c_y = 0$. The projection matrix is $\mathbf{\Pi} = \begin{bmatrix} 1 & 0 & 0 & 0 \\ 0 & 1 & 0 & 0 \\ 0 & 0 & 0 & 1 \end{bmatrix}$, which eliminates the Z component, effectively collapsing 3D points onto a plane.

\paragraph{Proof.}

In raster geodata such as \acp{DOP} and \acp{DSM} with $(o_x, o_y)$ describing the XY position of the origin in predefined 3D geographic reference coordinate system (e.g., UTM) and pixel size $(s_x, s_y)$ in metric space, a pixel coordinate $(x, y)$ can be mapped to its X and Y coordinates in that 3D reference coordinate system using a simple linear transformation:

\begin{equation}
    \begin{bmatrix}
        X\\ Y
    \end{bmatrix} = \begin{bmatrix}
        s_x x + o_x \\
        s_y y + o_y
    \end{bmatrix} \, .
\end{equation}

We can extract $x$ and $y$ as:

\begin{equation}
    \begin{aligned}
    x &= \frac{X - o_x}{s_x}\, , \\
    y &= \frac{Y - o_y}{s_y}\, .
    \end{aligned} 
\end{equation}

In matrix form, this gives:

\begin{equation}
    \begin{bmatrix}
        x\\ y\\ 1
    \end{bmatrix} = \begin{bmatrix}
        \frac{1}{s_x} & 0 & 0 & -\frac{o_x}{s_x} \\
        0 & \frac{1}{s_y} & 0 & -\frac{o_y}{s_y} \\
        0 & 0 & 0 & 1
    \end{bmatrix}
    \begin{bmatrix}
        X\\ Y\\ Z\\ 1
    \end{bmatrix} \, .
\end{equation}

Decomposing this compact projection matrix into the form $\mathbf{K}\mathbf{\Pi}\mathbf{T}$ yields:

\begin{equation}
\mathbf{K}\mathbf{\Pi}\mathbf{T} = 
    \begin{bmatrix}
        \frac{1}{s_x} & 0 & 0 \\
        0 & \frac{1}{s_y} & 0 \\
        0 & 0 & 1
    \end{bmatrix}
    \begin{bmatrix}
        1 & 0 & 0 & 0 \\
        0 & 1 & 0 & 0 \\
        0 & 0 & 0 & 1
    \end{bmatrix}
    \begin{bmatrix}
        1 & 0 & 0 & -o_x \\
        0 & 1 & 0 & -o_y \\
        0 & 0 & 1 & 0 \\
        0 & 0 & 0 & 1
    \end{bmatrix} \, .
\end{equation}

In standard geospatial data conventions, the origin is typically at the top-left corner of the raster, with $s_x > 0$ and $s_y < 0$. This negative $s_y$ accounts for the fact that the y-axis in pixel coordinates increases downward, while in world coordinates it increases upward. Given this convention, we have $f_x = 1/s_x > 0$ and $f_y = 1/s_y < 0$. This negative focal length has no direct physical counterpart in traditional optical systems.
Nevertheless, this mathematical abstraction effectively represents the orthographic projection found in aerial mapping data, while enabling us to represent both orthographic and perspective projections in a unified formulation.

\subsection{Lifting 2D Coordinates in DOP to 3D Coordinates using DSM}
\label{appendix:relation_dsm_dop}
Our framework leverages orthographic geodata in the form of \ac{DSM} and \ac{DOP} rasters, denoted as $\mathbf{R}^{\text{DSM}} \in \mathbb{R}^{W^{\text{DSM}} \times H^{\text{DSM}}}$ and $\mathbf{R}^{\text{DOP}} \in \mathbb{R}^{W^{\text{DOP}} \times H^{\text{DOP}} \times 3}$, respectively. These rasters are characterized by scales $(s_x^{\text{DSM}}, s_y^{\text{DSM}})$, $(s_x^{\text{DOP}}, s_y^{\text{DOP}})$, and origins $(o_x^{\text{DSM}}, o_y^{\text{DSM}})$, $(o_x^{\text{DOP}}, o_y^{\text{DOP}})$. Utilizing the linear correspondence between these representations, we derive 3D scene points from 2D coordinates in the \ac{DOP} raster as:
\begin{equation}
\mathbf{P}_i = \begin{bmatrix} {\mathbf{p}_i^{\text{DOP}}}^\top & \mathbf{R}^{\text{DSM}}(f(\mathbf{p}_i^{\text{DOP}})) \end{bmatrix}^\top \,,
\label{eq:dop_to_dsm}
\end{equation}

where $f: \mathbb{R}^2 \to \mathbb{R}^2$ is a linear transformation mapping coordinates between the rasters. This transformation is defined as:
\renewcommand{\arraystretch}{1.5}
\begin{equation}
\tilde{\mathbf{p}}_i^{\text{DSM}} =
\begin{pmatrix} 
\frac{s_x^{\text{DOP}}}{s_x^{\text{DSM}}} & 0 & \frac{o_x^{\text{DOP}} - o_x^{\text{DSM}}}{s_x^{\text{DSM}}} \\
0 & \frac{s_y^{\text{DOP}}}{s_y^{\text{DSM}}} & \frac{o_y^{\text{DOP}} - o_y^{\text{DSM}}}{s_y^{\text{DSM}}} \\
0 & 0 & 1
\end{pmatrix}
\tilde{\mathbf{p}}_i^{\text{DOP}},
\end{equation}
\renewcommand{\arraystretch}{1}

where $\tilde{\cdot}$ denotes homogeneous coordinates. In our dataset, both rasters share the same scales and origins, simplifying $f$ to the identity function. This formulation establishes 3D-2D correspondences between 2.5D orthographic geodata and \ac{UAV} imagery.

\subsection{Optimization of Camera Parameters}
Given a query image $I~\in~\mathbb{R}^{W^{I} \times H^{I} \times 3}$ captured by a \ac{UAV}, georeferenced camera calibration involves estimating camera parameters within a geospatial reference frame by minimizing a reprojection loss:

\begin{equation}
\begin{aligned}
    \mathbf{T}^*, \mathbf{K}^* &= \argmin_{\mathbf{T}, \mathbf{K}} \mathcal{L}_{\text{reproj}}, \\
    \mathcal{L}_{\text{reproj}} &= \sum_{i} \rho\left(\|\pi_p(\mathbf{P}_i, \mathbf{K}, \mathbf{T}) - \mathbf{p}^{I}_i\|_2\right),
\end{aligned}
\label{eq:camera_calibration}
\end{equation}

where $\mathbf{P}_i \in \mathbb{R}^3$ represents the 3D scene points, $\mathbf{p}_i^I \in \mathbb{R}^2$ corresponds to their associated 2D points in the image $I$, and $\rho(\cdot)$ is a robust cost function, specifically the Huber loss, which mitigates the impact of outlier correspondences. The calibration process involves optimizing the intrinsic and extrinsic parameters using the Levenberg-Marquardt~\cite{marquardt1963algorithm, zhang2002flexible} algorithm. For initialization, we assume the focal length is $f_x = f_y = \max(W^I, H^I)$ and derive the initial extrinsic parameters through RANSAC-EPnP~\cite{epnp}, employing a 5-pixel inlier threshold. In localization tasks, the intrinsic parameters $\mathbf{K}$ remain fixed, and only the extrinsic parameters $\mathbf{T}$ are estimated.

\section{\acs{dataset} Dataset Details}
\label{appendix:dataset_details}

\subsection{Dataset Statistics}
\label{appendix:dataset_stats}

Our dataset consists of 16,427 samples with raster sizes of 1024×1024 pixels and query image sizes of 1024×682 or 1024×767 pixels.

From the 51 locations in our dataset, 48 were split into training (13K samples) and validation (1.5K samples) sets to facilitate future work focused on training models using our data. Representative samples from these 48 locations were used to create an in-Place test set, reflecting performance on previously seen environments. The remaining 3 locations were reserved for an out-Place test set, designed to evaluate generalization to novel environments. The dataset samples are further categorized into three types: same domain, cross-domain within \ac{DOP}, and cross-domain between \ac{DOP} and \ac{DSM}.
\begin{table}[thpb]
\centering
\caption{\textbf{Distribution of Samples Across Dataset Splits.}}
\label{tab:samples_distribution}
\centering
\resizebox{0.8\textwidth}{!}{%
\begin{tabular}{l|r|rrrr}
\Xhline{1pt} \rowcolor{gray!15}
\textbf{Sample Type} & \textbf{All} & \textbf{Train} & \textbf{Val} & \textbf{Test In-Place} & \textbf{Test Out-Place} \\
\Xhline{0.1pt}
Same domain & 10,923 & 9,255 & 1,030 & 142 & 496  \\
\ac{DOP} cross-domain & 4,698 & 3,764 & 421 & 17 & 496 \\
\ac{DOP} \& \ac{DSM} cross-domains & 806 & 328 & 38 & 17 & 423 \\
\Xhline{0.1pt}
All types & 16,427 & 13,347 & 1,489 & 176 & 1,415 \\
\Xhline{1pt}
\end{tabular}%
}
\vspace{-0.5\baselineskip}
\end{table}

\begin{table}[thpb]
\centering
\caption{\textbf{Characteristics Across Sample Types and Dataset Splits.}}
\label{tab:dataset_characteristics}
\resizebox{\textwidth}{!}{%
\begin{tabular}{ll|r|rrr|rrrrr}
\Xhline{1pt}
\rowcolor{gray!15}
& & & \multicolumn{3}{c|}{\textbf{Sample Type}} & \multicolumn{4}{c}{\textbf{Dataset Split}} \\
\rowcolor{gray!15}
\textbf{Characteristic} & \textbf{Stats} & \textbf{All} 
 & \textbf{Same} & \textbf{DSP} & \textbf{DSP \&} & \textbf{Train} & \textbf{Val} & \textbf{Test} & \textbf{Test} \\
 \rowcolor{gray!15}
 &  &  & \textbf{domain} & \textbf{cross} & \textbf{DSM cross} &  &  & \textbf{inPlace} & \textbf{outPlace} \\
\Xhline{0.1pt}
\multirow{3}{*}{Obliqueness (deg)} & Mean & 14.6 & 14.3 & 14.6 & 18.6 & 14.0 & 14.4 & 18.6 & 20.4 \\
 & Min & 0.0 & 0.0 & 0.1 & 0.1 & 0.0 & 0.0 & 0.1 & 0.1 \\
 & Max & 86.8 & 86.8 & 56.3 & 30.9 & 86.8 & 55.8 & 59.7 & 56.3 \\
\Xhline{0.1pt}
\multirow{3}{*}{DSM area (m²)} & Mean & 37,512 & 36,061 & 40,860 & 37,660 & 36,903 & 37,477 & 49,434 & 41,810 \\
 & Min & 982 & 982 & 11,238 & 14,691 & 982 & 2,394 & 22,344 & 11,238 \\
 & Max & 370,686 & 370,686 & 370,686 & 60,569 & 370,686 & 337,931 & 241,607 & 370,686 \\
\Xhline{0.1pt}
\multirow{3}{*}{Elevation (m)} & Mean & 101.92 & 100 & 107 & 104 & 101 & 101 & 111 & 109 \\
 & Min & 23 & 23 & 72 & 74 & 23 & 24 & 99 & 72 \\
 & Max & 201 & 201 & 201 & 124 & 154 & 148 & 134 & 147 \\
\Xhline{0.1pt}
\multirow{3}{*}{Scale (cm/pix)} & Mean & 18.72 & 18.3 & 19.8 & 18.9 & 18.6 & 18.7 & 21.5 & 20.1 \\
 & Min & 4.1 & 4.1 & 11.3 & 11.9 & 4.1 & 4.8 & 15.9 & 11.3 \\
 & Max & 73.1 & 73.1 & 73.1 & 24.1 & 73.1 & 68.1 & 55.3 & 73.1 \\
\Xhline{0.1pt}
\multirow{3}{*}{Query visible area (m²)} & Mean & 18,402 & 17,329 & 20,896 & 18,406 & 18,061 & 18,822 & 23,666 & 20,524 \\
 & Min & 550 & 550 & 5,518 & 7,446 & 550 & 780 & 12,080 & 5,518 \\
 & Max & 286,785 & 286,785 & 286,785 & 26,678 & 286,785 & 271,372 & 148,775 & 286,785 \\
\Xhline{0.1pt}
\multirow{3}{*}{Covisibility (\%)} & Mean & 99.99 & 100.0 & 100.0 & 99.9 & 100.0 & 100.0 & 100.0 & 99.9 \\
 & Min & 95.7 & 99.4 & 99.6 & 95.7 & 95.7 & 98.7 & 99.9 & 95.7 \\
 & Max & 100.0 & 100.0 & 100.0 & 100.0 & 100.0 & 100.0 & 100.0 & 100.0 \\
\Xhline{1pt}
\end{tabular}%
}
\vspace{-1.5\baselineskip}
\end{table}

As shown in \Cref{tab:samples_distribution}, our dataset is well-distributed across different sample types, with the majority being same-domain samples (10,923 samples). The training set contains 13,347 samples (81.3\%), while validation and testing sets comprise 1,489 (9.1\%) and 1,591 (9.7\%) samples respectively.

\Cref{tab:dataset_characteristics} demonstrates the diversity in our dataset. We observe varying obliqueness angles (0° to 86.8°), elevations (23m to 201m), and scales (4.1cm/pix to 73.1cm/pix) across different sample types and splits. The covisibility remains consistently high (above 95.7\%) across all samples, ensuring quality matches between query and reference images.

The test sets feature higher average obliqueness angles and DSM areas compared to the training data, providing more challenging evaluation scenarios. This diversity across all characteristics makes our dataset well-suited for robust model training and evaluation across different geographical conditions.

\subsection{Samples Diversity}
\begin{figure}[ht]
    \centering
    \includegraphics[width=0.9\linewidth]{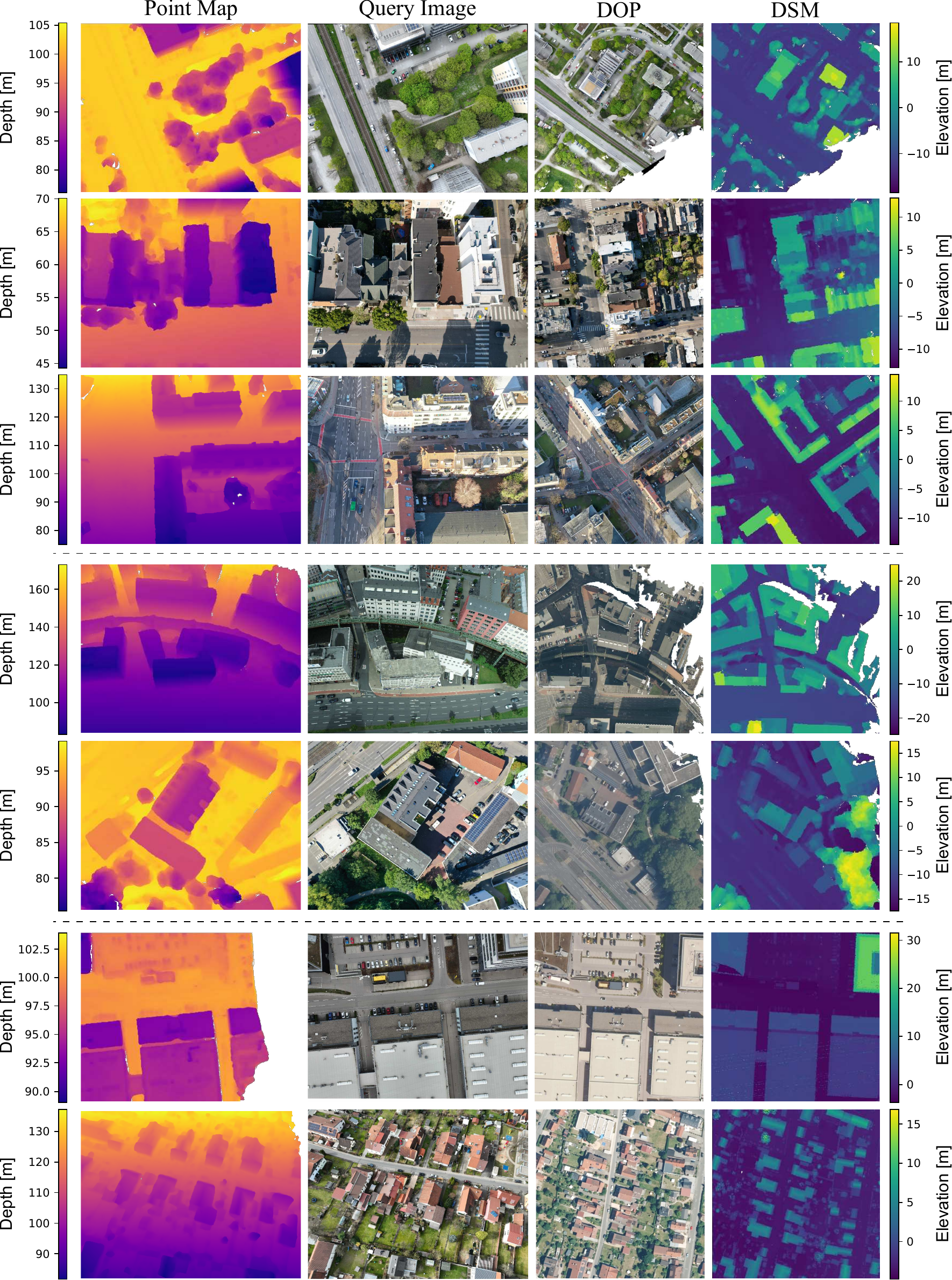}
    \caption{\textbf{Dataset Diversity Across Environments and Viewing Conditions.} Our dataset spans diverse scenes (urban, suburban, rural, industrial) and perspectives (nadir, oblique). \textbf{(Rows 1-3)}~Same domain for query data (query image, point map) and reference data (\acs{DOP}, \acs{DSM}); \textbf{(Rows 4-5)}~\ac{DOP} domain shifts; \textbf{(Rows 6-7)}~Combined \ac{DOP} and \ac{DSM} domain shifts.\protect\footnotemark}
    \label{fig:dataset_samples}
    \vspace{-1\baselineskip}
\end{figure}

\begin{figure}[htbp]
    \centering
    \begin{minipage}{0.245\textwidth}
    \includegraphics[width=\linewidth]{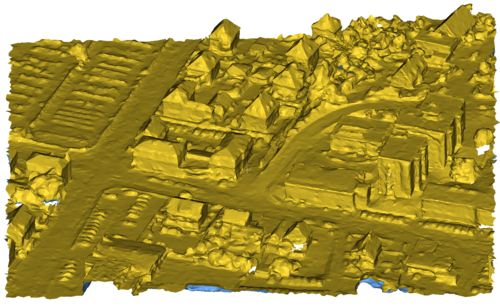}
    \end{minipage}
    \begin{minipage}{0.22455\textwidth}
        \includegraphics[width=\linewidth]{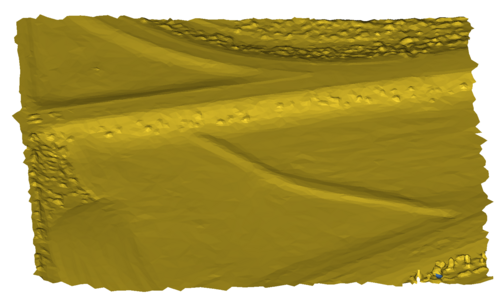}
    \end{minipage}
    \begin{minipage}{0.245\textwidth}
    \includegraphics[width=\linewidth]{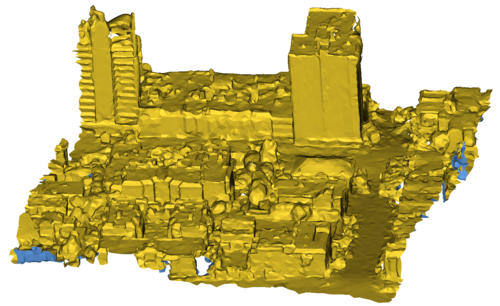}
    \end{minipage}
    \begin{minipage}{0.245\textwidth}
    \includegraphics[width=\linewidth]{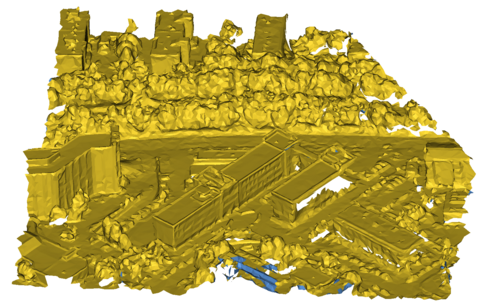}
    \end{minipage}
    \begin{minipage}{0.245\textwidth}
    \includegraphics[width=\linewidth]{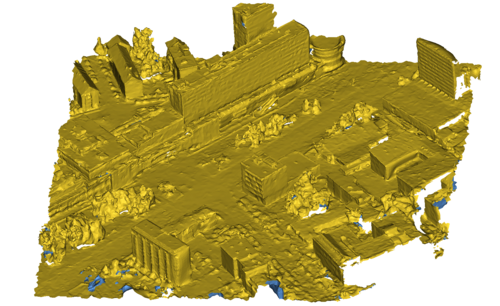}
    \end{minipage}
    \begin{minipage}{0.245\textwidth}
        \includegraphics[width=\linewidth]{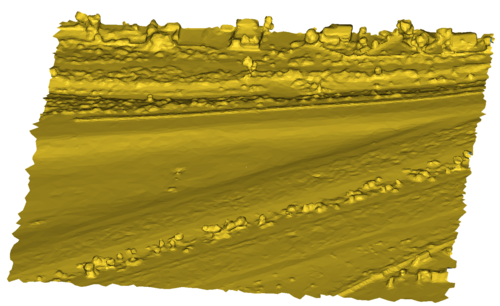}
    \end{minipage}
    \begin{minipage}{0.245\textwidth}
    \includegraphics[width=\linewidth]{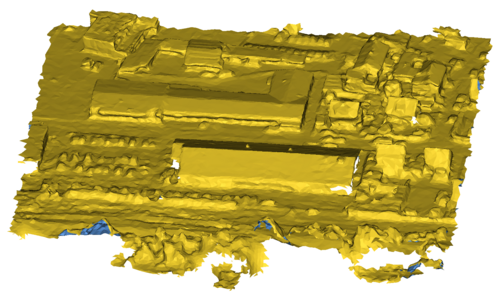}
    \end{minipage}
    \begin{minipage}{0.245\textwidth}
    \includegraphics[width=\linewidth]{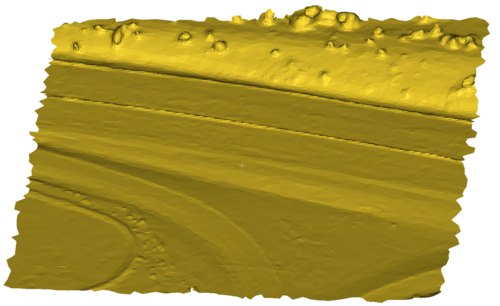}
    \end{minipage}
    \caption{\textbf{Examples of Local Meshes in \ac{dataset}.}}
    \label{fig:local_meshes}
\end{figure}

In addition to the statistics, we illustrate the diversity of our dataset by presenting representative samples from different environments and viewing conditions in~\Cref{fig:dataset_samples}. We also show local meshes for randomly picked samples from our dataset in~\Cref{fig:local_meshes}. These examples showcase the variability in scene content, viewpoint, and domain characteristics that make our dataset particularly challenging and representative of real-world conditions.

\subsection{Comparison with Existing Datasets}

We performed a detailed analysis of geometric consistency across recent aerial visual localization datasets to assess the accuracy of their ground-truth poses. Our evaluation projects 3D keypoints – extracted from high-precision \acp{DOP} and \acp{DSM} obtained from open geoportals – onto query images using the provided camera parameters, enabling visual verification of pose quality.

\begin{figure}[htbp]
\centering
\begin{subfigure}[b]{0.49\linewidth}
\includegraphics[width=\linewidth]{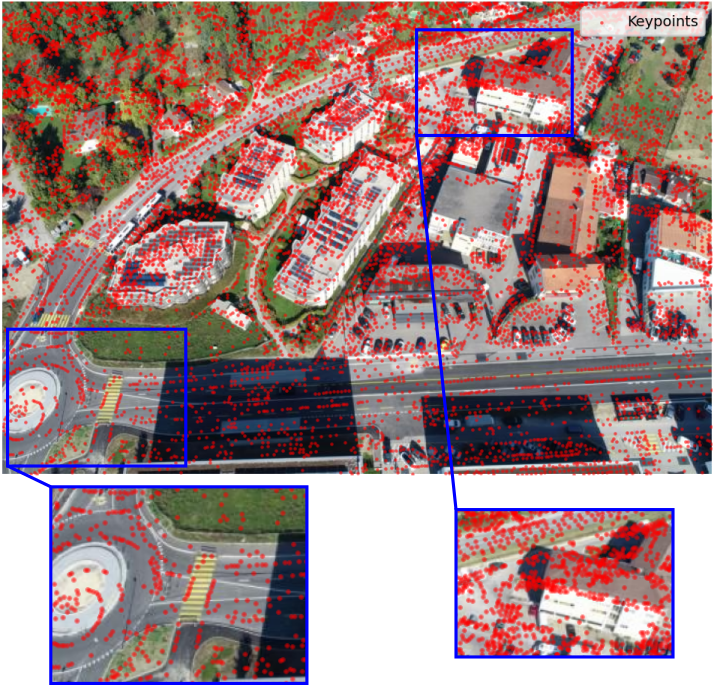}
\caption{CrossLoc~\cite{crossloc_22}}
\label{fig:crossloc_accuracy}
\end{subfigure}
\hfill
\begin{subfigure}[b]{0.49\linewidth}
\includegraphics[width=\linewidth]{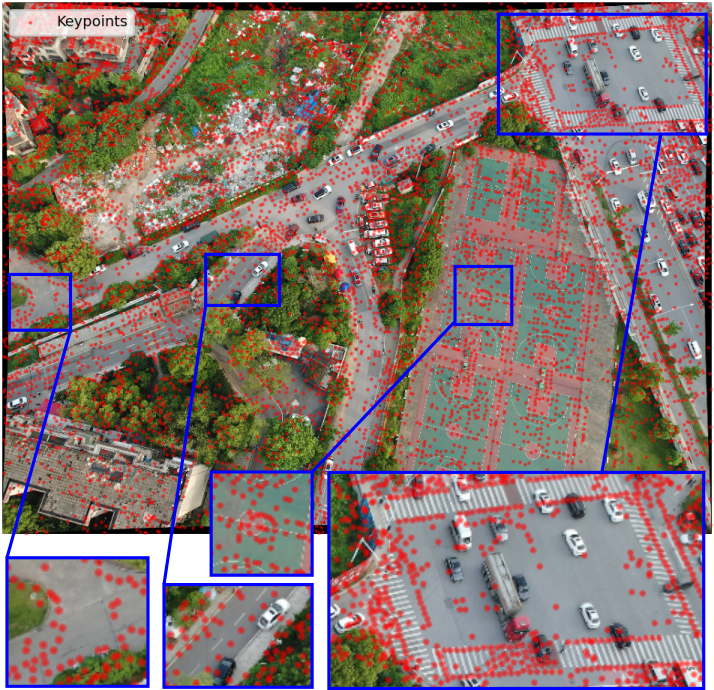}
\caption{UAVD4L~\cite{uavd4l_24}}
\label{fig:uavd4l_accuracy}
\end{subfigure}
\
\begin{subfigure}[b]{0.49\linewidth}
\includegraphics[width=\linewidth]{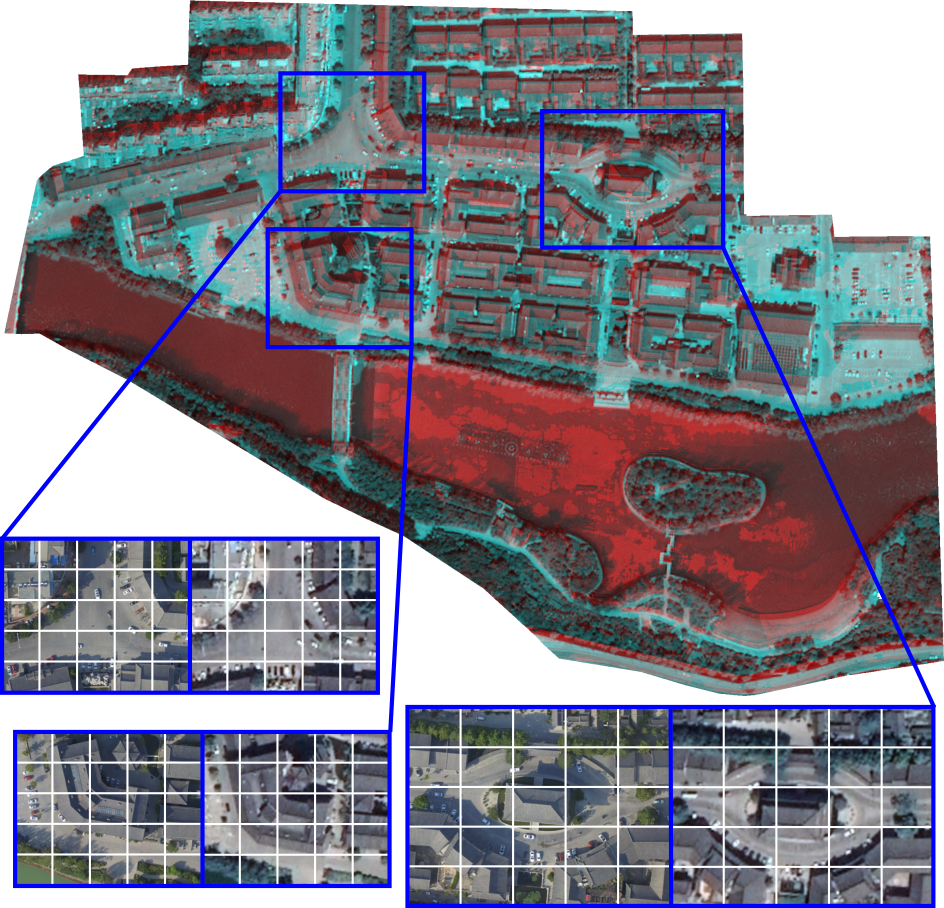}
\caption{AnyVisLoc~\cite{anyvisloc_25}}
\label{fig:anyvisloc_accuracy}
\end{subfigure}
\hfill
\begin{subfigure}[b]{0.49\linewidth}
\includegraphics[width=\linewidth]{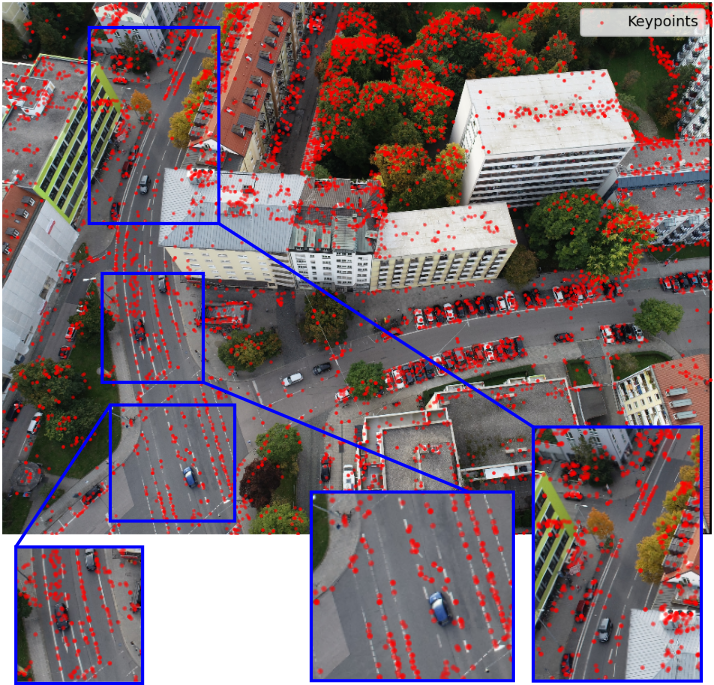}
\caption{\ac{dataset} (ours)}
\label{fig:our_dataset_accuracy}
\end{subfigure}
\caption{\textbf{Dataset Quality Assessment.} Geometric consistency evaluated by projecting 3D keypoints onto query images or showing alignment between \ac{DOP} from different sources. Our dataset shows superior consistency with accurate projections.}
\label{fig:accuracy_comparison_datasets}
\vspace{-1.0\baselineskip}
\end{figure}

As shown in Figure~\ref{fig:accuracy_comparison_datasets}, CrossLoc~\cite{crossloc_22} and UAVD4L~\cite{uavd4l_24} exhibit noticeable projection errors, indicating pose inaccuracies. While AnyVisLoc~\cite{anyvisloc_25} offers cross-domain augmentation using satellite imagery, this data is low resolution and poorly aligned with photogrammetry-based \acp{DOP}, limiting its suitability for realistic cross-domain experiments.

In contrast, our \ac{dataset} dataset delivers superior geometric consistency with accurately projected keypoints and well-aligned cross-domain data at resolutions comparable to official geoportal sources, enabling reliable cross-domain localization research.

\subsection{Dataset Creation Pipeline}
\label{appendix:dataset_pipeline}

The dataset is created by capturing data with drones, building 3D models through photogrammetry with georeferencing, extracting data like orthophotos and elevation rasters, and pairing query images with reference data followed by domain augmentation.

\subsubsection{Data Acquisition and Processing}
\label{appendix:data_acquisition}

\footnotetext{The geodata used to increase the domain shifts are ontained from different open geoportals. $4^{\text{th}}$ row: Geodata © Geobasis NRW, $5^{\text{th}}$ row: Geodata © HLBG, $6^{\text{th}}$ - $7^{\text{th}}$ rows: Geodata © LDBV Bayern}

\paragraph{Data collection.}
Our data collection encompassed 47 locations across 19 regions in Germany and the United States. We utilized a variety of commercial drones equipped with \ac{GPS} and \ac{RTK} technology to ensure precise positioning. To facilitate robust photogrammetric reconstruction, we implemented systematic flight paths, following established protocols in aerial mapping. 

Geodata were downloaded from public geoportals, with temporal misalignment considered explicitly. The dataset introduces time gaps of 2--8 years between geodata and \ac{UAV} imagery, matching common update cycles (2--3 years in urban areas, over 5 years in rural regions). This design provides a realistic benchmark for developing methods robust to outdated geodata.

\paragraph{Georeferenced 3D scene reconstruction.}
\label{appendix:reconstruction}

At each location, we acquire $N$ flight images $I_i$ ($1 \leq i \leq N$) and begin the pipeline by georeferencing them to constrain the subsequent \ac{SfM} optimization. We leverage \ac{GPS}, \ac{RTK}, or manually annotate \acp{GCP} to ensure accurate alignment. We execute a collection of \ac{SfM} pipelines---including DJI Terra~\cite{dji_terra}, PixPro~\cite{pixpro}, and COLMAP~\cite{colmap_16} with \ac{MVS}~\cite{openmvs_20}---and select the best output for each scene based on bundle-adjustment reprojection errors, GCP~\ac{RTK} residuals, and qualitative keypoint projections, following the procedure used in benchmarked localization methods. 

Formally, the pipeline extracts features from the images and constructs a pose graph. \ac{SfM} computes initial camera poses and a sparse point cloud $\mathcal{P}$, which \ac{MVS} densifies to refine poses and produce a denser 3D representation. We triangulate using Poisson surface reconstruction~\cite{poisson_2013} and apply texturing to generate a mesh with vertices $\mathcal{V} = \{\mathbf{P}_j\}_j$ and faces $\mathcal{F}$. 

\begin{figure}[t]
    \centering
    \includegraphics[width=1.0\linewidth]{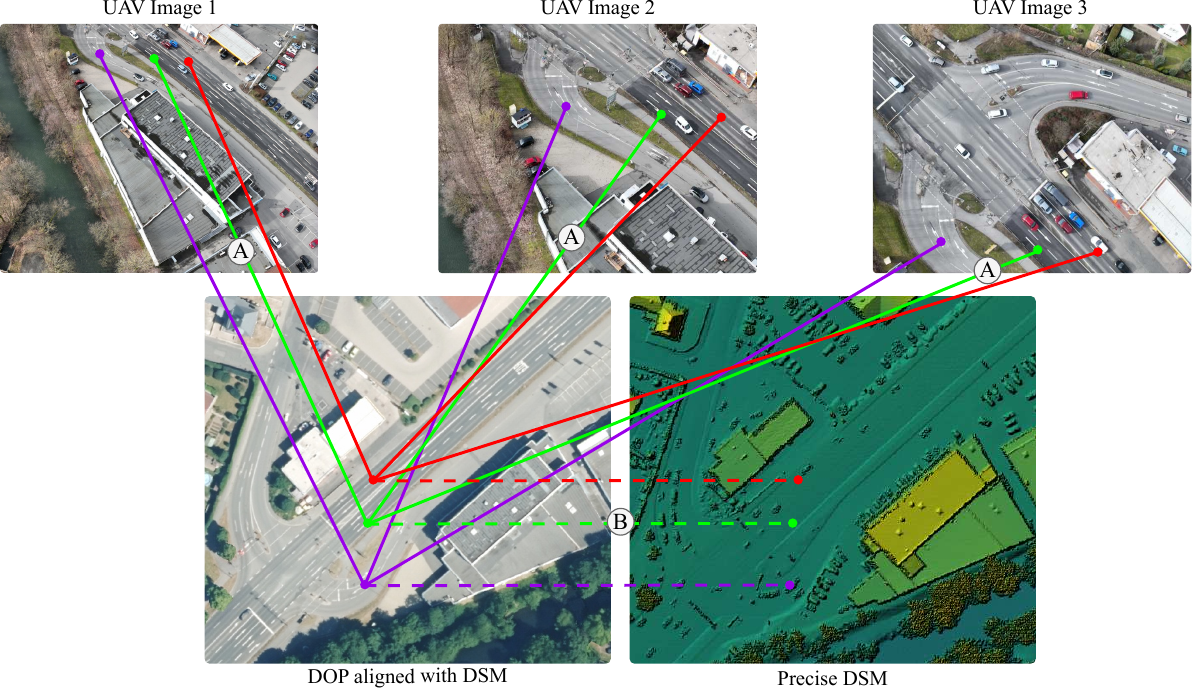}
\caption{\textbf{Manual \ac{GCP} Selection Procedure.} (A) We manually establish 2D–2D correspondences between the \ac{UAV} images and a \ac{DOP} (acquired from high-precision open geoportals), using visually distinctive features such as road markings. The feature positions are extracted in at least three images from the set of collected \ac{UAV} images. (B) The elevation of the corresponding pixel is then extracted to obtain 2D–3D correspondences.}
    \label{fig:gcps}
    \vspace{-1\baselineskip}
\end{figure}

We obtain precise 6-DoF ground-truth poses by jointly optimizing the scene geometry $\mathcal{P}$, camera extrinsics $\mathbf{T}_i$, and shared intrinsics $\mathbf{K}$, while incorporating georeferencing constraints from three complementary sources: \textbf{(1)} standard \ac{GPS} for coarse positioning, \textbf{(2)} \ac{RTK} for centimeter-level accuracy, and \textbf{(3)} manually annotated \acp{GCP} for high-precision alignment. Our \acp{GCP} were carefully selected following established best practices similar to those validated in~\cite{cledat2020mapping}. As shown in~\cref{fig:gcps}, we manually chose features with precise, visually distinctive characteristics, such as road marking edges, ensuring optimal visibility and spatial distribution across the mapping area. The corresponding 3D coordinates were obtained from either vehicle-based Mobile Laser Scanning point clouds or high-precision governmental geodata by sampling the \ac{DSM} at these locations. These 2D–3D correspondences are incorporated into the bundle adjustment during \ac{SfM} reconstruction as follows:
\begin{equation}
\begin{aligned} 
\mathbf{T}_i, \mathbf{K}, \mathcal{P} &= \argmin_{\mathbf{T}_i, \mathbf{K}, \mathcal{P}} \mathcal{L}_{\text{reproj}} + \lambda_{\text{GPS}} \mathcal{L}_{\text{GPS}} + \lambda_{\text{GCP}} \mathcal{L}_{\text{GCP}} \, ,\\
\mathcal{L}_{\text{reproj}} &= \sum_{i,j} \rho\left(  \|\pi_p(\mathbf{P}_j, \mathbf{K}, \mathbf{T}_i) - \mathbf{p}_{ij}\|_2 \right) \, ,\\
\mathcal{L}_{\text{GPS}} &= \sum_{i} \|\mathbf{C}_i - \mathbf{C}_i^{\text{GPS}}\|_2^2 \, ,\\
\mathcal{L}_{\text{GCP}} &= \sum_{k} \|\mathbf{P}_k - \mathbf{P}_k^{\text{GCP}}\|_2^2 \, .
\end{aligned}
\end{equation}
Here, $\mathbf{P}_j \in \mathbb{R}^3$ is a 3D scene point, $\mathbf{p}_{ij} \in \mathbb{R}^2$ is its projection in image $I_i$ through $\pi_p(\cdot)$, $\rho(\cdot)$ is a robust cost function, $\mathbf{C}_i \in \mathbb{R}^3$ is the camera center, and $\mathbf{C}_i^{\text{GPS}}$ is the measured \ac{GPS} or \ac{RTK} position. For ground control, $\mathbf{P}_k$ denotes a \ac{GCP} point, and $\mathbf{P}_k^{\text{GCP}}$ is its reference position. Weighting factors $\lambda_{\text{GPS}}$ and $\lambda_{\text{GCP}}$ are tuned to balance the reliability of each data source.
After optimization, the residual \ac{GCP} errors yield \ac{RMSE} values of 0.023 m, 0.030 m, and 0.042 m in x, y, and z, respectively, with an overall 3D \ac{RMSE} of 0.051 m.

While the optimization includes radial and tangential lens distortion coefficients, these terms are omitted from the equations for simplicity. The dataset provides undistorted images, allowing researchers to focus on focal length estimation, which is the most challenging aspect of \ac{UAV} camera calibration.

\paragraph{Rasterization and rendering.}
\label{appendix:rasterization}

The 3D mesh reconstruction is converted into two complementary geospatial representations: a \ac{DSM} matrix $\mathbf{R}^{\text{DSM}}$ and a \ac{DOP} matrix $\mathbf{R}^{\text{DOP}}$. The \ac{DSM} is generated by casting rays downward from a planar grid aligned with the XY plane at the maximum elevation. Each grid cell $(i,j)$ corresponds to a geographic position $(x,y)$, and the value at $\mathbf{R}^{\text{DSM}}(i,j)$ represents the highest elevation (z-coordinate) of the mesh intersected by the ray. Cells with no intersections are explicitly marked as invalid. The \ac{DOP} is created by rendering a nadir-oriented, orthographic view of the textured mesh with a virtual camera aligned along the negative z-axis. The resulting image is georeferenced and resampled to align with the resolution and coordinate system of the \ac{DSM}.

We also extract a sparse set $\mathcal{S}$ of SIFT~\cite{sift_2004} keypoints from the \ac{DOP}, which are lifted to 3D coordinates by mapping their positions to corresponding elevations in the \ac{DSM} using~\Cref{eq:dop_to_dsm}. These keypoints serve as reliable landmarks for pose verification in our evaluations.

\subsubsection{Data Pairing}
\label{appendix:data_pairing}

\paragraph{Selection of regions of interest.}
Given known camera parameters $(\mathbf{K}_i, \mathbf{T}_i)$ for each query image $I_i$, correspondences with geospatial representations are established through a precise geometric approach. To improve computational efficiency, the query image camera is downscaled by a factor of 8, and a grid of image coordinates ${\mathbf{p}_j^c}$ is generated for the downscaled image. For each coordinate, ray tracing is performed through the camera using $\mathbf{P}_j^c = \pi^{-1}(\mathbf{p}_j^c, \mathbf{K}_i, \mathbf{T}_i, d)$, where $\pi^{-1}(\cdot)$ is the inverse projection function and $d$ is the ray-mesh intersection depth.

Projected points with valid intersections ($d_j < \infty$) are filtered, and an irregular quadrilateral is fitted to the valid points. To introduce variability, stochastic perturbations are applied to the quadrilateral vertices as $\mathbf{P}_j^{c'} = \mathbf{P}_j^c + \boldsymbol{\epsilon}$, where $\boldsymbol{\epsilon} \sim \mathcal{U}(-20\text{m}, 20\text{m})$. The perturbed 3D points $\mathbf{P}_j^{c'}$ are then projected onto the \ac{DOP} and \ac{DSM} to define corresponding 2D regions, which are used to crop rasters $\mathbf{R}_i^{\text{DOP}}$ and $\mathbf{R}_i^{\text{DSM}}$ for each sample $i$.

To enrich data modalities, per-pixel raycasting generates point maps $\mathbf{P}_i \in \mathbb{R}^{W \times H \times 3}$, from which depth maps can derived using the extrinsic parameters. Additionally, visible mesh elements ($\mathcal{V}_i$, $\mathcal{F}_i$) and SIFT keypoints ($\mathcal{S}_i$) are also selected for each viewpoint, increasing the dataset’s modalities.

\paragraph{Data anonymization.}
To preserve geographic privacy while maintaining geometric relationships, we transform each data sample into a local coordinate system. For each sample, we apply a translation transform $\mathbf{v}_i \in \mathbb{R}^3$ defined by randomly selecting a finite 3D point from the visible scene. This translation is consistently applied to all geometric elements:

\begin{equation}
\begin{aligned}
\mathbf{P}_i' &= \mathbf{P}_i - \mathbf{v}_i\, , & \mathbf{R}_i' &= \mathbf{R}_i \, , & \mathbf{t}_i' &= \mathbf{t}_i + \mathbf{R}_i \mathbf{v}_i\, , \\
\text{DSM}' &= \text{DSM} - \mathbf{v}_{i,z} \, , & 
o_x' &= o_x - \mathbf{v}_{i,x} \, , & o_y' &= o_y - \mathbf{v}_{i,y} \, , \\
\mathcal{V}_i' &= \mathcal{V}_i - \mathbf{v}_i \, , & \mathcal{S}_i' &= \mathcal{S}_i - \mathbf{v}_i\, ,
\end{aligned}
\end{equation}

where $\mathbf{P}_i$ represents 3D points in the original point cloud, $\mathcal{V}_i$ and $\mathcal{S}_i$ denote visible points and scene points respectively, and $\mathbf{v}_{i,x}$, $\mathbf{v}_{i,y}$, and $\mathbf{v}_{i,z}$ stand for the x, y, and z coordinates of vector $\mathbf{v}_i$.

The camera transformation by translation is derived as follows: First, we convert the world-to-camera pose $\mathbf{T}_i$ (with rotation $\mathbf{R}_i$ and translation $\mathbf{t}_i$) to its inverse camera-to-world pose $\mathbf{T}_i^{-1}$ (with rotation $\mathbf{R}_i^\top$ and translation $-\mathbf{R}_i^\top \mathbf{t}_i$, which represents the camera center). After shifting this camera center by $\mathbf{v}_i$, we obtain a new camera-to-world pose with rotation $\mathbf{R}_i^\top$ and translation $-\mathbf{R}_i^\top \mathbf{t}_i - \mathbf{v}_i$. Converting back to the world-to-camera frame yields the original rotation matrix $(\mathbf{R}_i^\top)^\top = \mathbf{R}_i$ and a new translation vector $-\mathbf{R}^T (-\mathbf{R}_i^\top \mathbf{t}_i - \mathbf{v}_i) = \mathbf{t}_i + \mathbf{R}_i \mathbf{v}_i$.

This transformation suppresses absolute georeferencing while preserving all relative geometric relationships essential for localization evaluation. The random selection of transformation vectors ensures that geographic coordinates cannot be reliably reconstructed from the published dataset, protecting sensitive location information.

\section{Additional Experimental Details and Results}
\label{appendix:experiment_settings_results}

\subsection{Experiment Setting}
\label{appendix:experiment_settings}

Our experiments were conducted on a cluster using a computation node equipped with an Intel(R) Xeon(R) Gold 6254 CPU @ 3.10GHz and a single Quadro RTX 8000 GPU with 48GB memory. 

For benchmarking, we evaluated algorithms using the provided model weights without fine-tuning to assess inherent robustness. Images were resized to meet each algorithm's requirements, with resulting coordinates transformed back to full resolution before 3D lifting. Rotation-invariant algorithms processed each image in four orientations, selecting the output with the largest correspondence set.

The 6-DoF pose estimation was performed using PnP~\cite{review_pnp_94} with LO-RANSAC~\cite{lo_ransac_03} for outlier rejection, applying a 5-pixel reprojection threshold for inlier selection. For algorithms providing confidence scores, a 0.5 threshold was used to pre-filter correspondences. Optimization was restricted to estimating focal length, assuming a fixed aspect ratio and principal point at the image center, based on empirical validation.

\subsection{Qualitative Assessment of \acs{AdHoP} Performance}
\label{appendix:results_adhop}

This section offers a detailed qualitative evaluation of the \ac{AdHoP} strategy in various scenarios. While quantitative results in the main paper highlight consistent improvements in localization accuracy, visual analysis of reprojected keypoints provides further insights into the strengths and limitations of the method.

\begin{figure}[htbp]
    \centering
    \resizebox{\linewidth}{!}{
    \setlength{\tabcolsep}{1pt}
    \begin{tabular}{ccccc}
    \multicolumn{2}{c}{Improvement Cases} & \multicolumn{2}{c}{Degradation / Failure Cases} \\
    w/o \ac{AdHoP} & w \ac{AdHoP} & w/o \ac{AdHoP} & w \ac{AdHoP} \\
    \rotatebox{90}{\scriptsize SP+SG~\cite{superpoint_18,superglue_20}}
    \includegraphics[width=0.25\linewidth]{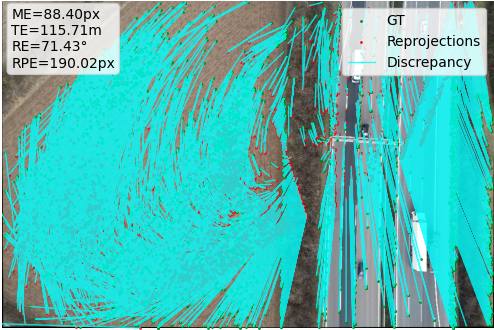} & 
    \includegraphics[width=0.25\linewidth]{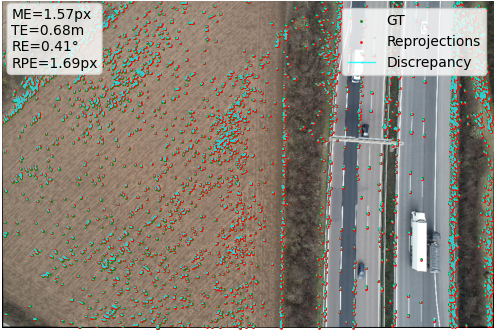} & 
    \includegraphics[width=0.25\linewidth]{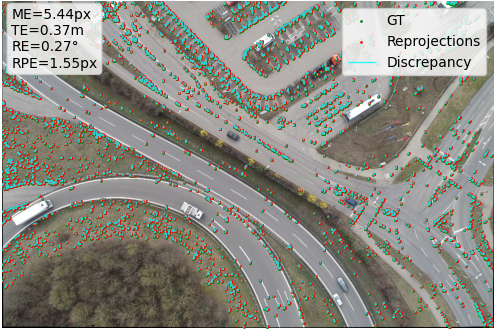} & 
    \includegraphics[width=0.25\linewidth]{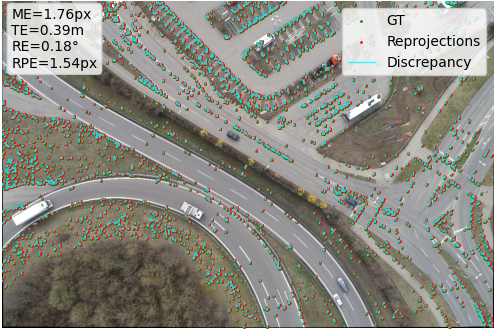} \\
\multicolumn{2}{c}{\scriptsize Sample: L50\_xDOP0031} & \multicolumn{2}{c}{\scriptsize Sample: L50\_xDOP0019e} \\
 
    \rotatebox{90}{\scriptsize DeDoDe~\cite{dedode_24}}
    \includegraphics[width=0.25\linewidth]{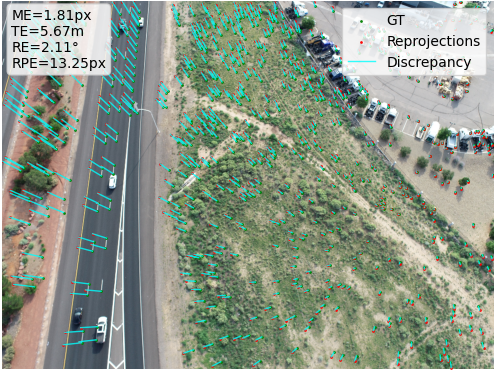} & 
    \includegraphics[width=0.25\linewidth]{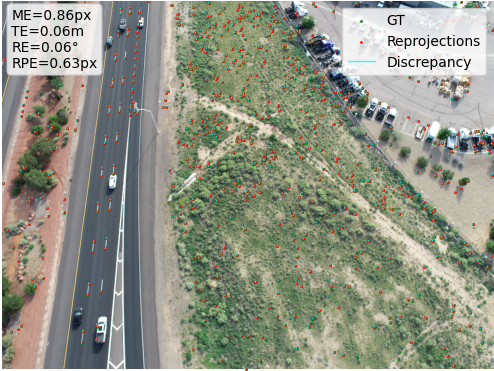} & 
    \includegraphics[width=0.25\linewidth]{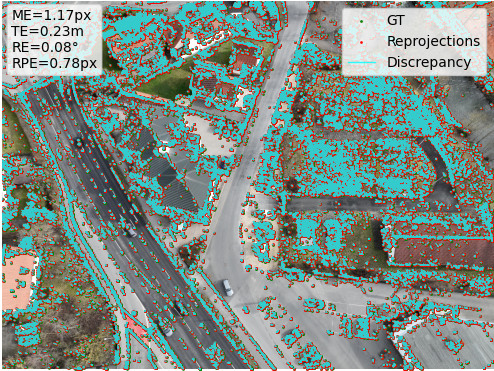} & 
    \includegraphics[width=0.25\linewidth]{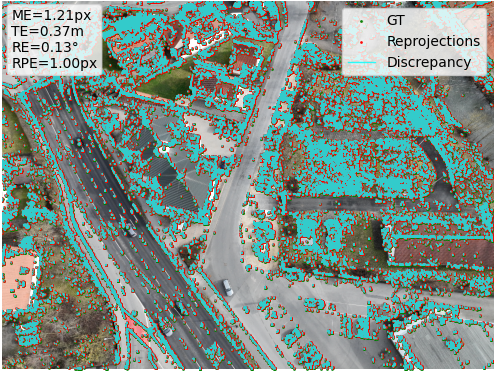} \\
\multicolumn{2}{c}{\scriptsize Sample: L01\_R0384}  & \multicolumn{2}{c}{\scriptsize Sample: L08\_R0228} \\

    \rotatebox{90}{\scriptsize LoFTR~\cite{loftr_21}}
    \includegraphics[width=0.25\linewidth]{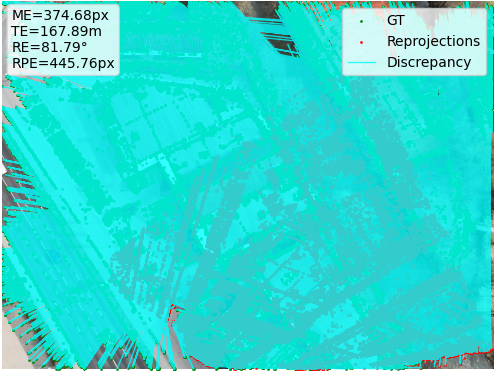} & 
    \includegraphics[width=0.25\linewidth]{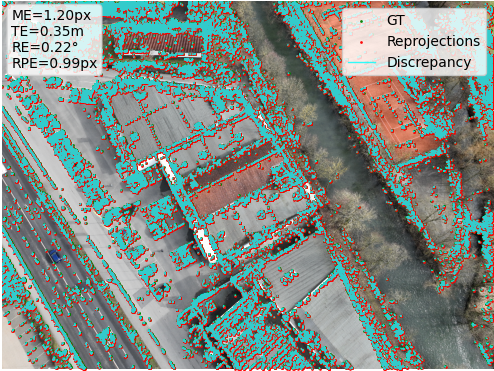} & 
    \includegraphics[width=0.25\linewidth]{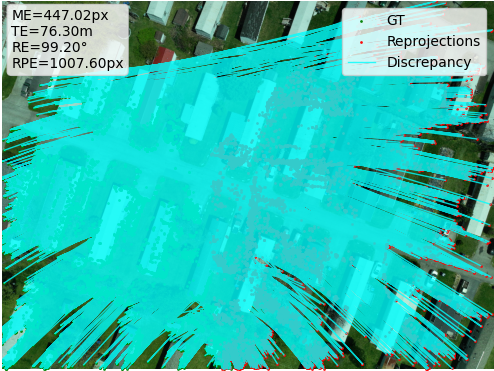} & 
    \includegraphics[width=0.25\linewidth]{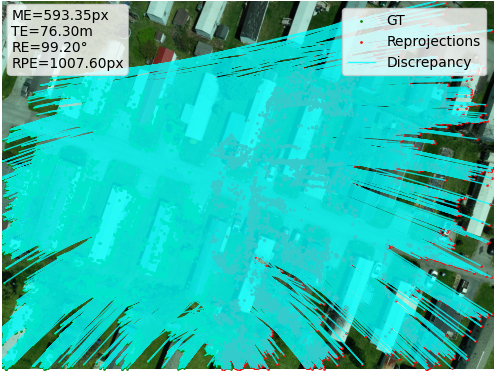} \\
\multicolumn{2}{c}{\scriptsize Sample: L08\_R0303} & \multicolumn{2}{c}{\scriptsize Sample: L02\_R0069} \\

    \rotatebox{90}{\scriptsize XFeat*~\cite{xfeat_24}}
    \includegraphics[width=0.25\linewidth]{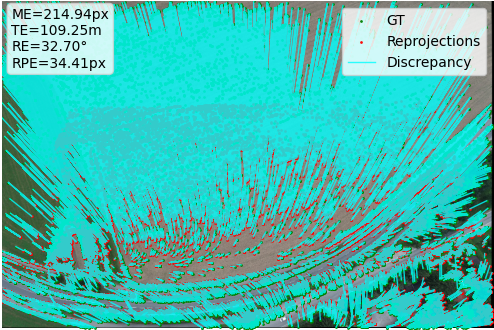} & 
    \includegraphics[width=0.25\linewidth]{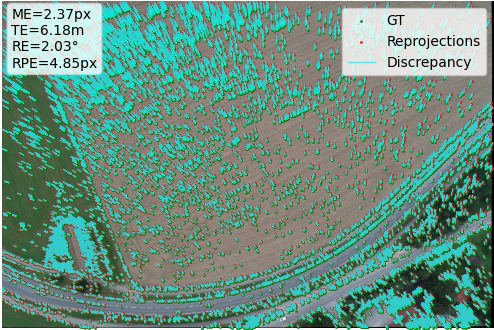} & 
    \includegraphics[width=0.25\linewidth]{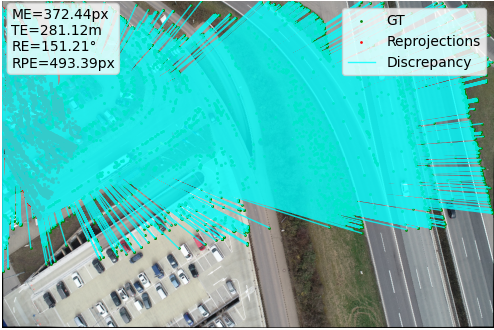} & 
    \includegraphics[width=0.25\linewidth]{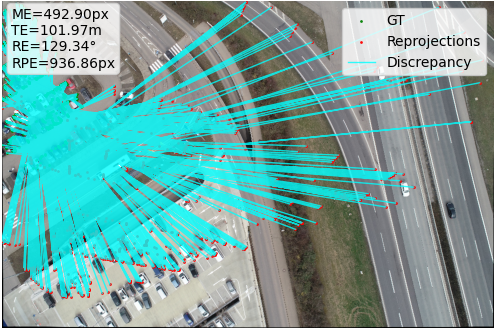} \\
\multicolumn{2}{c}{\scriptsize Sample: L51\_xDOP0019} & \multicolumn{2}{c}{\scriptsize Sample: L50\_xDOP0038} \\
    
    \rotatebox{90}{\scriptsize GIM+DKM~\cite{gim_24,dkm_23}}
    \includegraphics[width=0.25\linewidth]{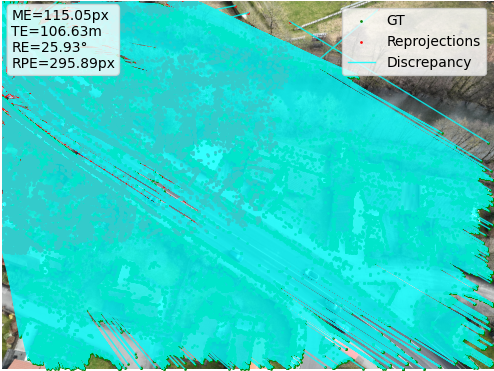} & 
    \includegraphics[width=0.25\linewidth]{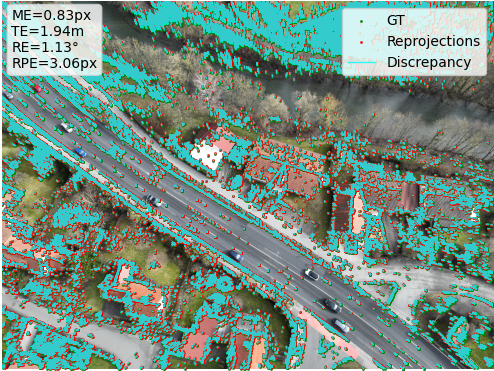} & 
    \includegraphics[width=0.25\linewidth]{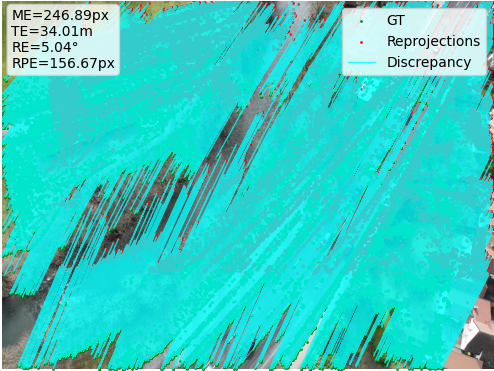} & 
    \includegraphics[width=0.25\linewidth]{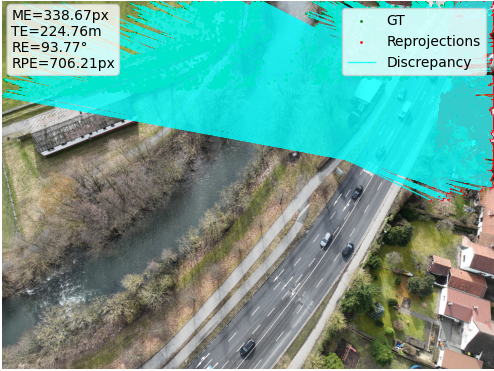} \\
\multicolumn{2}{c}{\scriptsize Sample: L08\_xDOPDSM0242} & \multicolumn{2}{c}{\scriptsize Sample: L08\_xDOPDSM0263} \\
    
    \rotatebox{90}{\scriptsize MASt3R~\cite{mast3r_24}}
    \includegraphics[width=0.25\linewidth]{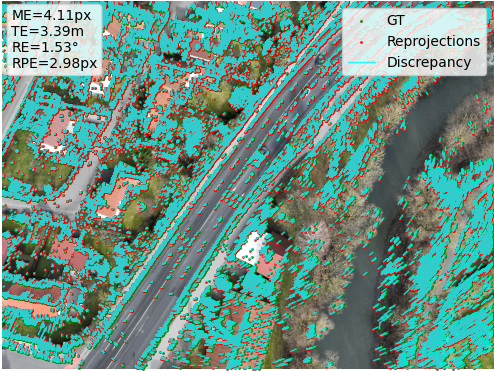} & 
    \includegraphics[width=0.25\linewidth]{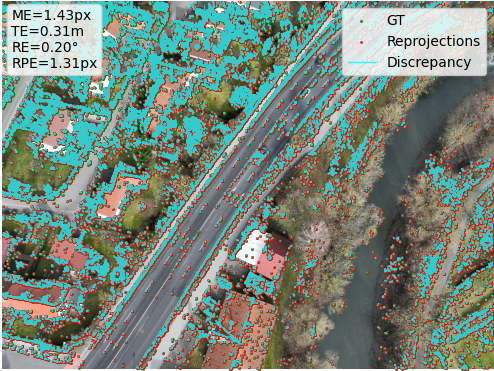} & 
    \includegraphics[width=0.25\linewidth]{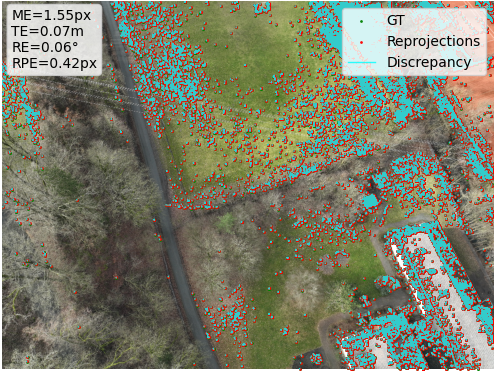} & 
    \includegraphics[width=0.25\linewidth]{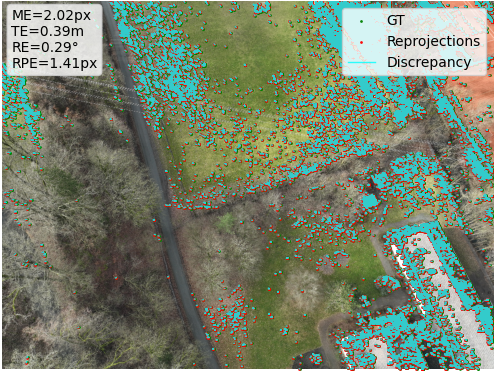} \\
\multicolumn{2}{c}{\scriptsize Sample: L08\_xDOP0350} & \multicolumn{2}{c}{\scriptsize Sample: L08\_R0145} \\
    
    \rotatebox{90}{\scriptsize RoMa~\cite{roma_24}}
    \includegraphics[width=0.25\linewidth]{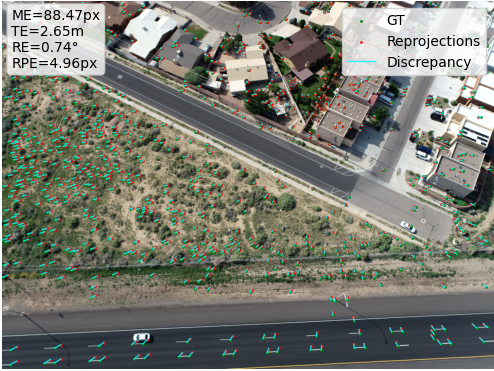} & 
    \includegraphics[width=0.25\linewidth]{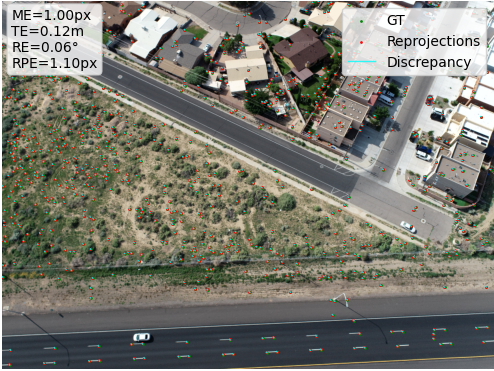} & 
    \includegraphics[width=0.25\linewidth]{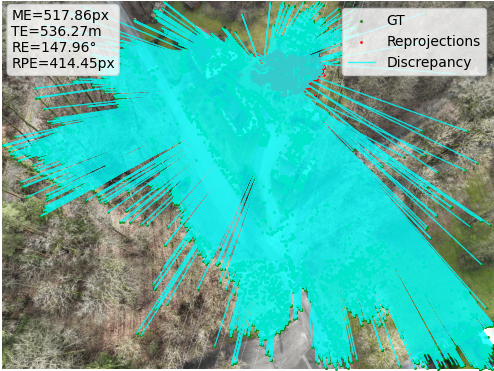} & 
    \includegraphics[width=0.25\linewidth]{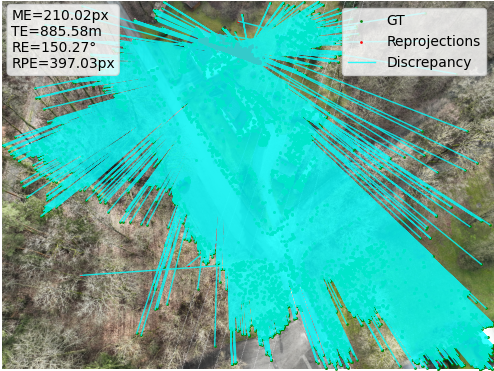} \\
\multicolumn{2}{c}{\scriptsize Sample: L01\_R0531} & \multicolumn{2}{c}{\scriptsize Sample: L08\_xDOP0153} \\

    \end{tabular}
    }
    \caption{\textbf{Qualitative Results of the Localization Using Our Baseline Method.} The left side illustrates successful improvements achieved using \ac{AdHoP}, while the right side presents degenerate or failure cases. \textcolor{darkgreen}{Green} and \textcolor{red}{red} points denote projections of the 3D keypoints $\mathcal{S}_i$ using the ground-truth and estimated poses, respectively. The \textcolor{cyan}{blue} lines indicate the discrepancies between these projections.}
    \label{fig:loc_reprojs}
\end{figure}

The left column of \Cref{fig:loc_reprojs} presents examples of challenging scenarios where \ac{AdHoP} achieves successful localization. These include cases where \ac{AdHoP} reduces large initial errors to achieve highly accurate poses with minimal reprojection error, enhances moderately accurate estimates to near-perfect precision, and improves the spatial distribution of correspondences across the image plane, leading to better geometric consistency.

The right column highlights instances where \ac{AdHoP} struggles to deliver improvements. In some cases, it slightly worsens performance by increasing matching errors or producing less accurate poses, often due to correspondences with worse geometric cues. In other scenarios, the approach fails to improve poorly initialized calibrations, as the homography-based warping introduces distortions that prevent effective matching. These issues are typically observed under extreme domain shifts, highly repetitive patterns, or significant discrepancies between the reference geodata and query images, where existing matchers exhibit poor performance. Notably, the failure modes of \ac{AdHoP} are captured in the computed projection error, allowing automatic rejection of results when the reprojection error increases.

\begin{figure}[htbp]
\centering
\includegraphics[width=1\textwidth]{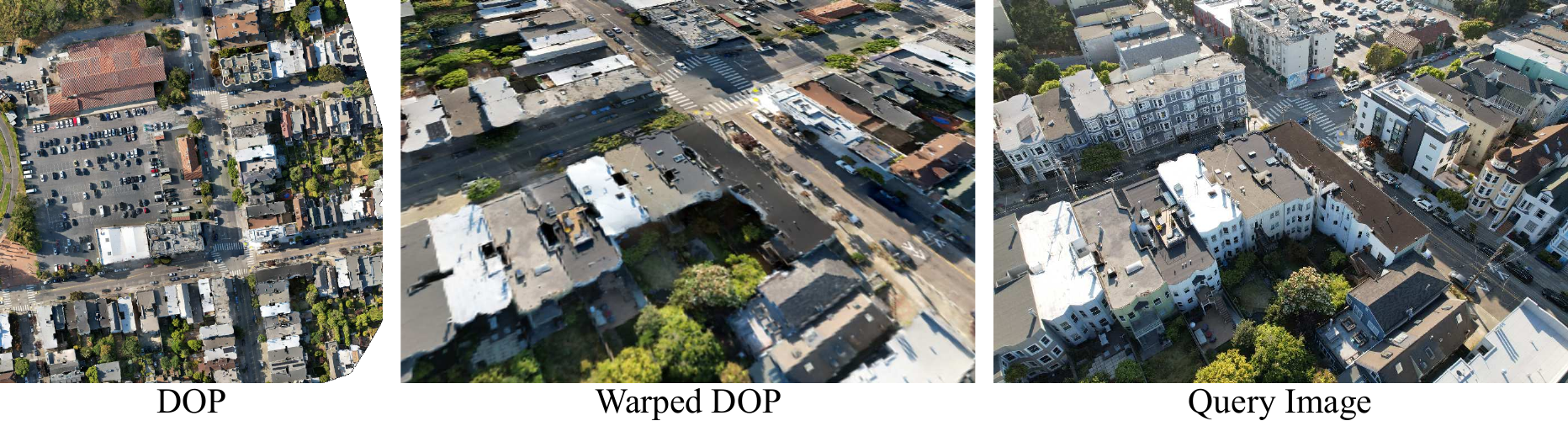}
\caption{\textbf{Effectiveness of Warped \ac{DOP} in Addressing Domain Shifts.} \textbf{Left:} The original \ac{DOP} exhibits significant discrepancies with \ac{UAV} imagery due to temporal changes, lighting variations, and pronounced viewpoint differences. \textbf{Middle:} The warped \ac{DOP} after applying the computed transformation demonstrates substantially improved alignment with \ac{UAV} imagery. \textbf{Right:} The corresponding \ac{UAV} query image to match.}
\label{fig:warping_of_dops}
\end{figure}

We also present an example of warped \ac{DOP} in \Cref{fig:warping_of_dops}, demonstrating how this transformation significantly improves alignment between the geodata and \ac{UAV} imagery by reducing domain shifts in appearance. The most accurately warped regions correspond to planar and non-occluded areas. However, some domain shifts persist, primarily in regions where: (1) building facades are missing in the \acp{DOP}, (2) areas occluded in the orthographic projection, (3) regions with strong shadows, and (4) inherent appearance differences in \ac{DOP} from different capture times.

This qualitative analysis supports our quantitative results, showing that \ac{AdHoP} improves accuracy in most cases.

\subsection{Camera Calibration Results}
\label{appendix:calibration_results}

\begin{table}[htpb]
  \centering
   \caption{\textbf{Quantitative Calibration Results on \ac{dataset} Test Sets.} Rankings between matchers are highlighted as \colorbox{colorFst}{first}, \colorbox{colorSnd}{second}, and \colorbox{colorTrd}{third}. \textbf{Bold} values indicate the best performance comparing without/with \ac{AdHoP}. RI indicates a rotation-invariant matcher (matching performed with 4 rotated versions, selecting the one with most correspondences). Abbreviations: SuperPoint~(SP), SuperGlue~(SG), LightGlue~(LG), Minima~(MM).}
  \label{tab:cal_w_wo_adhop_results}
  \scriptsize
  \setlength{\tabcolsep}{2pt}
  
\resizebox{\linewidth}{!}{\
\renewcommand{\arraystretch}{1.1}
  \begin{tabular}{c!{\vrule width 1pt}c
                  !{\vrule width 1pt}c|c|ccc|ccc|c
                  }
    \Xhline{1pt} \rowcolor{gray!15}
      \multirow{1}{*}{\textbf{Matcher}}
      & \multirow{1}{*}{\textbf{RI}} 
      & \textbf{ME} [px]$\downarrow$ & \textbf{RFE} [\%]$\downarrow$ & \textbf{TE} [m]$\downarrow$ & \textbf{RE} [°]$\downarrow$ & \textbf{RPE} [px]$\downarrow$ & \textbf{1m-1°} [\%]$\uparrow$ & \textbf{3m-3°} [\%]$\uparrow$ & \textbf{5m-5°} [\%]$\uparrow$ & \textbf{Speed} [s]$\downarrow$ \\
    \Xhline{1pt}

    SP+SG~\cite{superpoint_18,superglue_20} & \xmark & {\bf 2.2} / {\bf 2.2} & 2.6 / {\bf 2.4} & {\rd 3.09}  / {\bf 2.93} & {\nd 0.60}  / {\bf 0.57} & {\nd 16.4}  / {\bf 15.9} & {\bf {\rd 12.1} } / 11.1 & {\rd 40.0}  / {\bf {\rd 41.2} } & {\rd 50.7}  / {\bf {\rd 52.7} } & {\bf {\nd 0.2} } / {\nd 0.3}  \\ 
SP+LG~\cite{superpoint_18,lightglue_23} & \xmark & {\bf {\rd 2.0} } / {\bf {\rd 2.0} } & {\fs 1.9}  / {\bf {\rd 1.8} } & {\fs 2.49}  / {\bf {\rd 2.30} } & {\fs 0.54}  / {\bf {\rd 0.52} } & {\fs 15.2}  / {\bf {\rd 14.3} } & {\bf {\nd 13.1} } / {\bf {\nd 13.1} } & {\nd 46.1}  / {\bf {\nd 48.1} } & {\nd 54.6}  / {\bf {\nd 57.1} } & {\bf {\fs 0.1} } / {\fs 0.2}  \\ 
DeDoDe~\cite{dedode_24} & \xmark & {\bf {\fs 1.2} } / {\bf {\fs 1.2} } & {\rd 2.5}  / {\bf {\nd 1.7} } & 3.40 / {\bf {\nd 2.11} } & {\nd 0.60}  / {\bf {\fs 0.42} } & {\rd 17.0}  / {\bf {\fs 12.2} } & 5.3 / {\bf 7.4} & 17.7 / {\bf 21.7} & 21.4 / {\bf 25.0} & {\bf {\rd 0.3} } / {\bf {\nd 0.3} } \\ 
XFeat~\cite{xfeat_24} & \xmark & 257.0 / {\bf 60.5} & 100.0 / {\bf 60.1} & {\bf 180.53} / 192.47 & {\bf 169.09} / 173.66 & {\bf 648.8} / 809.0 & 0.0 / {\bf 0.1} & 0.0 / {\bf 1.6} & 0.0 / {\bf 2.5} & {\bf {\fs 0.1} } / {\nd 0.3}  \\ 
XFeat+LG~\cite{xfeat_24,lightglue_23} & \xmark & 4.3 / {\bf 3.2} & 6.1 / {\bf 2.7} & 7.40 / {\bf 3.35} & 1.06 / {\bf 0.73} & 29.1 / {\bf 19.5} & 4.4 / {\bf 6.6} & 17.4 / {\bf 30.7} & 26.0 / {\bf 38.5} & {\bf {\fs 0.1} } / {\nd 0.3}  \\ 
\Xhline{0.1pt}

LoFTR~\cite{loftr_21} & \xmark & 317.2 / {\bf 308.7} & 93.2 / {\bf 83.5} & 170.00 / {\bf 154.66} & 129.88 / {\bf 112.92} & 914.2 / {\bf 899.4} & 0.4 / {\bf 5.0} & 1.9 / {\bf 14.2} & 2.7 / {\bf 15.5} & {\bf {\fs 0.1} } / {\nd 0.3}  \\ 
MM+LoFTR~\cite{minima_24,lodloc_24} & \xmark & {\bf 261.1} / {\bf 261.1} & 86.0 / {\bf 54.2} & 193.17 / {\bf 143.00} & 112.46 / {\bf 91.81} & 1035.7 / {\bf 950.4} & 0.1 / {\bf 5.3} & 0.9 / {\bf 15.0} & 1.6 / {\bf 17.2} & {\bf {\rd 0.3} } / {\rd 0.6}  \\ 
eLoFTR~\cite{eloftr_24} & \xmark & 328.8 / {\bf 315.0} & 96.0 / {\bf 80.7} & 188.04 / {\bf 160.78} & 128.66 / {\bf 107.91} & 868.8 / {\bf 858.1} & 0.6 / {\bf 5.3} & 1.8 / {\bf 16.0} & 2.6 / {\bf 17.6} & {\bf {\fs 0.1} } / {\fs 0.2}  \\ 
XoFTR~\cite{xoftr_24} & \xmark & 286.6 / {\bf 282.2} & 91.9 / {\bf 71.6} & 180.41 / {\bf 146.27} & 135.07 / {\bf 107.65} & 942.8 / {\bf 829.1} & 0.3 / {\bf 4.1} & 1.9 / {\bf 12.6} & 2.5 / {\bf 14.5} & {\bf {\fs 0.1} } / {\fs 0.2}  \\ 
\Xhline{0.1pt}

DKM~\cite{dkm_23} & \cmark & 49.0 / {\bf 2.8} & 50.5 / {\bf 9.3} & 131.70 / {\bf 32.22} & 108.41 / {\bf 4.19} & 818.4 / {\bf 114.2} & 2.8 / {\bf {\rd 12.3} } & 12.4 / {\bf 30.0} & 17.5 / {\bf 34.8} & {\bf 0.8} / 1.7 \\ 
XFeat*~\cite{xfeat_24} & \xmark & 222.2 / {\bf 10.2} & 100.0 / {\bf 52.6} & 184.95 / {\bf 184.39} & {\bf 170.90} / 173.72 & {\bf 660.9} / 789.9 & 0.0 / {\bf 0.2} & 0.0 / {\bf 2.8} & 0.0 / {\bf 4.0} & {\bf {\fs 0.1} } / {\fs 0.2}  \\ 
GIM+DKM~\cite{gim_24,dkm_23} & \cmark & {\nd 1.5}  / {\bf {\nd 1.4} } & {\nd 2.4}  / {\bf {\fs 1.6} } & {\nd 3.07}  / {\bf {\fs 2.09} } & {\rd 0.65}  / {\bf {\nd 0.47} } & 17.8 / {\bf {\nd 13.1} } & {\fs 16.2}  / {\bf {\fs 21.8} } & {\fs 49.5}  / {\bf {\fs 59.0} } & {\fs 58.2}  / {\bf {\fs 67.8} } & {\bf 1.3} / 2.6 \\ 
DUSt3R~\cite{dust3r_24} & \cmark & 5.0 / {\bf 4.9} & 12.5 / {\bf 12.3} & {\bf 14.68} / 17.02 & {\bf 2.39} / 2.82 & {\bf 75.3} / 86.9 & {\bf 0.3} / 0.2 & {\bf 5.7} / 5.5 & {\bf 12.7} / 11.4 & {\bf 1.5} / 2.1 \\ 
MASt3R~\cite{mast3r_24} & \cmark & 2.4 / {\bf 2.3} & 4.9 / {\bf 3.9} & 6.10 / {\bf 4.72} & 0.95 / {\bf 0.79} & 28.7 / {\bf 23.3} & 6.0 / {\bf 8.4} & 31.1 / {\bf 38.0} & 45.3 / {\bf 51.6} & {\bf 2.2} / 3.3 \\ 
RoMa~\cite{roma_24} & \cmark & 20.8 / {\bf 2.5} & 91.9 / {\bf 7.0} & 150.82 / {\bf 10.62} & 149.98 / {\bf 1.73} & 616.0 / {\bf 46.2} & 1.1 / {\bf 5.2} & 8.0 / {\bf 27.6} & 12.5 / {\bf 38.8} & {\bf 1.1} / 2.1 \\ 
MM+RoMa~\cite{minima_24,roma_24} & \cmark & 71.1 / {\bf 4.3} & 99.3 / {\bf 13.5} & 165.60 / {\bf 26.39} & 142.71 / {\bf 3.70} & 646.2 / {\bf 97.7} & 1.1 / {\bf 3.3} & 7.1 / {\bf 20.7} & 11.4 / {\bf 30.5} & {\bf 1.1} / 2.2 \\ 
\Xhline{1pt}

  \end{tabular}
  }
\end{table}

\Cref{tab:cal_w_wo_adhop_results} presents our full camera calibration results across different feature matching approaches, both without and with our proposed \ac{AdHoP} strategy. The integration of \ac{AdHoP} significantly enhances calibration performance across all matchers, with GIM+DKM~\cite{gim_24, dkm_23} achieving the best focal length estimation with an \ac{RFE} of 1.6\%. The most dramatic improvement is observed with RoMa~\cite{roma_24}, where translation errors are reduced from 150.82\,m to 10.62\,m.

Our experimental results show a strong correlation between focal length errors and translation errors, driven by the inherent mathematical ambiguity of the perspective projection model. In the next section, we provide a mathematical proof of the correlation between focal length and the camera translation.

\subsection{Proof of Focal Length and Translation Ambiguity}
We denote a 3D point as $\mathbf{P} = [X, Y, Z]^\top$ and its corresponding 2D image point as $\mathbf{p} = [p_x, p_y]^\top$. For simplicity, we assume that the focal lengths in the intrinsic matrix $\mathbf{K}$ are equal, i.e., $f_x = f_y = f$, which is a reasonable assumption for most modern cameras. Let $\mathbf{r}_1, \mathbf{r}_2, \mathbf{r}_3$ denote the three rows of the rotation matrix $\mathbf{R}$.

Using the projection model defined in \Cref{eq:general_projection}, the projection of a single 3D point onto the image plane is expressed as:
\begin{equation}
p_x = f \cdot \frac{\mathbf{r}_1 \cdot \mathbf{P} + t_x}{\mathbf{r}_3 \cdot \mathbf{P} + t_z} + c_x\, ,
\end{equation}
\begin{equation}
p_y = f \cdot \frac{\mathbf{r}_2 \cdot \mathbf{P} + t_y}{\mathbf{r}_3 \cdot \mathbf{P} + t_z} + c_y\, ,
\end{equation}

where $\mathbf{r}_i \cdot \mathbf{P}$ represents the dot product between the $i$-th row of $\mathbf{R}$ and the 3D point $\mathbf{P}$.

For the reprojection error, we use the Mean Squared Error (MSE) for this proof. Given a set of $N$ corresponding 2D-3D point pairs $\{(\mathbf{p}_i, \mathbf{P}_i)\}_{i=1}^N$:

\begin{equation}
E_{\text{reproj}} = \frac{1}{N}\sum_{i=1}^N \|\mathbf{p}_i - \hat{\mathbf{p}}_i\|^2 = \frac{1}{N}\sum_{i=1}^N \left( (p_{x,i} - \hat{p}_{x,i})^2 + (p_{y,i} - \hat{p}_{y,i})^2 \right)\, ,
\end{equation}

where $\hat{\mathbf{p}}_i = [\hat{p}_{x,i}, \hat{p}_{y,i}]^\top$ is the projection of $\mathbf{P}_i$ using the estimated camera parameters.

For notation simplicity, let:
\begin{equation}
\begin{aligned}
a_i &= \mathbf{r}_1 \cdot \mathbf{P}_i + t_x\, , \\
b_i &= \mathbf{r}_2 \cdot \mathbf{P}_i + t_y\, , \\
c_i &= \mathbf{r}_3 \cdot \mathbf{P}_i + t_z\, .
\end{aligned}
\end{equation}

Then the projected points become:
\begin{equation}
\hat{p}_{x,i} = f \cdot \frac{a_i}{c_i} + c_x, \quad \hat{p}_{y,i} = f \cdot \frac{b_i}{c_i} + c_y \, .
\end{equation}

The partial derivative of $\hat{p}_{x,i}$ with respect to $f$ is:
\begin{equation}
\frac{\partial \hat{p}_{x,i}}{\partial f} = \frac{a_i}{c_i} \, .
\end{equation}

Similarly for $\hat{p}_{y,i}$:
\begin{equation}
\frac{\partial \hat{p}_{y,i}}{\partial f} = \frac{b_i}{c_i} \, .
\end{equation}

The partial derivative of the reprojection error with respect to $f$ is:
\begin{equation}
\begin{aligned}
\frac{\partial E_{\text{reproj}}}{\partial f} &= \frac{2}{N}\sum_{i=1}^N \left[ (p_{x,i} - \hat{p}_{x,i}) \cdot \left(-\frac{a_i}{c_i}\right) + (p_{y,i} - \hat{p}_{y,i}) \cdot \left(-\frac{b_i}{c_i}\right) \right] \\
&= -\frac{2}{N}\sum_{i=1}^N \left[ (p_{x,i} - \hat{p}_{x,i}) \cdot \frac{a_i}{c_i} + (p_{y,i} - \hat{p}_{y,i}) \cdot \frac{b_i}{c_i} \right]
\end{aligned} \, .
\end{equation}

\paragraph{Mathematical approximation for $\frac{\partial E_{\text{reproj}}}{\partial f} \propto \frac{1}{f}$.}
Near the optimal solution, the reprojection errors $(p_{x,i} - \hat{p}_{x,i})$ and $(p_{y,i} - \hat{p}_{y,i})$ become very small. To understand the relationship with $f$, consider that at near-optimal parameters, we can approximate $p_{x,i} \approx \hat{p}_{x,i}$ and use small perturbations $\delta f$ around the current estimate of $f$. The change in projected coordinates would be:

\begin{equation}
\delta \hat{p}_{x,i} = \delta f \cdot \frac{a_i}{c_i} \, .
\end{equation}

For identical relative changes in focal length ($\frac{\delta f}{f}$), the absolute change in projection is proportional to $f$, as:

\begin{equation}
\delta \hat{p}_{x,i} = f \cdot \frac{\delta f}{f} \cdot \frac{a_i}{c_i} \, .
\end{equation}

This means the sensitivity of the projection (and consequently the error gradient) to absolute changes in $f$ scales with $\frac{1}{f}$. For larger $f$ values, the same absolute change has less impact on the projection. Therefore, $\frac{\partial E_{\text{reproj}}}{\partial f} \propto \frac{1}{f}$.

For the translation component $t_z$, we compute:
\begin{equation}
\frac{\partial \hat{p}_{x,i}}{\partial t_z} = -f \cdot \frac{a_i}{c_i^2} \, .
\end{equation}

Similarly:
\begin{equation}
\frac{\partial \hat{p}_{y,i}}{\partial t_z} = -f \cdot \frac{b_i}{c_i^2} \, .
\end{equation}

The partial derivative of the reprojection error with respect to $t_z$ is:
\begin{equation}
\begin{aligned}
\frac{\partial E_{\text{reproj}}}{\partial t_z} &= \frac{2}{N}\sum_{i=1}^N \left[ (p_{x,i} - \hat{p}_{x,i}) \cdot \left(f \cdot \frac{a_i}{c_i^2}\right) + (p_{y,i} - \hat{p}_{y,i}) \cdot \left(f \cdot \frac{b_i}{c_i^2}\right) \right] \\
&= \frac{2f}{N}\sum_{i=1}^N \left[ (p_{x,i} - \hat{p}_{x,i}) \cdot \frac{a_i}{c_i^2} + (p_{y,i} - \hat{p}_{y,i}) \cdot \frac{b_i}{c_i^2} \right]
\end{aligned} \, .
\end{equation}

\paragraph{Proving that $f$ and $t_z$ are coupled.}
In aerial imagery, $t_z$ is typically much larger than the variations in scene depth, so $c_i \approx t_z$ for most points. With this approximation:

\begin{equation}
\frac{\partial E_{\text{reproj}}}{\partial t_z} \propto -\frac{f}{t_z^2} \, .
\end{equation}

The critical insight comes from examining how $f$ and $t_z$ affect the projection. Consider a simplified projection model with $c_i \approx t_z$:

\begin{equation}
\hat{p}_{x,i} \approx f \cdot \frac{a_i}{t_z} + c_x \, .
\end{equation}

If we simultaneously scale $f$ by a factor $\alpha$ and $t_z$ by the same factor $\alpha$, the projection remains unchanged:

\begin{equation}
(\alpha f) \cdot \frac{a_i}{(\alpha t_z)} + c_x = f \cdot \frac{a_i}{t_z} + c_x \, .
\end{equation}

This exact mathematical compensation creates a "valley" in the optimization landscape where different combinations of $f$ and $t_z$ produce nearly identical reprojection errors, making their individual values ambiguous. 

For comparison, the partial derivative with respect to $t_x$ is:
\begin{equation}
\frac{\partial \hat{p}_{x,i}}{\partial t_x} = f \cdot \frac{1}{c_i} \, .
\end{equation}

Comparing these derivatives reveals the key relationships:
\begin{equation}
\frac{\partial E_{\text{reproj}}}{\partial f} \propto \frac{1}{f}, \quad \frac{\partial E_{\text{reproj}}}{\partial t_z} \propto -\frac{f}{t_z^2}, \quad \frac{\partial \hat{p}_{x,i}}{\partial t_x} \propto \frac{f}{c_i} \, .
\end{equation}

\paragraph{Importance of data variation for robust estimation.}
The coupling between $f$ and $t_z$ creates an ill-posed optimization problem when 3D points lie approximately on a plane, as is common in aerial imagery. Spatial diversity in point correspondences is crucial for breaking this ambiguity for two key reasons:
\begin{itemize}
    \item \textbf{Depth variation}: Points at different depths create different sensitivity patterns to $f$ and $t_z$. When the scene contains significant depth variations, the exact compensation relationship between $f$ and $t_z$ breaks down, as the $c_i = \mathbf{r}_3 \cdot \mathbf{P}_i + t_z$ term varies more significantly across points.
    \item \textbf{Geometric constraints}: Points distributed across the image plane, especially toward the borders, experience different projection behaviors than points clustered in the center. The peripheral points are more sensitive to focal length changes, providing stronger constraints during optimization.
\end{itemize}

Our \ac{AdHoP} strategy specifically addresses this challenge by encouraging spatially diverse correspondence distributions across the image plane. By ensuring correspondences span different image regions with varying depths, we better constrain the parameter space and reduce the inherent focal length-translation ambiguity, making the optimization more likely to converge to the correct parameter values.
This explains why our experimental results show significantly improved focal length and translation estimates when using \ac{AdHoP}, as the strategy effectively breaks the mathematical coupling that would otherwise lead to ambiguous solutions.

\subsection{Domain Shift Analysis}
\label{appendix:domain_shift_analysis}

\begin{table}[htpb]
  \centering
   \caption{\textbf{Quantitative Results of Localization on \ac{dataset} Test Sets Across Domain Configurations.} We evaluate matchers under three scenarios: same domain (reference and query from identical sources), DOP cross-domain (different orthophoto sources), and DOP+DSM cross-domain (different orthophoto and elevation model sources). For each metric, results are presented as: same domain~/~DOP cross-domain~/~DOP+DSM cross-domain. Rankings between matchers are highlighted as \colorbox{colorFst}{first}, \colorbox{colorSnd}{second}, and \colorbox{colorTrd}{third}. \textbf{Bold} values indicate the best performance within each domain configuration group. RI indicates a rotation-invariant matcher (matching performed with 4 rotated versions, selecting the one with most correspondences). Abbreviations: SuperPoint~(SP), SuperGlue~(SG), LightGlue~(LG), Minima~(MM).}
  \label{tab:xcross_loc_results}
  \setlength{\tabcolsep}{2pt}
  
\resizebox{\linewidth}{!}{\
\renewcommand{\arraystretch}{1.1}
  \begin{tabular}{c!{\vrule width 1pt}c
                  !{\vrule width 1pt}c|ccc|ccc
                  }
    \Xhline{1pt} \rowcolor{gray!15}
      \multirow{1}{*}{\textbf{Matcher}}
      & \multirow{1}{*}{\textbf{RI}} 
      & \textbf{ME} [px]$\downarrow$ & \textbf{TE} [m]$\downarrow$ & \textbf{RE} [°]$\downarrow$ & \textbf{RPE} [px]$\downarrow$ & \textbf{1m-1°} [\%]$\uparrow$ & \textbf{3m-3°} [\%]$\uparrow$ & \textbf{5m-5°} [\%]$\uparrow$ \\
    \Xhline{1pt}

    SP+SG~\cite{superpoint_18,superglue_20} & \xmark & {\bf 1.5} / 2.8 / 2.8 & {\bf 0.20} / {\fs 0.42}  / {\nd 1.09}  & {\bf {\rd 0.08} } / {\fs 0.16}  / {\nd 0.52}  & {\bf 1.6} / {\fs 3.2}  / {\nd 9.2}  & {\bf 96.2} / {\fs 75.4}  / {\rd 41.5}  & {\bf {\rd 99.3} } / {\fs 85.1}  / {\fs 83.5}  & {\bf {\nd 99.5} } / {\fs 87.4}  / {\fs 86.3}  \\ 
SP+LG~\cite{superpoint_18,lightglue_23} & \xmark & {\bf 1.5} / {\rd 2.5}  / {\rd 2.5}  & {\bf 0.21} / {\fs 0.42}  / {\fs 1.07}  & {\bf {\nd 0.07} } / {\nd 0.17}  / {\fs 0.50}  & {\bf 1.6} / {\nd 3.4}  / {\fs 9.0}  & {\bf 94.4} / {\nd 71.3}  / {\nd 42.3}  & {\bf 98.1} / {\rd 80.4}  / {\nd 80.4}  & {\bf {\rd 98.6} } / {\nd 83.3}  / {\nd 83.0}  \\ 
DeDoDe~\cite{dedode_24} & \xmark & {\bf {\rd 1.2} } / {\nd 2.2}  / {\nd 2.1}  & {\bf 0.33} / 2.89 / 3.45 & {\bf 0.13} / 1.18 / 1.56 & {\bf 2.6} / 23.3 / 28.1 & {\bf 68.5} / 1.8 / 0.7 & {\bf 76.2} / 5.5 / 4.8 & {\bf 79.0} / 6.8 / 6.6 \\ 
XFeat~\cite{xfeat_24} & \xmark & {\bf 2.6} / 210.4 / 214.9 & {\bf 0.23} / 16.43 / 27.62 & {\bf 0.09} / 8.68 / 11.88 & {\bf 1.9} / 136.0 / 243.8 & {\bf 92.8} / 28.8 / 15.7 & {\bf 94.5} / 42.5 / 41.4 & {\bf 94.7} / 45.8 / 44.5 \\ 
XFeat+LG~\cite{xfeat_24,lightglue_23} & \xmark & {\bf 1.8} / 3.8 / 3.8 & {\bf {\rd 0.19} } / {\rd 0.75}  / 1.36 & {\bf {\rd 0.08} } / {\rd 0.33}  / 0.62 & {\bf {\nd 1.4} } / 5.8 / 11.4 & {\bf {\rd 99.1} } / 56.0 / 28.3 & {\bf {\nd 99.5} } / 67.9 / 67.8 & {\bf {\nd 99.5} } / 70.1 / 70.7 \\ 
\Xhline{0.1pt}

LoFTR~\cite{loftr_21} & \xmark & {\bf 291.3} / 313.4 / 331.4 & {\bf 110.48} / 123.28 / 129.28 & {\bf 101.50} / 107.81 / 111.09 & {\bf 1198.5} / 1505.7 / 1511.8 & {\bf 30.4} / 18.3 / 10.5 & {\bf 31.5} / 22.0 / 21.4 & {\bf 32.6} / 22.4 / 21.8 \\ 
MM+LoFTR~\cite{minima_24,lodloc_24} & \xmark & 294.3 / {\bf 216.6} / 228.6 & 90.15 / {\bf 82.47} / 84.96 & 100.86 / {\bf 93.99} / 96.14 & 960.8 / 783.0 / {\bf 777.7} & {\bf 27.1} / 15.5 / 8.7 & {\bf 27.9} / 20.2 / 19.6 & {\bf 28.4} / 20.9 / 20.3 \\ 
eLoFTR~\cite{eloftr_24} & \xmark & {\bf 285.0} / 330.5 / 334.3 & {\bf 102.97} / 124.06 / 127.49 & {\bf 96.61} / 106.57 / 107.73 & {\bf 1176.8} / 1718.6 / 1745.8 & {\bf 33.2} / 19.5 / 11.8 & {\bf 35.4} / 22.8 / 21.8 & {\bf 35.9} / 23.6 / 23.2 \\ 
XoFTR~\cite{xoftr_24} & \xmark & {\bf 273.2} / 283.4 / 299.7 & {\bf 103.26} / 114.13 / 118.62 & {\bf 100.65} / 108.60 / 114.40 & {\bf 1254.4} / 1268.0 / 1322.8 & {\bf 29.5} / 20.1 / 11.8 & {\bf 29.7} / 22.4 / 20.7 & {\bf 30.0} / 23.0 / 21.8 \\ 
\Xhline{0.1pt}

DKM~\cite{dkm_23} & \cmark & {\bf {\nd 1.1} } / 96.9 / 123.0 & {\bf 0.20} / 61.30 / 60.74 & {\bf {\rd 0.08} } / 63.16 / 69.68 & {\bf {\rd 1.5} } / 518.3 / 526.9 & {\bf 65.0} / 33.1 / 19.8 & {\bf 66.1} / 40.0 / 37.5 & {\bf 66.5} / 40.7 / 38.4 \\ 
XFeat*~\cite{xfeat_24} & \xmark & {\bf 2.2} / 89.1 / 94.4 & {\bf 0.32} / 1.57 / 1.81 & {\bf 0.11} / 0.66 / 0.90 & {\bf 2.6} / 11.3 / 17.1 & {\bf 95.0} / 44.2 / 26.8 & {\bf 96.9} / 55.9 / 55.9 & {\bf 96.9} / 58.1 / 58.2 \\ 
GIM+DKM~\cite{gim_24,dkm_23} & \cmark & {\bf {\fs 1.0} } / {\fs 1.8}  / {\fs 1.8}  & {\bf {\fs 0.16} } / {\nd 0.48}  / {\rd 1.10}  & {\bf {\fs 0.05} } / {\nd 0.17}  / {\rd 0.54}  & {\bf {\fs 1.1} } / {\rd 3.6}  / {\rd 9.4}  & {\bf {\fs 100.0} } / {\rd 70.4}  / {\fs 45.5}  & {\bf {\fs 100.0} } / {\nd 80.7}  / {\rd 78.6}  & {\bf {\fs 100.0} } / {\rd 81.7}  / {\rd 79.5}  \\ 
DUSt3R~\cite{dust3r_24} & \cmark & {\bf 3.9} / 6.5 / 6.8 & {\bf 2.43} / 5.27 / 4.99 & {\bf 1.08} / 2.05 / 2.05 & {\bf 18.9} / 37.2 / 38.8 & {\bf 15.2} / 0.6 / 0.5 & {\bf 59.2} / 16.8 / 16.8 & {\bf 76.3} / 31.2 / 33.0 \\ 
MASt3R~\cite{mast3r_24} & \cmark & {\bf 1.6} / 3.2 / 3.2 & {\bf 0.25} / 1.05 / 1.48 & {\bf 0.10} / 0.44 / 0.66 & {\bf 2.0} / 8.0 / 12.5 & {\bf {\nd 99.4} } / 47.7 / 30.1 & {\bf {\fs 100.0} } / 70.3 / 70.3 & {\bf {\fs 100.0} } / 74.8 / 74.0 \\ 
RoMa~\cite{roma_24} & \cmark & {\bf {\nd 1.1} } / 5.6 / 5.1 & {\bf {\nd 0.18} } / 1.59 / 1.84 & {\bf {\nd 0.07} } / 0.60 / 0.92 & {\bf {\nd 1.4} } / 11.7 / 16.5 & {\bf 78.7} / 46.2 / 29.5 & {\bf 79.0} / 55.8 / 55.7 & {\bf 79.0} / 58.3 / 58.9 \\ 
MM+RoMa~\cite{minima_24,roma_24} & \cmark & {\bf {\nd 1.1} } / 28.3 / 33.2 & {\bf 0.20} / 3.24 / 4.57 & {\bf {\nd 0.07} } / 1.44 / 2.24 & {\bf {\rd 1.5} } / 35.1 / 46.8 & {\bf 72.1} / 39.2 / 23.0 & {\bf 72.4} / 49.1 / 47.5 & {\bf 72.4} / 51.5 / 50.2 \\ 
\Xhline{1pt}

  \end{tabular}
  }
  \vspace{-1 \baselineskip}
\end{table}

\Cref{tab:xcross_loc_results} presents the quantitative localization results using our baseline with \ac{AdHoP}, incorporating reference data from different domains.

For same-domain scenarios, the majority of the models reach high performance (except semi-dense matchers and Dust3R~\cite{dust3r_24} with recall 1m-1° below 50\%). GIM+DKM~\cite{gim_24,dkm_23} and Mast3R~\cite{mast3r_24} demonstrate particularly high accuracy in these conditions with recall 1m-1° 100\% and 99.4\%, respectively.

When employing \acp{DOP} from visually distinct domains, the increased appearance gap between query and reference data leads to noticeable performance degradation. The extent of this degradation varies significantly across matching algorithms, with some exhibiting greater robustness to appearance changes than others. Even the best performing GIM+DKM~\cite{gim_24,dkm_23} degrades by 29.6\% in cross-domain scenarios. DeDoDe~\cite{dedode_24} is very sensitive to domain shifts, with recall 1m-1° dropping drastically from 68.5\% to 1.8\%. Dust3R~\cite{dust3r_24}, on the other hand, struggles across all domains, highlighting its limitations in aerial views.

\begin{figure}[htbp]
\centering
\begin{minipage}{1\textwidth}
  \centering
  \includegraphics[width=0.32\textwidth]{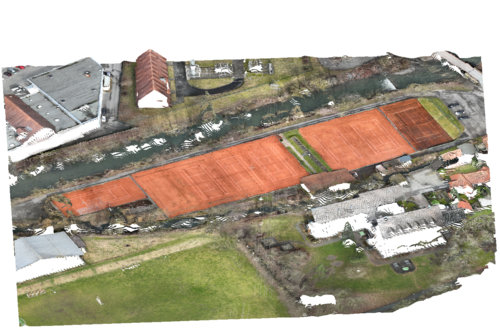}
  \includegraphics[width=0.32\textwidth]{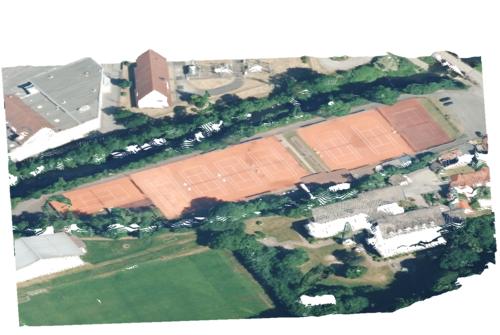}
  \includegraphics[width=0.32\textwidth]{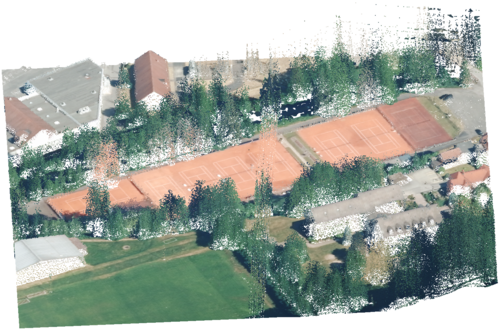}
\end{minipage}%
\hfill
\begin{minipage}{1\textwidth}
  \centering
  \includegraphics[width=0.32\textwidth]{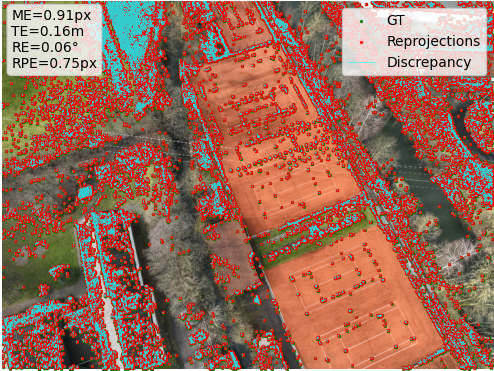}
  \includegraphics[width=0.32\textwidth]{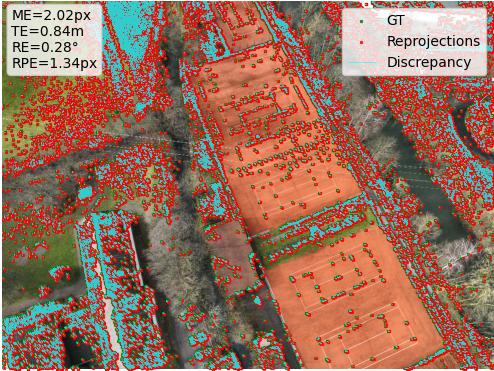}
  \includegraphics[width=0.32\textwidth]{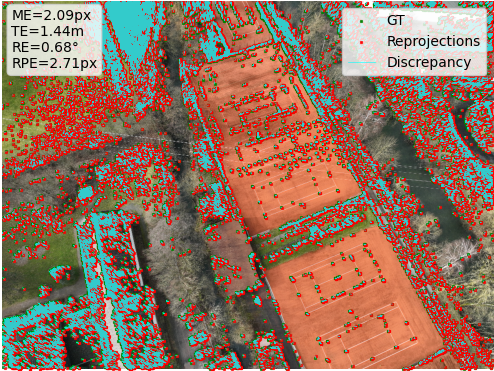}
\end{minipage}
\caption{\textbf{Example of Domain Shift Impact When Using GIM+DKM~\cite{gim_24} and \ac{AdHoP}.} No shift (left), \ac{DOP} shift (middle), \ac{DOP}+\ac{DSM} shifts (right). Top: colored point clouds (created using \ac{DSM} with colors from \ac{DOP}); Bottom: reprojections with \textcolor{darkgreen}{green} (ground-truth), \textcolor{red}{red} (estimated) points and \textcolor{cyan}{blue} discrepancy lines indicating reprojection errros.}
\label{fig:domain_shifts_results}
\end{figure}

Further increasing the domain shift by additionally incorporating \acp{DSM} from different sources generally degrades performance, highlighting the importance of geometry cues for pose estimation. Some algorithms degrade strongly as they find matches primarily on edges or object boundaries where DSMs differ significantly between sources, while others show more resilience by matching features in geometrically stable regions where elevation remains consistent across different DSM sources. An example of performance degradation is illustrated in~\Cref{fig:domain_shifts_results}.

\subsection{Resolution Analysis}
\label{appendix:resolution_analysis}

\begin{figure}[t]
\centering
\includegraphics[width=1.0\textwidth]{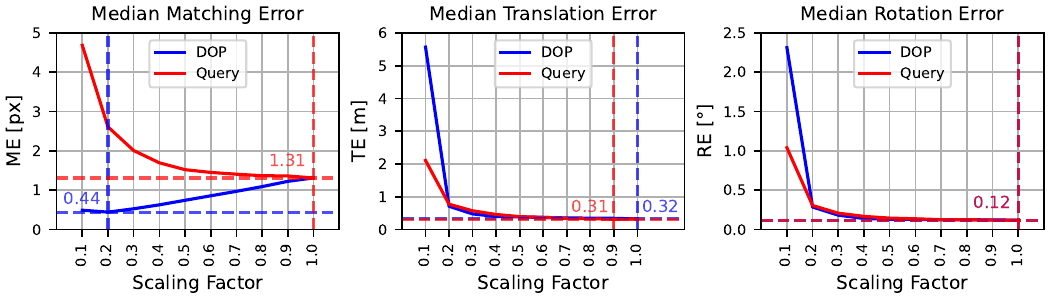}
\caption{\textbf{Resolution Impact on Localization:} Performance of GIM+DKM~\cite{gim_24,dkm_23} + \ac{AdHoP} across varying raster and query image resolutions.}
\label{fig:resolution_results}
\vspace{-0.75\baselineskip}
\end{figure}

We evaluate how raster resolution affects our lightweight localization system, an important factor for storage-constrained \ac{UAV} platforms. Additionally, we analyze the effects of query image resolution on matching performance, with results shown in~\Cref{fig:resolution_results}.

Our findings indicate that localization performance remains robust down to 512px raster resolution, with noticeable degradation only at lower resolutions. At 256px, translation error increases by 44\% and rotation error by 33\% compared to the highest resolution. This suggests that significant storage savings can be achieved with minimal performance impact by using moderately reduced resolution reference data.

Query image resolution shows similar patterns, maintaining adequate performance down to 512px before exhibiting significant degradation. The balance between computational efficiency and localization accuracy becomes particularly important for onboard processing in real-time \ac{UAV} applications.

\subsection{Covisibility Analysis}
\label{appendix:covisibility_analysis}

\begin{figure}[ht]
\centering
\includegraphics[width=1.0\textwidth]{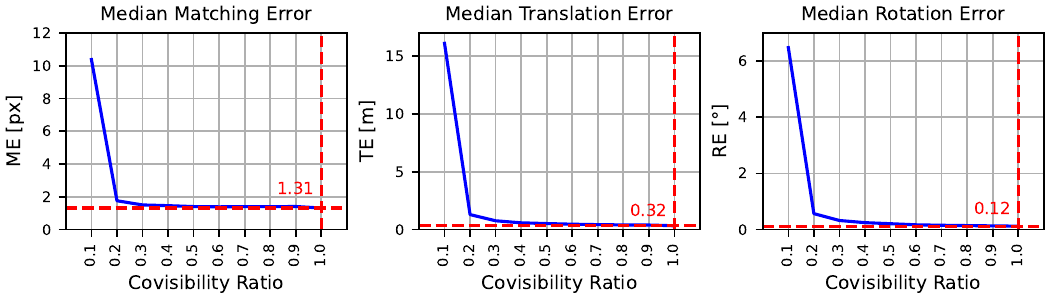}
\caption{\textbf{Covisibility Ratio Impact:} Localization performance of GIM+DKM~\cite{gim_24,dkm_23} + \ac{AdHoP} across different covisibility ratios.}
\label{fig:covisibility_error}
\vspace{-0.75\baselineskip}
\end{figure}

To assess algorithm robustness in real-world scenarios where perfect image retrieval cannot be guaranteed, we systematically reduce the covisibility ratio between query and reference images by cropping the reference raster. Our experiments reveal that having only a subset of potential correspondences significantly degrades performance, as shown in~\Cref{fig:covisibility_error}.

When the covisibility ratio drops below 20\%, we observe a sharp increase in both translation and rotation errors. This degradation occurs because the distribution of matched points becomes non-uniform across the image. This non-uniformity causes PnP to overfit to specific image regions, creating an underdetermined problem that compromises localization accuracy.

\begin{figure}[htbp]
\centering
\begin{minipage}{\textwidth}
  \centering
  \includegraphics[width=0.32\textwidth]{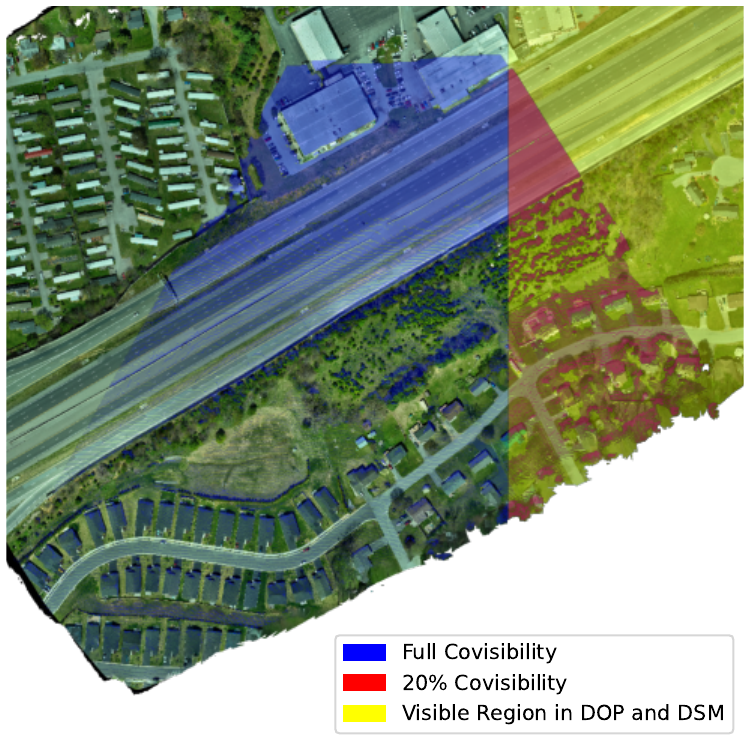}
  \includegraphics[width=0.32\textwidth]{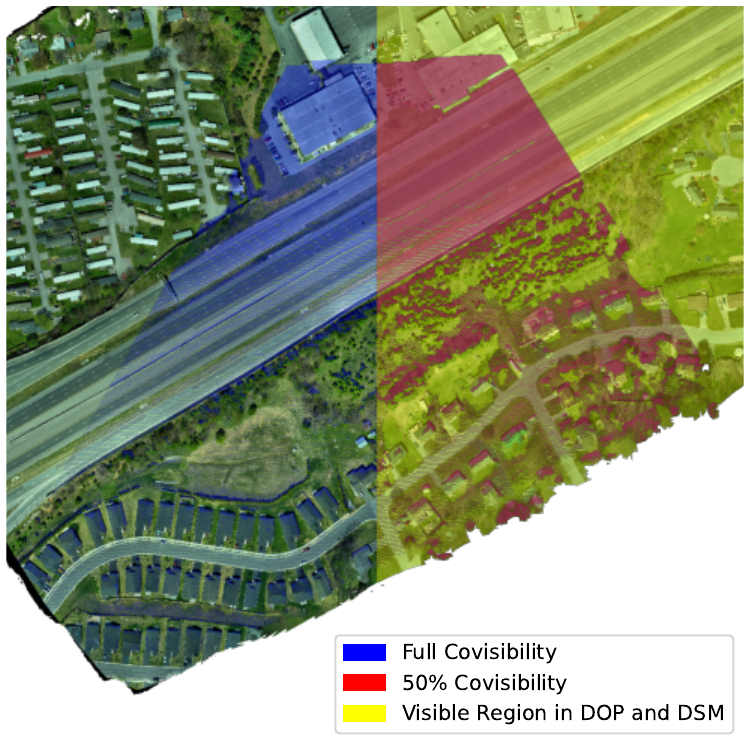}
  \includegraphics[width=0.32\textwidth]{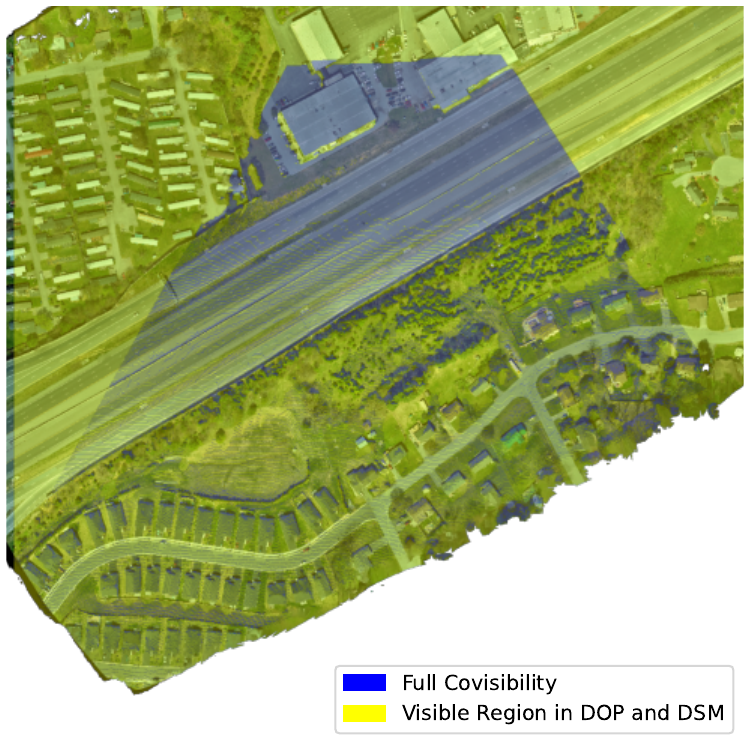}
\end{minipage}%
\hfill
\begin{minipage}{\textwidth}
  \centering
  \includegraphics[width=0.32\textwidth]{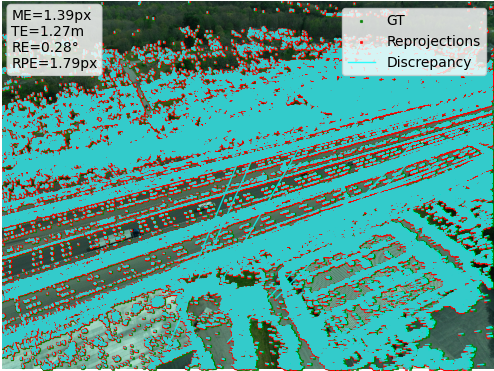}
  \includegraphics[width=0.32\textwidth]{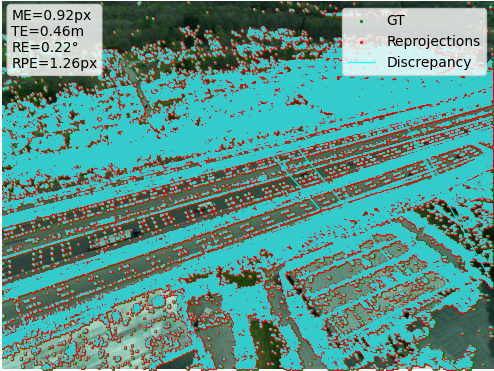}
  \includegraphics[width=0.32\textwidth]{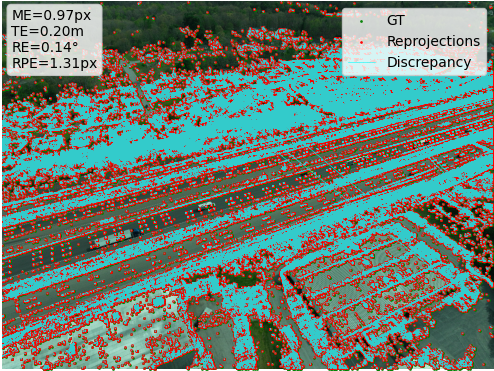}
\end{minipage}
\caption{\textbf{Covisibility Comparison in Localization Setup.} (left) 20\% coverage, (middle) 50\% coverage, (right) full coverage. The top row shows query-raster covisibility, while the bottom row displays reprojection of the keypoints $\mathcal{S}_i$ using the estimated pose. Note how the distribution of points significantly affects the quality of calibration.}
\label{fig:localization_covisibility}
\end{figure}

Figure~\ref{fig:localization_covisibility} demonstrates how the same query image produces different localization results depending on whether the reference points are well-distributed or concentrated in a particular area of the image.

These findings have important implications for real-world applications, suggesting that image retrieval systems should prioritize maximizing overlap between query and reference images.

\subsection{Utilizing Multi-Modal Data for Ground-Truth Geometry-Aware Correspondences}
Our dataset enables computing geometry-aware confidences that can guide network training. We show that filtering correspondences based purely on geometric constraints, using ground-truth data, provides perfect pose estimation.

\paragraph{Geometry-aware confidences computation.}

Given the 3D point maps and \ac{DSM} in our dataset, we establish ground-truth correspondences with associated confidence values. For each pixel $\mathbf{p}^I_i$ in the query image $I$, we:

\begin{enumerate}
    \item Backproject it to a 3D point $\mathbf{P}_i$ using the perspective camera model and the point map.
    \item Project $\mathbf{P}_i$ onto the \ac{DSM} plane using the orthographic projection model.
\end{enumerate}

This process, illustrated in \Cref{fig:stereo_projection}, reveals a fundamental limitation in perspective-to-orthographic projection with 2.5D geodata. Unlike full 3D meshes where visible points have one-to-one correspondence with 3D space, raster geodata creates a many-to-one mapping. When ray-casting from the perspective view, multiple points ($\mathbf{p}^p_i$ and $\mathbf{p}^p_j$) can map to the same orthographic position ($\mathbf{p}^o_i = \mathbf{p}^o_j$) because 2.5D representations store only a single height value per $(x,y)$ coordinate. As shown in \Cref{fig:stereo_projection}, points along vertical structures (like building facades) in the perspective view map to identical locations in the orthographic view, creating ambiguous correspondences. To address this ambiguity, we introduce a geometry-aware confidence measure $\alpha_i$ for each correspondence using: 
\begin{align}
    \alpha_i &= \exp(-\gamma \cdot d_{i}) \, , \\
    \mathbf{P}_i &= \pi_p^{-1}(\mathbf{p}^I_i) \, , \\
    d_{i} &= \|\mathbf{P}_i - \pi_o^{-1}(\pi_o(\mathbf{P}_i))\|_2 \, ,
\end{align}

where $\pi_p^{-1}$ is the perspective back-projection, $\pi_o$ is the orthographic projection, $\pi_o^{-1}$ is the orthographic back-projection (which assigns the \ac{DSM} height to the 2D coordinates), and $\gamma$ is a scaling parameter (we set it to 1). Note that the composition $\pi_o^{-1} \circ \pi_o$ is not an identity due to the dimensional reduction in orthographic projection, as illustrated in \Cref{fig:stereo_projection}.

\subsection{Are 2.5D Rasters Sufficient for Accurate Localization?}

To understand the practical potential of commonly available 2.5D geodata for UAV localization, we investigate how geometric ambiguities affect pose estimation accuracy and whether simple filtering strategies can overcome these challenges.

Table~\ref{tab:results_loc_gt} summarizes the localization results achieved using ground-truth correspondences with geometry-aware confidences. Different thresholds $\tau$ are used to filter points.

\begin{table}[thpb]
  \centering
  \caption{\textbf{Quantitative Results of Localization on OrthoLoC Test Sets (Same Domain) Using Ground-Truth Matchings.}}
  \vspace{0.2\baselineskip}
  \label{tab:results_loc_gt}

  \setlength{\tabcolsep}{2pt}
  
  \resizebox{0.8\linewidth}{!}{%
    \renewcommand{\arraystretch}{1.1}
    \begin{tabular}{c!{\vrule width 1pt}ccc|ccc}
      \Xhline{1pt}
      \rowcolor{gray!15}
      \textbf{Filtering Condition} & \textbf{TE} [m]$\downarrow$ & \textbf{RE} [°]$\downarrow$ & \textbf{RPE} [px]$\downarrow$ & 
      \textbf{1m-1°} [\%]$\uparrow$ & \textbf{3m-3°} [\%]$\uparrow$ & \textbf{5m-5°} [\%]$\uparrow$ \\ 
      \Xhline{1pt}
      $\alpha_i > 0.0$  & 0.03 & 0.00 & 0.2 & 100.0 & 100.0 & 100.0 \\ 
      $\alpha_i > 0.5$  & 0.03 & 0.01 & 0.2 & 100.0 & 100.0 & 100.0 \\ 
      $\alpha_i > 0.95$ & 0.00 & 0.00 & 0.0 & 100.0 & 100.0 & 100.0 \\ 
      $\alpha_i > 0.99$ & 0.00 & 0.00 & 0.0 & 100.0 & 100.0 & 100.0 \\ 
      \Xhline{0.1pt}
    \end{tabular}
  }
\end{table}

In the unrestricted 2.5D case ($\alpha_i > 0.0$), all valid points from the 2.5D \ac{DSM} are used, including those from vertical structures or occluded areas. This approach introduces minor errors, due to ambiguities in the many-to-one mapping of vertical structures. As $\tau$ increases, filtering progressively excludes ambiguous points, improving data purity. At higher thresholds (e.g., $\alpha_i > 0.95$), the mapping becomes close to a one-to-one relationship, and pose estimation achieves perfect localization with no observable errors.

These findings demonstrate the sufficiency of 2.5D orthographic geodata for accurate \ac{UAV} localization when paired with robust geometric filtering. By carefully selecting $\tau$, 2.5D geodata can achieve performance levels comparable to full 3D representations, motivating further research into leveraging 2.5D geodata capabilities.

\end{document}